\documentclass[10pt]{book}

\usepackage[table]{xcolor}
\usepackage{mathrsfs} % for mathscr
\usepackage{amssymb,amsmath}
\usepackage{graphicx}
\usepackage{hyperref} 
\usepackage{xspace}
\usepackage{booktabs}
\usepackage{caption} 
\usepackage{longtable}

\usepackage{pdfpages}
\usepackage{caption}

\hypersetup{colorlinks,linkcolor={red!50!black},citecolor={blue!50!black},urlcolor={blue!80!black}}
\usepackage[top=1.2in, bottom=1.2in, left=1.4in, right=1.4in]{geometry}

\usepackage{tikz}
\usetikzlibrary{shapes,calc,positioning,automata,arrows,trees}

\usepackage{listingsutf8}
\usepackage{textcomp}
\usepackage{verbatim}

\definecolor{v3lgray}{gray}{0.98}
\definecolor{v2lgray}{gray}{0.85}
\definecolor{vlgray}{gray}{0.92}
\definecolor{dgray}{rgb}{0.4,0.4,0.4}
\definecolor{dblue}{RGB}{0,0,99}
\definecolor{dred}{RGB}{175,6,54}
\definecolor{dgreen}{RGB}{47,135,7}
\definecolor{dviolet}{RGB}{102,0,153}
\definecolor{mblue}{RGB}{0,0,180}
\definecolor{dorange}{RGB}{204, 82, 0}

\def\xt{{\rm XCSP3}\xspace}

\def\xt{{\rm XCSP3}\xspace}

\newcommand{\gb}[1]{{\tt #1}} % global constraint names
\newcommand{\mn}[1]{\texttt{#1()}} % function/methods names
\newcommand{\vn}[1]{\mathtt{#1}} % variable names

\lstdefinelanguage{json}{
    basewidth  = {.6em,0.6em},
    basicstyle=\normalfont\ttfamily,
    breaklines=true,
    morestring=[b]',
    morestring=[b]", 
    sensitive=false,
    stringstyle=\color[rgb]{0.227,0.226,0.441}\ttfamily, 
    escapechar=!,
    showstringspaces=false,
    xleftmargin=20pt, %,xrightmargin=-20pt,
    breaklines=true,basicstyle=\ttfamily\small,inputencoding=utf8/latin9,texcl
}
\lstnewenvironment{json}{\lstset{language=json}}{}

\lstdefinelanguage{mcsp}{
  keywords={forall,array,block,class,implements,model,public,slide},
  basewidth  = {.6em,0.6em},
  keywordstyle=\color{dred}\bfseries,
  ndkeywords={intension,lessThan,lessEqual,greaterEqual,greaterThan,equal,different,implication,equivalence,conjunction,disjunction,extension,regular,mdd,allDifferent,allDifferentMatrix,allEqual,ordered,increasing,decreasing,strictlyIncreasing,strictlyDecreasing,lex,lexMatrix,sum,count,atMost,atLeast,exactly,atMost1,atleast1,exactly1,element,channel,maximum,minimum,cardinality,nValues,noOverlap,cumulative,instantiation,clause,circuit,minimize,maximize},
  ndkeywordstyle=\color{mblue}\bfseries,
  identifierstyle=\color{black},
  sensitive=false,
  comment=[l]{//},
  morecomment=[s]{<!--}{-->},
  commentstyle=\color{dgreen}\ttfamily,
  stringstyle=\color{dgreen}\rmfamily, %normalfont,
  morestring=[b]',
  morestring=[b]",
  escapechar=~,
  showstringspaces=false,
  classoffset=2, morekeywords={private},keywordstyle=\color{gray},
  classoffset=3, morekeywords={dom,size,when},keywordstyle=\color{dorange},
  xleftmargin=-22pt,xrightmargin=-22pt,
  breaklines=true,basicstyle=\ttfamily\footnotesize,backgroundcolor=\color{v3lgray},inputencoding=utf8/latin9,texcl
}
\lstnewenvironment{mcsp}{\lstset{language=mcsp}}{}

\title{\textcolor{dred}{XCSP3 Competition 2018\\ Proceedings}}
\author{Christophe Lecoutre and Olivier Roussel \\
CRIL CNRS, UMR 8188\\ University of Artois, France \\
\{lecoutre,roussel\}@cril.fr
}

\date{December 3, 2018} %\\~ \\\href{www.xcsp.org}{www.xcsp.org}}

\begin{document}
\maketitle

~ \\
~ \\

\bigskip

This document represent the proceedings of the \xt Competition 2018.
%This is a preliminary version containing the description of problems (models) selected for the competition.
%Soon, it will also contain the description of the solvers and the results.
%\bigskip
The website containing all {\bf detailed results} is available at:
\begin{quote}
  {\Large \href{http://www.cril.fr/XCSP18/}{http://www.cril.fr/XCSP18/}}
\end{quote}

\tableofcontents

\chapter{About the Selection of Problems in 2018}

Remember that the complete description, {\bf Version 3.0.5}, of the format (\xt) used to represent combinatorial constrained problems can be found in \cite{BLP_xcsp3}.
For the 2018 competition, we have limited \xt to its kernel, called \xt-core.
This means that the scope of \xt is restricted to:
\begin{itemize}
\item integer variables,
\item CSP and COP problems, 
\item a set of 21 popular (global) constraints for Standard tracks:
  \begin{itemize}
  \item generic constraints: \gb{intension} and \gb{extension}
  \item language-based constraints: \gb{regular} and \gb{mdd}
  \item comparison constraints: \gb{allDifferent}, \gb{allEqual}, \gb{ordered} and \gb{lex}
  \item counting/summing constraints: \gb{sum}, \gb{count}, \gb{nValues} and \gb{cardinality}
  \item connection constraints: \gb{maximum}, \gb{minimum}, \gb{element} and \gb{channel}
  \item packing/scheduling constraints: \gb{noOverlap} and \gb{cumulative}
  \item \gb{circuit}, \gb{instantiation} and \gb{slide}
  \end{itemize}
  and a small set of constraints for Mini-solver tracks.
\end{itemize}

For the 2018 competition, 41 problems have been selected.
They are succinctly presented in Table \ref{fig:problems}.
For each problem, the type of optimization is indicated (if any), as well as the involved constraints.
At this point, do note that making a good selection of problems/instances is a difficult task.
In our opinion, important criteria for a good selection are:
\begin{itemize}
\item the novelty of problems, avoiding constraint solvers to overfit already published problems;
\item the diversity of constraints, trying to represent all of the most popular constraints (those from \xt-core) while paying attention to not over-representing some of them (in particular, second class citizens);
\item the scaling up of problems.
\end{itemize}

\begin{table}
  \begin{small}
  \begin{tabular}{p{3.5cm}p{3cm}p{8cm}}
    \toprule
    Problem & Optimization & Constraints \\
    \midrule
\rowcolor{vlgray}{}    {\em Auction} & $\max$ SUM & \gb{count}, \gb{sum} \\
    {\em BACP} &  $\min$ MAXIMUM & \gb{intension}, \gb{extension}, \gb{count}, \gb{sum} \\
\rowcolor{vlgray}{}    BIBD &  & \gb{sum}, \gb{lexMatrix} \\
    Car Sequencing & & \gb{extension}, \gb{sum}, \gb{cardinality} \\
\rowcolor{vlgray}{}    Coloured Queens & & \gb{allDifferent}, \gb{allDifferentMatrix} \\
    Crosswords & & \gb{extension} \\
\rowcolor{vlgray}{}  {\em Crosswords Design} & $\max$ SUM & \gb{extension} ($*$) \\
Dubois & & \gb{extension} \\
\rowcolor{vlgray}{} {\em Eternity} & & \gb{intension}, \gb{extension}, \gb{allDifferent} \\
{\em FAPP} & $\min$ SUM & \gb{intension}, \gb{extension} \\
\rowcolor{vlgray}{} FRB & & \gb{extension} \\
Golomb Ruler & $\min$ VAR & \gb{intension}, \gb{allDifferent} \\
\rowcolor{vlgray}{} Graceful Graph & & \gb{intension}, \gb{allDifferent} \\
Graph Coloring &  $\min$ MAXIMUM & \gb{intension} \\
\rowcolor{vlgray}{} Haystacks & & \gb{extension} \\
Knapsack &  $\max$ SUM & \gb{sum} \\
\rowcolor{vlgray}{} Langford & & \gb{intension}, \gb{element} \\
Low Autocorrelation &  $\min$ SUM & \gb{intension}, \gb{sum} \\
\rowcolor{vlgray}{} Magic Hexagon & & \gb{intension}, \gb{sum} and \gb{allDifferent} \\
Magic Square & & \gb{allDifferent}, \gb{sum}, \gb{instantiation} \\
\rowcolor{vlgray}{} Mario & $\max$ SUM & \gb{intension}, \gb{extension}, \gb{sum}, \gb{circuit} \\
{\em Mistery Shopper} & & \gb{intension}, \gb{extension}, \gb{allDifferent}, \gb{lexMatrix}, \gb{channel} \\
\rowcolor{vlgray}{} {\em Nurse Rostering} &  $\min$ SUM & \gb{intension}, \gb{extension}, \gb{sum}, \gb{count}, \gb{regular}, \gb{instantiation}, \gb{slide} \\
{\em Peacable Armies} &  $\max$ SUM & \gb{intension}, \gb{sum}, \gb{count} \\
\rowcolor{vlgray}{} {\em Pizza Voucher} & $\min$ SUM & \gb{intension}, \gb{count} \\
Pseudo-Boolean &  $\min$ SUM & \gb{sum} \\
\rowcolor{vlgray}{} Quadratic Assignment & $\min$ SUM & \gb{extension}, \gb{allDifferent} \\
QuasiGroup & & \gb{intension}, \gb{allDifferentMatrix}, \gb{instantiation}, \gb{element} \\
\rowcolor{vlgray}{} RCPSP & $\min$ VAR & \gb{intension}, \gb{cumulative} \\
{\em RLFAP} & $\min$\;MAXIMUM / $\min$ NVALUES & \gb{intension}, \gb{instantiation} \\
\rowcolor{vlgray}{} Social Golfers & & \gb{intension}, \gb{instantiation}, \gb{cardinality}, \gb{lexMatrix} \\
Sports Scheduling & & \gb{intension}, \gb{extension}, \gb{instantiation}, \gb{allDifferent}, \gb{count}, \gb{cardinality} \\
\rowcolor{vlgray}{} {\em Steel Mill Slab} & $\min$ SUM & \gb{intension}, \gb{extension}, \gb{ordered}, \gb{sum} \\
Still Life & $\max$ VAR & \gb{intension}, \gb{extension}, \gb{instantiation}, \gb{sum} \\
\rowcolor{vlgray}{} Strip Packing & & \gb{intension}, \gb{extension}, \gb{noOverlap} \\
Subgraph Isomorphism & & \gb{extension}, \gb{allDifferent} \\
\rowcolor{vlgray}{} {\em Sum Coloring} & $\min$ SUM & \gb{intension} \\
{\em TAL} & $\min$ SUM & \gb{intension}, \gb{extension}, \gb{count} \\
\rowcolor{vlgray}{} {\em Template Design} & $\min$ SUM & \gb{intension}, \gb{ordered}, \gb{sum} \\
{\em Traveling Tournament} & $\min$ SUM & \gb{intension}, \gb{extension} ($*$), \gb{allDifferent}, \gb{element}, \gb{cardinality}, \gb{regular} \\
\rowcolor{vlgray}{} Travelling Salesman & $\min$ SUM & \gb{extension}, \gb{allDifferent} \\
\bottomrule
  \end{tabular}
  \end{small}
  \caption{Selected Problems for the 2018 Competition. New problems are indicated in italic font. VAR means that a variable must be optimized. For RLFAP, the type of objective differs depending on instances. When \gb{extension} is followed by ($*$), it means that short tables are considered.}\label{fig:problems}
\end{table}

\paragraph{Novelty.} More than one third of the problems are new (15 out of 41, i.e. 36.5\%). They are Auction, BACP, Crosswords Design, Eternity, FAPP, Mistery Shopper, Nurse Rostering, Peaceable Armies, Pizza Voucher, RLFAP, Steel Mill Slab, Sum Coloring, TAL, Template Design, and Traveling Tournament.
Some of these new problems are quite challenging; in particular, Crosswords design (an optimization problem with a total freedom on the position of black cells), FAPP (the original optimization instances from the ROADEF challenge), Nurse Rostering (an optimization problem involving many types of constraints), RLFAP (the original optimization instances from the ``Centre d'Electronique de l'Armement'') and TAL (an optimization problem of natural language processing).
It is important to note that the optimization FAPP and RLFAP instances, mentioned here, are far more difficult that the simplified satisfaction versions published in XCSP 2.1 some years ago. 

\paragraph{Diversity.} On the one hand, 5 constraints from \xt-core were not involved in 2018. They are \gb{mdd}, \gb{allEqual}, \gb{nValues}, \gb{minimum} and \gb{maximum}.
Very certainly, we shall try to foster \gb{mdd} in next editions because of the growing interest \cite{CY_mdd,PR_GAC4,BCHH_decision,P_thesis,VLS_CMDD} for that constraint, but note that \gb{regular} is quite related to \gb{mdd}.
The constraint \gb{allEqual} is basically an ease of modeling, because it trivially corresponds to a set of binary equality constraints.
A related form of the constraint \gb{nValues} is indirectly represented in some RLFAP instances where the type of the objective is NVALUES.
Similarly, a related form of \gb{maximum} is indirectly represented in 3 problems where the type of the objective is MAXIMUM.
On the other hand, in many problems, one can observe the presence of \gb{intension} and \gb{extension}.
The frequent occurrence of \gb{intension} is quite natural since in many problems a few primitives (e.g., like $x < y$) are required.
The frequent occurrence of \gb{extension} can be explained by the usefulness of that constraint. 
Sometimes, ordinary and short tables simply happen to be simple and natural choices for dealing with tricky situations.
This is the case when no known (global) constraint exists or when converting a logical combination of (small) constraints into a table is needed for filtering efficiency reasons. 
Basically, table constraints offer the user a direct way to handle disjunction (a choice between tuples), and this is clearly emphasized with smart tables \cite{MDL_smart}, which could be introduced in next editions (possibly, in a special track).
Another argument showing the importance of universal structures like tables, and also MDDs (Multi-valued Decision Diagrams), is the rising of tabulation techniques, i.e., the the process of converting sub-problems into tables (or MDDs), by hand, by means of heuristics \cite{AGJMNS_automatic} or by annotations \cite{DBCFM_auto}.

\paragraph{Scaling up.} It is always interesting to see how constraint solvers behave when the instances of a problem become harder and harder.
This is what we call the scaling behavior of solvers.
For most of the problems in the 2018 competition, we have selected series of instances with regular increasing difficulty.
For example, selected instances for Crosswords Design follow an increasing order (size of the grid) from 4 to 15.
Similarly, selected instances for Eternity also follow an increasing order (size of the puzzle) from 4 to 15.

\paragraph{Selection.} This year, the selection of problems and instances has been performed by Christophe Lecoutre.
As a consequence, the solver AbsCon didn't enter the competition.

%\section{About the models}

\bigskip

\chapter{Problems and Models}

In the next sections, you will find all the models that have been developed for generating the \xt instances.
All these models are written in MCSP3 1.1 \cite{L_mcsp3}, which is the new version of MCSP3, officially released in December 2018 (with a full detailed documentation).

\section{Auction}

This is Problem \href{http://www.csplib.org/Problems/prob063/}{063} on CSPLib, called Winner Determination Problem (Combinatorial Auction).

\subsection*{Description {\small (from Patrick Prosser on CSPLib)}}
\begin{quote}
 ``There is a bunch of people bidding for things. A bid has a value, and the bid is for a set of items. If we have two bids, call them A and B, and there is an intersection on the items they bid for, then we can accept bid A or bid B, but we cannot accept both of them. However, if A and B are bids on disjoint sets of items then these two bids are compatible with each other, and we might accept both. The problem then is to accept compatible bids such that we maximize the sum of the values of those bids.'' % (i.e., make most money).''
\end{quote}

\subsection*{Data}

As an illustration of data specifying an instance of this problem, we have:

%!\bef!
\begin{json} 
{
  "bids": [
    { "value": 10, "items": [1, 2] },
    { "value": 20, "items": [1, 3] },
    { "value": 30, "items": [2, 4] },
    { "value": 40, "items": [2, 3, 4] },
    { "value": 14, "items": [1] }
  ]
} 
\end{json} 
%!\aft!

\subsection*{Model}
The MCSP3 model used for the competition is:

\begin{mcsp}
class Auction implements ProblemAPI {
  Bid[] bids;

  class Bid {
    int value;
    int[] items;
  }

  public void model() {
    int[] allItems = singleValuesFrom(bids, bid -> bid.items); // distinct sorted items
    int[] bidValues = valuesFrom(bids, bid -> bid.value);
    int nBids = bids.length, nItems = allItems.length;

    Var[] b = array("b", size(nBids), dom(0, 1),
      "b[i] is 1 iff the ith bid is selected");

    forall(range(nItems), i -> {
      int[] itemBids = select(range(nBids), j -> contains(bids[j].items, allItems[i]));
      if (itemBids.length > 1) {
	Var[] scope = select(b, itemBids);
	if (modelVariant("cnt"))
  	  atMost1(scope, takingValue(1));
	if (modelVariant("sum"))
	  sum(scope, LE, 1);
      }
    }).note("avoiding intersection of bids");

    maximize(~SUM~, b, weightedBy(bidValues))
      .note("maximizing summed value of selected bids");
  }
}
\end{mcsp}
%ç\aftç

The model is rather elementary, involving 0/1 variables, and either the global constraint \gb{count} (\gb{atMost1}) with model variant 'cnt', or the global constraint \gb{sum} with model variant 'sum'.
To observe the efficiency of solvers with respect to these two global constraints, two series of 10 instances have been selected for the competition.
Also, note that only the model variant 'sum' is compatible with the restrictions imposed for the mini-track.
%The auxiliary method \mn{bidsWith} is called to compute an array containing the indexes of the bids involving a given item $v$. %a useful 2-dimensional array from original data.

\section{BACP}

This is Problem \href{http://www.csplib.org/Problems/prob030/}{030} on CSPLib, called Balanced Academic Curriculum Problem (BACP).

\subsection*{Description {\small (from Brahim Hnich, Zeynep Kiziltan and Toby Walsh on CSPLib)}}
\begin{quote}
 ``The BACP is to design a balanced academic curriculum by assigning periods to courses in a way that the academic load of each period is balanced, i.e., as similar as possible.
  An academic curriculum is defined by a set of courses and a set of prerequisite relationships among them.
  Courses must be assigned within a maximum number of academic periods.
  Each course has associated a number of credits or units that represent the academic effort required to successfully follow it.
  The curriculum must obey the following regulations:
  \begin{itemize}
    \item Minimum academic load: a minimum number of academic credits per period is required to consider a student as full time.
    \item Maximum academic load: a maximum number of academic credits per period is allowed in order to avoid overload.
    \item Minimum number of courses: a minimum number of courses per period is required to consider a student as full time.
    \item Maximum number of courses: a maximum number of courses per period is allowed in order to avoid overload.
  \end{itemize}
The goal is to assign a period to every course in a way that the minimum and maximum academic load for each period, the minimum and maximum number of courses for each period, and the prerequisite relationships are satisfied. An optimal balanced curriculum minimizes the maximum academic load for all periods.''
\end{quote}

\subsection*{Data}

As an illustration of data specifying an instance of this problem, we have:
\begin{json}
{
  "nPeriods": 5,
  "minCredits": 6,
  "maxCredits": 15,
  "minCourses": 2,
  "maxCourses": 6,
  "credits": [2, 3, 2, 4, 1, 3, 3, 3, 3, 3, 3, 3, 2, 3, 3, 3],
  "prerequisites": [[6,0], [7,5], [10,4], [10,5], [11,10], [13,8], [14,8], [15,9]]
}
\end{json}

\subsection*{Model}
The MCSP3 model used for the competition is: 

\begin{mcsp}
class Bacp implements ProblemAPI {
  int nPeriods;
  int minCredits, maxCredits;
  int minCourses, maxCourses;
  int[] credits;
  int[][] prerequisites;

  private int[][] channelingTable(int c) {
    int[][] tuples = new int[nPeriods][nPeriods + 1];
    for (int p = 0; p < nPeriods; p++) {
      tuples[p][p] = credits[c];
      tuples[p][nPeriods] = p;
    }
    return tuples;
  }
  
  public void model() {
    int nCourses = credits.length, nPrerequisites = prerequisites.length;

    Var[] s = array("s", size(nCourses), dom(range(nPeriods)),
      "s[c] is the period (schedule) for course c");
    Var[] co = array("co", size(nPeriods), dom(range(minCourses, maxCourses + 1)),
      "co[p] is the number of courses at period p");
    Var[] cr = array("cr", size(nPeriods), dom(range(minCredits, maxCredits + 1)),
      "cr[p] is the number of credits at period p");
    Var[][] cp = array("cp", size(nCourses, nPeriods), (c, p) -> dom(0, credits[c]),
      "cp[c][p] is 0 if course c is not planned at period p, the number of credits for c otherwise");

    forall(range(nCourses), c -> extension(vars(cp[c], s[c]), channelingTable(c)))
      .note("channeling between arrays cp and s");
    forall(range(nPeriods), p -> count(s, takingValue(p), EQ, co[p]))
      .note("counting the number of courses in each period");
    forall(range(nPeriods), p -> sum(columnOf(cp, p), EQ, cr[p]))
      .note("counting the number of credits in each period");  
    forall(range(nPrerequisites), i -> lessThan(s[prerequisites[i][0]],s[prerequisites[i][1]]))
      .note("handling prerequisites");

    minimize(~MAXIMUM~, cr)
      .note("minimizing the maximum number of credits in periods");

    decisionVariables(s);
  }
}
\end{mcsp}

This model involves 4 arrays of variables and 4 types of constraints: \gb{extension}, \gb{count}, \gb{sum} and \gb{intension} (primitive \gb{lessThan}).
Actually, two series 'm1' and 'm2' of 12 instances each have been selected. The series 'm1' corresponds to the model described above whereas 'm2' (obtained after some reformulations) is compatible with the restrictions imposed for the mini-track.
Decision variables are indicated, except for the instances (\xt files) whose name contains 'nodv' (no decision variables).

\section{BIBD}

This is Problem \href{http://www.csplib.org/Problems/prob028/}{028} on CSPLib, called Balanced Incomplete Block Designs (BIBD).

\subsection*{Description {\small (from Steven Prestwich on CSPLib)}}
\begin{quote}
  ``BIBD generation is described in most standard textbooks on combinatorics. A BIBD is defined as an arrangement of $v$ distinct objects into $b$ blocks such that each block contains exactly $k$ distinct objects, each object occurs in exactly $r$ different blocks, and every two distinct objects occur together in exactly $\lambda$ blocks.
  Another way of defining a BIBD is in terms of its incidence matrix, which is a $v$ by $b$ binary matrix with exactly $r$ ones per row, $k$ ones per column, and with a scalar product of $\lambda$ between any pair of distinct rows.
  A BIBD is therefore specified by its parameters $(v,b,r,k,\lambda)$''
\end{quote}

\subsection*{Data}

As an illustration of data specifying an instance of this problem, we have $(v=7, b=7, r=3, k=3, lambda=1)$.

\subsection*{Model}

The MCSP3 model used for the competition is: 

\begin{mcsp}
class Bibd implements ProblemAPI {
  int v, b, r, k, lambda;

  public void model() {
    b = b != 0 ? b : (lambda * v * (v-1)) / (k * (k-1)); // when b is 0, we compute it
    r = r != 0 ? r : (lambda * (v-1)) / (k-1);           // when r is 0, we compute it

    Var[][] x = array("x", size(v, b), dom(0, 1),
      "x[i][j] is the value at row i and column j of the matrix");

    forall(range(v), i -> sum(x[i], EQ, r))
      .note("constraints on rows");
    forall(range(b), j -> sum(columnOf(x, j), EQ, k))
      .note("constraints on columns");
    forall(range(v), i -> forall(range(i + 1, v), j -> sum(x[i], weightedBy(x[j]), EQ, lambda)))
      .note("scalar constraints with respect to lambda");

    lexMatrix(x, ~INCREASING~).tag(SYMMETRY_BREAKING);
      .note("Increasingly ordering both rows and columns")
  }
}
\end{mcsp}

This model involves 1 array of variables and 2 types of constraints: \gb{sum} and \gb{lexMatrix}, the latter being used to break some variable symmetries.
Two series 'sum' and 'sc' of 6 instances each have been selected.
The series 'sum' corresponds to the model variant described above whereas 'sc' is a model variant obtained by introducing auxiliary variables.

\section{Car Sequencing}

This is Problem \href{http://www.csplib.org/Problems/prob001/}{001} on CSPLib. %, proposed by Barbara Smith.

\subsection*{Description {\small (from Barbara Smith on CSPLib)}}

\begin{quote}
``A number of cars are to be produced; they are not identical, because different options are available as variants on the basic model. The assembly line has different stations which install the various options (air-conditioning, sun-roof, etc.). These stations have been designed to handle at most a certain percentage of the cars passing along the assembly line. Furthermore, the cars requiring a certain option must not be bunched together, otherwise the station will not be able to cope. Consequently, the cars must be arranged in a sequence so that the capacity of each station is never exceeded. For instance, if a particular station can only cope with at most half of the cars passing along the line, the sequence must be built so that at most 1 car in any 2 requires that option.'' 
\end{quote}

\subsection*{Data}

As an illustration of data specifying an instance of this problem, we have:
\begin{json}
{
  "carClasses": [
    { "demand": 1, "options": [1, 0, 1, 1, 0] },
    { "demand": 1, "options": [0, 0, 0, 1, 0] },
    { "demand": 2, "options": [0, 1, 0, 0, 1] },
    { "demand": 2, "options": [0, 1, 0, 1, 0] },
    { "demand": 2, "options": [1, 0, 1, 0, 0] },
    { "demand": 2, "options": [1, 1, 0, 0, 0] }
  ],
  "optionLimits": [
    { "num": 1, "den": 2 },
    { "num": 2, "den": 3 },
    { "num": 1, "den": 3 },
    { "num": 2, "den": 5 },
    { "num": 1, "den": 5 }
  ]
}
\end{json}

\subsection*{Model}

The MCSP3 model used for the competition is: 

\begin{mcsp}
class CarSequencing implements ProblemAPI {
  CarClass[] classes;
  OptionLimit[] limits;

  class CarClass {
    int demand;
    int[] options;
  }

  class OptionLimit {
    int num;
    int den;
  }

  private Table channelingTable() {
    Table table = table();
    for (int i = 0; i < classes.length; i++)
      table.add(i, classes[i].options); // indexing car class options
    return table;
  }

  public void model() {
    int[] demands = valuesFrom(classes, cla -> cla.demand);
    int nCars = sumOf(demands), nOptions = limits.length, nClasses = classes.length;
    Range allClasses = range(nClasses);
    
    Var[] c = array("c", size(nCars), dom(allClasses)),
      "c[i] is the class of the ith assembled car");
    Var[][] o = array("o", size(nCars, nOptions), dom(0, 1),
      "o[i][k] is 1 if the ith assembled car has option k");
    
    cardinality(c, allClasses, occurExactly(demands))
      .note("building the right numbers of cars per class");

    forall(range(nCars), i -> extension(vars(c[i], o[i]), channelingTable()))
      .note("linking cars and options");

    forall(range(nOptions).range(nCars), (k, i) -> {
      if (i <= nCars - limits[k].den) {
        Var[] scp = select(columnOf(o, k), range(i, i + limits[k].den));
        sum(scp, LE, limits[k].num);
      }
    }).note("constraints about option frequencies"); 

    forall(range(nOptions).range(nCars), (k, i) -> {
      // i stands for the number of blocks set to the maximal capacity
      int nOptionOccurrences = sumOf(valuesFrom(classes, cla -> cla.options[k] * cla.demand));
      int nOptionsRemainingToSet = nOptionOccurrences - i * limits[k].num;
      int nOptionsPossibleToSet = nCars - i * limits[k].den;
      if (nOptionsRemainingToSet > 0 && nOptionsPossibleToSet > 0) {
        Var[] scp = select(columnOf(o, k), range(nOptionsPossibleToSet));
        sum(scp, GE, nOptionsRemainingToSet);
      }
    }).tag(REDUNDANT_CONSTRAINTS);
  }
}
\end{mcsp}

This model involves 2 arrays of variables and 3 types of constraints: \gb{cardinality}, \gb{extension} and \gb{sum}.
Note that instead of posting \gb{extension} constraints, we could have used binary \gb{intension} constraints with a predicate like $c_i = j \Rightarrow o_{i,k} = v$ where $v$ is the value (0 or 1) of the kth option of the jth class.
Also, note that we could have used a cache for the table built by \mn{matchs}.
The last group of constraints corresponds to redundant constraints.
A series of 17 instances has been selected for the competition.

\section{Coloured Queens}

\subsection*{Description}
\begin{quote}
  The queens graph is a graph with n*n nodes corresponding to the squares of a chess-board. There is an edge between nodes iff they are on the same
  row, column, or diagonal, i.e., if two queens on those squares would attack each other. The coloring problem is to color the queens graph with $n$
  colors. See \cite{KLR_developments}.
\end{quote}

\subsection*{Data}

As an illustration of data specifying an instance of this problem, we simply have $n=8$.

\subsection*{Model}

The MCSP3 model used for the competition is: 

\begin{mcsp}
class ColouredQueens implements ProblemAPI {
  int n;

  public void model() {
    Var[][] x = array("x", size(n, n), dom(range(n)),
      "x[i][j] is the color at row i and column j");
    Var[][] dn = diagonalsDown(x), up = diagonalsUp(x); // precomputing scopes

    allDifferentMatrix(x)
      .note("different colors on rows and columns");
    forall(range(dn.length), i -> allDifferent(dn[i])
      .note("different colors on downward diagonals"));
    forall(range(up.length), i -> allDifferent(up[i])
      .note("different colors on upward diagonals"));
  }
} 
\end{mcsp}

This model only involves 1 array of variables and 2 types of constraints: \gb{allDifferent} and \gb{allDifferentMatrix}.
A series of 12 instances has been selected for the competition.

\section{Crosswords (Satisfaction)}

This problem has already been used in previous XCSP competitions, because it notably permits to compare filtering algorithms on large table constraints.
%Hence, we very succinctly introduce it.

\subsection*{Description}
\begin{quote}
``Given a grid with imposed black cells (spots) and a dictionary, the problem is to fulfill the grid with the words contained in the dictionary.''
\end{quote}

\subsection*{Data}

As an illustration of data specifying an instance of this problem, we have:
\begin{json}
{
  "spots": [[0,1,0,0,0], [0,0,0,0,0], [0,0,1,0,0], [0,0,0,0,0], [0,0,0,0,1]],
  "dictFileName": "ogd"
}
\end{json}

\subsection*{Model}

The MCSP3 model used for the competition is: 

\begin{mcsp}
class Crossword implements ProblemAPI {
  int[][] spots;
  String dictFileName;
  
  private Map<Integer, List<int[]>> loadWords() {
    Map<Integer, List<int[]>> words = new HashMap<>();
    readFileLines(dictFileName).forEach(w -> 
      words.computeIfAbsent(w.length(), k -> new ArrayList<>()).add(Utilities.wordAsIntArray(w))
    );
    return words;
  }

  private class Hole {
    int row, col, ~size~;
    boolean horizontal;
    
    Hole(int row, int col, int ~size~, boolean horizontal) {
      this.row = horizontal ? row : col;
      this.col = horizontal ? col : row;
      this.~size~ = ~size~;
      this.horizontal = horizontal;
    }
    
    Var[] scope(Var[][] x) {
      return variablesFrom(range(~size~), i -> horizontal ? x[row][col + i] : x[row + i][col]);
    }
  }

  private List<Hole> findHoles(int[][] t, boolean untransposed) {
    int nRows = t.length, nCols = t[0].length;
    List<Hole> list = new ArrayList<>();
    for (int i = 0; i < nRows; i++) {
      int start = -1;
      for (int j = 0; j < nCols; j++)
        if (t[i][j] == 1) { // if spot (black cell)
	  if (start != -1 && j - start >= 2)
	    list.add(new Hole(i, start, j - start, untransposed));
	  start = -1;
        } else {
	  if (start == -1)
	    start = j;
	  else if (j == nCols - 1 && nCols - start >= 2)
	    list.add(new Hole(i, start, nCols - start, untransposed));
        }
    }
    return list;
  }

  private Hole[] findHoles() {
    List<Hole> list = findHoles(spots, true);
    list.addAll(findHoles(transpose(spots), false));
    return list.toArray(new Hole[0]);
  }
      
  public void model() {
    Map<Integer, List<int[]>> words = loadWords();
    Hole[] holes = findHoles();
    int nRows = spots.length, nCols = spots[0].length, nHoles = holes.length;

    Var[][] x = array("x", size(nRows, nCols), (i, j) -> dom(range(26)).when(spots[i][j] == 0),
      "x[i][j] is the letter, number from 0 to 25, at row i and column j (when no spot)");
    
    forall(range(nHoles), i -> extension(holes[i].scope(x), words.get(holes[i].~size~)))
      .note("fill the grid with words");
  }    
}
\end{mcsp}

This satisfaction problem only involves 1 array of variables and 1 type of constraints: \gb{extension} (ordinary table constraints).
For clarity, we use an auxiliary class \texttt{Hole}.
A series of 13 instances, with only blank grids, has been selected for the competition.

\section{Crosswords (Optimization)}

This problem is the subject of a regular Romanian challenge, and has been studied in XX. %\cite{LM_subtables}.

\subsection*{Description}
\begin{quote}
``Given a main dictionary containing ordinary words, and a second dictionary containing thematic words, the objective is to fill up a grid with words of both dictionary.
  Each word from the thematic word has a value (benefit) equal to its length.
  The objective is to maximize the overall value.
  The problem is difficult because black cells are not imposed, i.e., can be put anywhere in the grid (but no adjacency of black cells is authorized).''
\end{quote}

\subsection*{Data}

As an illustration of data specifying an instance of this problem, we have:
\begin{json}
{
  "n": 10,
  "nMaxWords": 5,
  "mainDict": "mainDictRomanian.txt",
  "thematicDict": "thematicDictRomanian2017.txt"
}
\end{json}

\subsection*{Model}
The MCSP3 model used for the competition is: 

\begin{mcsp}
class CrosswordDesign implements ProblemAPI {
  int n; // size of the grid (number of rows and number of columns)
  int nMaxWords; // maximum number of words that can be put on a same row or column
  String mainDict, thematicDict;

  @NotData
  String[] words; // words of the two merged dictionaries

  @NotData
  int[] wordsPoints; // value of each word

  private void loadWords() {
    words = Stream.concat(readFileLines(mainDict), readFileLines(thematicDict))).toArray(String[]::new);
    wordsPoints = new int[words.length];
    List<String> list = Arrays.asList(words);
    readFileLines(thematicDict).forEach(w -> {
      int pos = list.indexOf(w);
      if (pos != -1)
        wordsPoints[pos] = w.length(); // thematic words have some value (their lengths)
    });
  }
  
  private Table shortTable(int k) {
    boolean lastWord = k == nMaxWords - 1;
    List<int[]> list = new ArrayList<>();
    if (k != 0)
      list.add(range(n + 4).map(i -> i == 0 || i == 1 | i == 3 ? -1 : i == 2 ? 0 : STAR));
    int[] possiblePositions = k == 0 ? vals(0, 1) : vals(range(2 * k, n));
    for (int p : possiblePositions)
      for (int i = 0; i < words.length; i++) {
        int bp = p + words[i].length(); // position of the black point, just after the word
        if (bp <= n) {
	  int[] tuple = new int[n + 4];
	  tuple[0] = p;
	  tuple[1] = i;
	  tuple[2] = wordsPoints[i];
	  if (lastWord && bp < n - 1)
	    continue;
	  tuple[3] = lastWord ? -1 : bp <= n - 2 ? bp + 1 : -1;
	  for (int j = 4; j < tuple.length; j++) {
	    if (j - 4 == p - 1 || j - 4 == bp)
	      tuple[j] = 26; // black points
	    else if (p <= j - 4 && j - 4 < p + words[i].length())
	      tuple[j] = words[i].charAt(j - 4 - p) - 97;
	    else
	      tuple[j] = STAR;
	  }
	  list.add(tuple);
        }
      }
    return table().add(list);
  }

  private int[] positionValues(int k) {
    return k == 0 ? vals(0, 1) : k == nMaxWords ? vals(-1) : vals(range(-1, n));
  }
  
  public void model() {
    loadWords();
    int nWords = words.length;
    
    Var[][] x = array("x", size(n, n), dom(range(27)),
      "x[i][j] is the (number for) letter at row i and col j; 26 stands for a black point");
    Var[][] r = array("r", size(n, nMaxWords), dom(range(-1, nWords)),
      "r[i][k] is the (index of) kth word at row i; -1 means no word");
    Var[][] c = array("c", size(n, nMaxWords), dom(range(-1, nWords)),
      "c[j][k] is the (index of) kth word at col j; -1 means no word");
    Var[][] pr = array("pr", size(n, nMaxWords + 1), (i, k) -> dom(positionValues(k)),
      "pr[i][k] is the position (index of col) of the kth word at row i; -1 means no word");
    Var[][] pc = array("pc", size(n, nMaxWords + 1), (j, k) -> dom(positionValues(k)),
      "pc[j][k] is the position (index of row) of the kth word at col j; -1 means no word");
    Var[][] br = array("br", size(n, nMaxWords), dom(range(n + 1)),
      "br[i][k] is the benefit of the kth word at row i");
    Var[][] bc = array("bc", size(n, nMaxWords), dom(range(n + 1)),
      "bc[j][k] is the benefit of the kth word at col j");
 
    forall(range(n).range(nMaxWords), (i, k) ->
      extension(vars(pr[i][k], r[i][k], br[i][k], pr[i][k+1], x[i]), shortTable(k)))
    .note("putting words on rows");

    forall(range(n).range(nMaxWords), (j, k) ->
      extension(vars(pc[j][k], c[j][k], bc[j][k], pc[j][k+1], columnOf(x, j)), shortTable(k)))
    .note("putting words on columns");

      maximize(~SUM~, vars(br, bc))
        .note("maximizing the summed benefit of words put in the grid");
  }
}
\end{mcsp}

This optimization problem involves 7 arrays of variables and simply the constraint \gb{extension}. However, do note that such constraints are built with large short tables (i.e., tables involving '*', denoted by STAR in the code).
The auxiliary methods \mn{loadWords} and \mn{shortTable} are respectively useful for loading the dictionaries and computing the short tables.
A series of 13 instances has been selected for the competition.

\section{Dubois}

This problem has been conceived by Olivier Dubois, and submitted to the second DIMACS Implementation Challenge.
Dubois's generator produces contradictory 3-SAT instances that seem very difficult to be solved by any general method.

\subsection*{Description}
\begin{quote}
``Given an integer $n$, called the degree, Dubois's process allows us to construct a 3-SAT contradictory instance with $3 \times n$ variables and $2 \times n$ clauses, each of them having 3 literals.''
\end{quote}

\subsection*{Data}

As an illustration of data specifying an instance of this problem, we simply have $n=10$.

\subsection*{Model}

The MCSP3 model used for the competition is: 

\begin{mcsp}
class Dubois implements ProblemAPI {
  int n;

  public void model() {
    Table table1 = table("(0,0,1)(0,1,0)(1,0,0)(1,1,1)"), table2 = table("(0,0,0)(0,1,1)(1,0,1)(1,1,0)");

    Var[] x = array("x", size(3 * n), dom(0, 1))
      .note("x[i] is the Boolean value (0/1) of the ith variable of Dubois's sequence");

    extension(vars(x[2*n - 2], x[2*n - 1], x[0]), table1);
    forall(range(n - 2), i -> extension(vars(x[i], x[2*n + i], x[i + 1]), table1));
    forall(range(2), i -> extension(vars(x[n - 2 + i], x[3*n - 2], x[3 * n - 1]), table1));
    forall(range(n, 2*n - 2), i -> extension(vars(x[i], x[4*n - 3 - i], x[i - 1]), table1));
    extension(vars(x[2 * n - 2], x[2 * n - 1], x[2 * n - 3]), table2);
  }
}
\end{mcsp}

This model involves 1 array of variables and 1 type of constraints: \gb{extension}.
A series of 12 instances has been selected for the competition.

\section{Eternity}

Eternity II is a famous edge-matching puzzle, released in July 2007 by TOMY, with a 2 million dollars prize for the first submitted solution. See, for example, \cite{BB_fast}.
Here, we are interested in instances derived from the original problem by the \href{http://becool.info.ucl.ac.be/}{BeCool} team of the UCL (``Universit\'e Catholique de Louvain'') who proposed them for the competition.

\subsection*{Description}
\begin{quote}
``On a board of size $n \times m$, you have to put square tiles (pieces) that are described by four colors (one for each direction : top, right, bottom and left).
All adjacent tiles on the board must have matching colors along their common edge. All edges must have color '0' on the border of the board.''
\end{quote}

\subsection*{Data}

As an illustration of data specifying an instance of this problem, we have:
\begin{json}
{
  "n": 3,
  "m": 3,
  "pieces": [[0,0,1,1], [0,0,1,2], [0,0,2,1], [0,0,2,2], [0,1,3,2], [0,1,4,1], [0,2,3,1], [0,2,4,2], [3,3,4,4]]
}
\end{json}

\subsection*{Model}

The MCSP3 model used for the competition is: 

\begin{mcsp}
class Eternity implements ProblemAPI {
  int n, m;
  int[][] pieces;

  private Table piecesTable() {
    Table table = table();
    for (int i = 0; i < n * m; i++)
      for (int r = 0; r < 4; r++) // handling rotation
        table.add(i, pieces[i][r % 4], pieces[i][(r+1) %4 ], pieces[i][(r+2) % 4], pieces[i][(r+3) % 4]);
    return table;
  }
    
  public void model() {
    int maxValue = maxOf(valuesIn(pieces)); // max possible value on pieces

    Var[][] id = array("id", size(n, m), dom(range(n * m)),
      "id[i][j] is the id of the piece at row i and column j");
    Var[][] top = array("top", size(n, m), dom(range(0, maxValue)),
      "top[i][j] is the value at the top of the piece put at row i and column j");
    Var[][] left = array("left", size(n, m), dom(range(0, maxValue)),
      "left[i][j] is the value at the left of the piece put at row i and column j");
    Var[] bot = array("bot", size(m), dom(range(0, maxValue)),
      "bot[j] is the value at the bottom of the piece put at the bottommost row and column j");
    Var[] right = array("right", size(n), dom(range(0, maxValue)),
      "right[i] is the value at the right of the piece put at the row i and the rightmost column");

    allDifferent(id)
      .note("all pieces must be placed (only once)");

    forall(range(n).range(m), (i, j) -> {
      Var lr = j < m - 1 ? left[i][j + 1] : right[j], tb = i < n - 1 ? top[i + 1][j] : bot[j];
      extension(vars(id[i][j], top[i][j], lr, tb, left[i][j]), piecesTable());
    }).note("pieces must be valid (i.e. correspond to those given initially, possibly after rotation)");

    block(() -> {
      forall(range(n), i -> equal(left[i][0], 0));
      forall(range(n), i -> equal(right[i], 0));
      forall(range(m), j -> equal(top[0][j], 0));
      forall(range(m), j -> equal(bot[j], 0));
    }).note("put special value 0 on borders");
  }
}
\end{mcsp}

This model involves 5 arrays of variables and 3 types of constraints: \gb{allDifferent}, \gb{extension} and \gb{intension} (primitive \gb{equal}).
Note that we could have stored and reused the table, instead of creating it systematically. 
A series of 15 instances has been selected for the competition.

\section{FAPP}

The frequency assignment problem with polarization constraints (FAPP) is an optimization problem\footnote{This is an extended subject of the CALMA European project} that was part of the ROADEF'2001 challenge\footnote{See \url{http://uma.ensta.fr/conf/roadef-2001-challenge/}}.
In this problem, there are constraints concerning frequencies and polarizations of radio links.
Progressive relaxation of these constraints is explored: the relaxation level is between 0 (no relaxation) and 10 (the maximum relaxation).  
Whereas we used simplified CSP instances of this problem in previous XCSP competitions, do note here that we have considered the original COP instances. 

\subsection*{Description}
\begin{quote}
The description is rather complex. Hence, we refer the reader to:\\ \href{https://uma.ensta-paristech.fr/conf/roadef-2001-challenge/distrib/fapp\_roadef01\_rev2\_tex.pdf}{https://uma.ensta-paristech.fr/conf/roadef-2001-challenge/distrib/fapp\_roadef01\_rev2\_tex.pdf}
\end{quote}

\subsection*{Data}

As an illustration of data specifying an instance of this problem, we have:
\begin{json}
{
  "domains": {
    "0": [1, 2, 3, 4, 5, 6, 7, 8, 9, 10, 11, 12, 13, 14, 15, 16, 17, 18, 19],
    "1": [25, 26, 27, 28, 29, 30, 31, 32, 33, 34, 35, 36, 37, 38, 39, 40],
    "2": [55, 56, 57, 63, 64, 65]},
  "routes": [
    { "domain": 1, "polarization": -1 },
    { "domain": 1, "polarization": 0 },
    ...
  ],
  "hards": [
    { "route1": 1, "route2": 2, "frequency": true, "equality": true, "gap": 36 },
    { "route1": 0, "route2": 1, "frequency": true, "equality": true, "gap": 0 },
    ...
  ],
  "softs": [
    { "route1": 0, "route2": 2, "eqRelaxations": [46,44,42,42,40,40,35,35,35,30,20],
      "neRelaxations": [39,37,35,35,33,33,28,28,28,23,13] },
    { "route1": 0, "route2": 8, "eqRelaxations": [30,29,28,27,26,25,24,23,22,21,20],
      "neRelaxations": [29,28,27,26,25,24,23,22,21,20,19] },
    { "route1": 1, "route2": 3, "eqRelaxations": [33,32,30,30,29,28,27,22,17,12,7],
      "neRelaxations": [29,28,27,27,27,26,25,20,15,10,5] },
    ...
  ]
}
\end{json}

\subsection*{Model}

The MCSP3 model used for the competition is: 

\begin{mcsp}
class Fapp implements ProblemAPI {

  Map<Integer, int[]> domains;
  Route[] routes;
  HardCtr[] hards;
  SoftCtr[] softs;
  
  class Route {
    int domain, polarization;

    int[] polarizationValues() {
      return polarization == 0 ? vals(0, 1) : polarization == 1 ? vals(1) : vals(0);
    }
  }
  
  class HardCtr {
    int route1, route2;
    boolean frequency, equality;
    int gap;
  }
  
  class SoftCtr {
    int route1, route2;
    int[] eqRelaxations;
    int[] neRelaxations;
  }
  
  private boolean softLink(int i, int j) {
    return firstFrom(softs, c -> c.route1 == i && c.route2 == j || c.route1 == j && c.route2 ==i) != null;
  }
  
  private int[] distances(int i, int j) {
    int[] dom1 = domains.get(routes[i].domain), dom2 = domains.get(routes[j].domain);
    return IntStream.of(dom1).flatMap(f1 -> IntStream.of(dom2).map(f2 -> Math.abs(f1 - f2))).toArray(); 
  }
 
  private CtrEntity imperativeConstraint(Var[] f, Var[] p, HardCtr c) {
    int i = c.route1, j = c.route2;
    if (c.frequency) {
      if (c.gap == 0)
        return c.equality ? equal(f[i], f[j]) : different(f[i], f[j]);
      else
        return c.equality ? equal(dist(f[i], f[j]), c.gap) : different(dist(f[i], f[j]), c.gap);
    }
    return c.equality ? equal(p[i], p[j]) : different(p[i], p[j]);
  }

  private Table relaxTable(SoftCtr c) {
    Table table = table();
    Set<Integer> set = new HashSet<>();
    int i = c.route1, j = c.route2;
    for (int fi : domains.get(routes[i].domain))
      for (int fj : domains.get(routes[j].domain)) {
        int dist = Math.abs(fi - fj);
        if (set.contains(dist))
          continue; // because already encountered
        for (int pol = 0; pol < 4; pol++) {
	  int pi = pol < 2 ? 0 : 1;
	  int pj = pol == 1 || pol == 3 ? 1 : 0;
	  if (routes[i].polarization == 1 && pi == 0 || routes[j].polarization == 1 && pj == 0)
	    continue;
	  if (routes[i].polarization == -1 && pi == 1 || routes[j].polarization == -1 && pj == 1)
	    continue;
	  int[] t = pi == pj ? c.eqRelaxations : c.neRelaxations;
	  for (int k = 0; k <= 11; k++) {
	    if (k == 11 || dist >= t[k]) { // for k=11, we suppose t[k] = 0
	      int ~sum~ = IntStream.range(0, k - 1).map(l -> dist >= t[l] ? 0 : 1).~sum~();
  	      table.add(dist, pi, pj, k, k == 0 || dist >= t[k - 1] ? 0 : 1, k <= 1 ? 0 : ~sum~);
	    }
	  }
        }
        set.add(dist);
      }
    return table;
  }
  
 public void model() {
   int n = routes.length, nHards = hards == null ? 0 : hards.length, nSofts = softs.length;

   Var[] f = array("f", size(n), i -> dom(domains.get(routes[i].domain)),
     "f[i] is the frequency of the ith radio-link");
   Var[] p = array("p", size(n), i -> dom(routes[i].polarizationValues())),
     "p[i] is the polarization of the ith radio-link");
   Var[][] d = array("d", size(n, n), (i, j) -> dom(distances(i, j)).when(i < j && softLink(i,j)),
     "d[i][j] is the distance between the ith and the jth frequencies, for i < j when a soft link exists");
   Var[] v1 = array("v1", size(nSofts), dom(0, 1),
     "v1[q] is 1 iff the qth pair of radio constraints is violated when relaxing another level");
   Var[] v2 = array("v2", size(nSofts), dom(range(11)),
     "v2[q] is the number of times the qth pair of radio constraints is violated when relaxing more than one level");
   Var k = ~\textcolor{dred}{var}~("k", dom(range(12)),
     "k is the relaxation level to be optimized");

   forall(range(n).range(n), (i, j) -> {
     if (i < j && softLink(i,j))
       equal(d[i][j], dist(f[i], f[j]));
   }).note("computing intermediary distances");

   forall(range(nHards), q -> imperativeConstraint(f, p, hards[q])))
     .note("imperative constraints");    
   
   forall(range(nSofts), q -> {
     int i = softs[q].route1, j = softs[q].route2;
     extension(vars(i < j ? d[i][j] : d[j][i], p[i], p[j], k, v1[q], v2[q]), relaxTable(softs[q]));
   }).note("relaxable radioelectric compatibility constraints");

   int[] coeffs = vals(10 * nSofts * nSofts, repeat(10 * nSofts, nSofts), repeat(1, nSofts));
   minimize(~SUM~, vars(k, v1, v2), weightedBy(coeffs))
     .note("minimizing sophisticated relaxation cost");
 }
}
\end{mcsp}

%Progressive relaxation produces eleven CSP instances from any single original FAPP optimization instance.

This model involves 5 arrays of variables (as well as the stand-alone variable $k$) and two types of constraints: \gb{extension} and \gb{intension}.
A cache could be used for avoiding creating similar tables, and also for avoiding checking several times whether a given pair $(i,j$) is subject to a soft link. 
Two series 'm2s' and 'ext' of respectively 18 and 10 instances have been selected.
The series 'm2s' corresponds to the model described above whereas the series 'ext' (obtained after some reformulations) is compatible with the restrictions imposed for the mini-track.

\section{FRB}

This problem has been already used in previous XCSP competitions.
Hence, we very succinctly introduce it.

\subsection*{Description}
\begin{quote}
These instances are randomly generated using Model RB \cite{XBHL_random}, while guaranteeing satisfiability.
\end{quote}

This satisfaction problem only involves (ordinary) table constraints.
A series of 16 instances has been selected.
Using model RB, some forced binary CSP instances have been generated by choosing $k=2$, $\alpha=0.8$,  $r=0.8$ and $n$ varying from $40$ to $59$.
Each such instance is prefixed by \gb{frb-n}.

\section{Golomb Ruler}

This is Problem \href{http://www.csplib.org/Problems/prob006}{006} on CSPLib, called Golomb Ruler.

\subsection*{Description {\small (from Peter van Beek on CSPLib)}}
\begin{quote}
``The problem is to find the ruler with the smallest length where we can put $n$ marks such that the distance between any two pairs of marks is distinct.'' 
\end{quote}

\subsection*{Data}

As an illustration of data specifying an instance of this problem, we simply have $n=8$.

\subsection*{Model}

The MCSP3 model used for the competition is: 

\begin{mcsp}
class GolombRuler implements ProblemAPI {
  int n;
  
  public void model() {
    int rulerLength = n * n + 1; // a trivial upper-bound
    
    Var[] x = array("x", size(n), dom(range(rulerLength)),
      "x[i] is the position of the ith tick");
    Var[][] y = array("y", size(n, n), (i, j) -> dom(range(1, rulerLength)).when(i < j),
      "y[i][j] is the distance between x[i] and x[j], for i < j");
      
    allDifferent(y)
      .note("all distances are different");
    forall(range(n), i -> forall(range(i + 1, n), j -> equal(x[j], add(x[i], y[i][j]))))
      .note("computing distances");
      
    minimize(x[n - 1])
      .note("minimizing the position of the rightmost tick");

    decisionVariables(x);
  }
}
\end{mcsp}

This model involves 2 arrays of variables and 2 types of constraints: \gb{allDifferent} and \gb{intension} (\gb{equal}).
A series of $10+3$ instances has been chosen ($n$ varying from 7 to 16).
Decision variables are indicated, except for the 3 instances (\xt files) whose name contains 'nodv' (no decision variables).

\section{Graceful Graph}

This is Problem \href{http://www.csplib.org/Problems/prob053}{053} on CSPLib, called 
Graceful Graph. See, for example, \cite{SP_constraint}.

\subsection*{Description {\small (from Karen Petrie on CSPLib)}}
\begin{quote}
``A labelling $f$ of the nodes of a graph with $q$ edges is graceful if $f$ assigns each node a
unique label from $\{0, 1, \dots q\}$ and when each edge $(x,y)$ is labelled with $|f(x) - f(y)|$,
the edge labels are all different. (Hence, the edge labels are a permutation of $1, 2, \dots, q$.)''
\end{quote}

We focused on graphs of the form $K_k \times P_p$ that consist of $p$ copies of a clique K of size $k$ with corresponding nodes of the cliques also forming the nodes of a path of length $p$.

\subsection*{Data}

As an illustration of data specifying an instance of this problem, we have $(k=5, p=2)$.

\subsection*{Model}

The MCSP3 model used for the competition is: 

\begin{mcsp}
class GracefulGraph implements ProblemAPI {
  int k; // size of each clique K (number of nodes)
  int p; // size of each path P (or equivalently, number of cliques)
  
  public void model() {
    int nEdges = ((k * (k - 1)) * p) / 2 + k * (p - 1);
    
    Var[][] cn = array("cn", size(p, k), dom(range(nEdges + 1)),
      "cn[i][j] is the color of the jth node of the ith clique");
    Var[][][] ce = array("ce", size(p, k, k), (i, j1, j2) -> dom(range(1, nEdges + 1)).when(j1 < j2),
      "ce[i][j1][j2] is the color of the edge (j1, j2) of the ith clique, for j1 < j2");
    Var[][] cp = array("cp", size(p - 1, k), dom(range(1, nEdges + 1)),
      "cp[i][j] is the color of the jth edge of the ith path");
    
    allDifferent(cn)
      .note("all nodes are colored differently");
    allDifferent(vars(ce, cp))
      .note("all edges are colored differently");
    
    block(() -> {
      forall(range(p).range(k), (i, j1) -> forall(range(j1 + 1, k), j2 ->
        equal(ce[i][j1][j2], dist(cn[i][j1], cn[i][j2]))));
      forall(range(p - 1).range(k), (i, j) -> equal(cp[i][j], dist(cn[i][j], cn[i + 1][j])));
    }).note("computing colors of edges from colors of nodes");
  }
}
\end{mcsp}

This model involves 3 arrays of variables and 2 types of constraints: \gb{allDifferent} and \gb{intension} (\gb{equal}).
A series of 11 instances has been selected.

\section{Graph Coloring}

This well-known problem has been already used in previous XCSP competitions.
%Hence, we succinctly introduce it.

\subsection*{Description}
\begin{quote}
``Given a graph $G=(V,E)$, the objective is to find the minimum number of colors such that it is possible to color each node of $G$ while ensuring that no two adjacent nodes share the same color.''
\end{quote}

\subsection*{Model}

The MCSP3 model used for the competition is: 

\begin{mcsp}
class Coloring implements ProblemAPI {
  int nNodes, nColors;
  int[][] edges;
  
  public void model() {
    int nEdges = edges.length;

    Var[] x = array("x", size(nNodes), dom(range(nColors)),
      "x[i] is the color assigned to the ith node of the graph");
    
    forall(range(nEdges), i -> different(x[edges[i][0]], x[edges[i][1]]))
      .note("all adjacent nodes must be colored differently");

    minimize(~MAXIMUM~, x)
      .note("minimizing the maximum used color index (and, consequently, the number of colors)");
  }
}
\end{mcsp}

This model only involves 1 array of variables and 1 type of constraint: \gb{intension} (\gb{different}).
A series of 11 instances has been selected for the competition.

\section{Haystacks}

This problem, introduced by Marc Van Dongen, has been already used in previous XCSP competitions.

\subsection*{Description  {\small (from Marc Van Dongen)}}
\begin{quote}
``The problem instance of order $p$ has $p \times p$ variables with domain $\{0,\dots,p-1\}$. 
The constraint graph is highly regular, consisting of $p$ clusters: one central cluster and $p-1$ outer clusters, each one being a $p$-clique. 
The instances are designed so that if the variables in the central cluster are instantiated, only one of the outer clusters contains an inconsistency: this cluster is the haystack. 
The task is to find the haystack and decide that it is unsatisfiable, thereby providing a proof that the current instantiation of the variables in the central cluster is inconsistent.''
\end{quote}

A series of 10 instances has been selected for the competition.

\section{Knapsack}

This is Problem \href{http://www.csplib.org/Problems/prob133}{133} on CSPLib, called %proposed by Özgür Akgün, and called 
Knapsack. %See, for example, \cite{SP_constraint}.

\subsection*{Description}
\begin{quote}
``Given a set of items, each with a weight and a value, determine which items to include in a collection so that the total weight is less than or equal to a given capacity and the total value is as large as possible.''
\end{quote}

\subsection*{Data}

As an illustration of data specifying an instance of this problem, we have:
\begin{json}
{
  "capacity": 10,
  "items": [
    { "weight": 2, "value": 54 },
    { "weight": 2, "value": 92 },
    { "weight": 1, "value": 62 },
    { "weight": 2,"value": 20 },
    { "weight": 2,"value": 55 }
  ]
}
\end{json}

\subsection*{Model}

The MCSP3 model used for the competition is: 

\begin{mcsp}
class Knapsack implements ProblemAPI {
  int capacity;
  Item[] items;
  
  class Item {
    int weight;
    int value;
  }
  
  public void model() {
    int[] weights = valuesFrom(items, item -> item.weight);
    int[] values = valuesFrom(items, item -> item.value);
    int nItems = items.length;
    
    Var[] x = array("x", size(nItems), dom(0, 1),
      "x[i] is 1 iff the ith item is selected");
    
    sum(x, weightedBy(weights), LE, capacity)
      .note("the capacity of the knapsack must not be exceeded");

    maximize(~SUM~, x, weightedBy(values))
      .note("maximizing summed up value (benefit)");
  }
}
\end{mcsp}

This model only involves 1 array of variables and 1 type of constraint: \gb{sum}.
A series of 14 instances has been selected for the competition.

\section{Langford}

This is Problem \href{http://www.csplib.org/Problems/prob024}{024} on CSPLib, called Langdford's number problem.

\subsection*{Description {\small (from Toby Walsh on CSPLib)}}
\begin{quote}
``Given two integers $k$ and $n$, the problem $L(k,n)$ is to arrange $k$ sets of numbers $1$ to $n$, so that each appearance of the number $m$ is $m$ numbers on from the last.''
\end{quote}

\subsection*{Data}
Here, we focus on the model proposed in \cite{GJM_watched} for $k=2$.
The MCSP3 model used for the competition is:

\begin{mcsp}
class LangfordBin implements ProblemAPI {
  int n;
  
  public void model() {
    Var[] v = array("v", size(2 * n), dom(range(1, n + 1)),
      "v[i] is the ith value of the Langford vector");
    Var[] p = array("p", size(2 * n), dom(range(2 * n)),
      "p[j] is the first (resp., second) position of 1+j/2 in v if j is even (resp., odd)");
    
    forall(range(n), i -> element(v, at(p[2 * i]), takingValue(i + 1)))
      .note("computing the position of the 1st occurrence of i");   
    forall(range(n), i -> element(v, at(p[2 * i + 1]), takingValue(i + 1)))
      .note("computing the position of the 2nd occurrence of i"); 
    forall(range(n), i -> equal(p[2 * i], add(i + 2, p[2 * i + 1])))
      .note("the distance between two occurrences of i must be respected");
  }
}
\end{mcsp}

This model involves 2 arrays of variables and 2 types of constraints: \gb{element} and \gb{intension} (\gb{equal}).
A series of 11 instances has been generated for the competition, by varying $n$ from 6 to 16.

\section{Low Autocorrelation}

This is Problem \href{http://www.csplib.org/Problems/prob005}{005} on CSPLib, called Low Autocorrelation Binary Sequences.

\subsection*{Description {\small (from Toby Walsh on CSPLib)}}
\begin{quote}
``The objective is to construct a binary sequence $S_i$ of length $n$ that minimizes the autocorrelations between bits.
  Each bit in the sequence takes the value $+1$ or $-1$.
  With non-periodic (or open) boundary conditions, the kth autocorrelation, $C_k$ is defined to be $\sum_{i=0}^{n-k-1} S_i \times S_{i+k}$.
%  With periodic (or cyclic) boundary conditions, the kth autocorrelation, $C_k$ is defined to be $\sum_{i=0}^{n-1} S_i \times S_{i+ k\%n}$.
  The aim is to minimize the sum of the squares of these autocorrelations, i.e., to minimize $E=\sum_{k=1}^{n-1}C_k^2$.
\end{quote}

\subsection*{Data}

As an illustration of data specifying an instance of this problem, we have $n=10$.

\subsection*{Model}

The MCSP3 model used for the competition is: 

\begin{mcsp}
class LowAutocorrelation implements ProblemAPI {
  int n;

  public void model() {
    Var[] x = array("x", size(n), dom(-1, 1),
      "x[i] is the ith value of the sequence to be built.");
    Var[][] y = array("y", size(n - 1, n - 1), (k, i) -> dom(-1, 1).when(i < n - k - 1),
      "y[k][i] is the ith product value required to compute the kth autocorrelation");
    Var[] c = array("c", size(n - 1), k -> dom(range(-n + k + 1, n - k)),
      "c[k] is the value of the kth autocorrelation");
    Var[] s = array("s", size(n - 1), k -> dom(range(n - k).map(v -> v * v)),
      "s[k] is the square of the kth autocorrelation");
    
    forall(range(n - 1), k -> forall(range(n - k - 1), i -> equal(y[k][i], mul(x[i], x[i + k + 1]))))
      .note("computing product values");
    forall(range(n - 1), k -> sum(y[k], EQ, c[k]))
      .note("computing the values of the autocorrelations");
    forall(range(n - 1), k -> equal(s[k], mul(c[k], c[k])))
      .note("computing the squares of the autocorrelations");
    
    minimize(~SUM~, s)
      .note("minimizing the sum of the squares of the autocorrelation");
  }
}
\end{mcsp}

This model involves 4 arrays of variables and 2 types of constraints: \gb{sum} and \gb{intension} (\gb{equal}).
A series of 14 instances has been generated for the competition.

\section{Magic Hexagon}

This is Problem \href{http://www.csplib.org/Problems/prob023}{023} on CSPLib, called  Magic Hexagon.

\subsection*{Description}
\begin{quote}
``A magic hexagon consists of the numbers 1 to 19 arranged in a hexagonal pattern such that all diagonals sum to 38.''
\end{quote}

The description is given here for order $n=3$ (the length of the first row of the hexagon) and starting value $s=1$ (the first value of the sequence of numbers).

\subsection*{Model}

The MCSP3 model used for the competition is: 

\begin{mcsp}
class MagicHexagon implements ProblemAPI {
  int n; // order
  int s; // start

  private Var[] scopeForDiagonal(Var[][] x, int i, boolean right) {
    int d = x.length;
    int v1 = right ? Math.max(0, d / 2 - i) : Math.max(0, i - d / 2), v2 = d / 2 - v1;
    Range r = range(d - Math.abs(d / 2 - i));
    return variablesFrom(r, j -> x[j + v1][i - Math.max(0, right ? v2 - j : j - v2)]);
  }
  
  public void model() {
    int gap = 3 * n * n - 3 * n + 1;
    int magic = sumOf(range(s, s + gap)) / (2 * n - 1);
    int d = n + n - 1; // longest diameter
    
    Var[][] x = array("x", size(d, d), (i, j) -> dom(range(s, s + gap)).when(j < d - Math.abs(d/2 - i)),
      "x represents the hexagon; on row x[i], only the first n - |n/2 - i| cells are useful.");
   
    allDifferent(x)
      .note("all values must be different");;
    forall(range(d), i -> sum(x[i], EQ, magic))
      .note("all rows sum to the magic value");
    forall(range(d), i -> sum(scopeForDiagonal(x, i, true), EQ, magic))
      .note("all right-sloping diagonals sum to the magic value");
    forall(range(d), i -> sum(scopeForDiagonal(x, i, false), EQ, magic))
      .note("all left-sloping diagonals sum to the magic value");
    
    block(() -> {
      lessThan(x[0][0], x[0][n - 1]);
      lessThan(x[0][0], x[n - 1][d - 1]);
      lessThan(x[0][0], x[d - 1][n - 1]);
      lessThan(x[0][0], x[d - 1][0]);
      lessThan(x[0][0], x[n - 1][0]);
      lessThan(x[0][n - 1], x[n - 1][0]);
    }).tag(SYMMETRY_BREAKING);
  }
}
\end{mcsp}

This model involves 1 array of variables and 3 types of constraints: \gb{allDifferent}, \gb{sum} and \gb{intension} (\gb{lessThan}).
The intensional constraints are here for  breaking a few symmetries.
A series of 11 instances has been generated for the competition.

\section{Magic Square}

This is Problem \href{http://www.csplib.org/Problems/prob019}{019} on CSPLib, called Magic Square.

\subsection*{Description}
\begin{quote}
``A magic square of order $n$ is a $n$ by $n$ matrix containing the numbers 1 to $n^2$, where each row, column and main diagonal sum up to the same value.''
\end{quote}

\subsection*{Data}

As an illustration of data specifying an instance of this problem, we have:
\begin{json}
{
  "n": 9,
  "clues": [
    [0, 0, 0, 31, 0, 0, 0, 0, 0],
    [0, 0, 0, 0, 0, 0, 0, 12, 0],
    ...
  ]
}
\end{json}

%In some cases, clues (pre-set values) are given.
\noindent When there is no clue (pre-set value) at all, we have:
\begin{json}
{
  "n": 10,
  "clues": null
}
\end{json}

or, equivalently:
\begin{json}
{
  "n": 10,
  "clues": "null"
}
\end{json}

\subsection*{Model}

The MCSP3 model used for the competition is: 

\begin{mcsp}
class MagicSquare implements ProblemAPI {
  int n;
  int[][] clues;
  
  public void model() {
    int magic = n * (n * n + 1) / 2;
    
    Var[][] x = array("x", size(n, n), dom(range(1, n * n + 1)),
      "x[i][j] is the value at row i and column j of the magic square");
    
    allDifferent(x)
      .note("all values must be different");

    forall(range(n), i -> sum(x[i], EQ, magic))
      .note("all rows sum up to the magic value");
    forall(range(n), j -> sum(columnOf(x, j), EQ, magic))
      .note("all columns sum up to the magic value");

    block(() -> {
      sum(diagonalDown(x), EQ, magic);
      sum(diagonalUp(x), EQ, magic);
    }).note("the two (main) diagonals sum up to the magic value");

    instantiation(x, takingValues(clues), onlyOn((i, j) -> clues[i][j] != 0)).tag(CLUES)
      .note("respecting specified clues (if any)");
  }
}
\end{mcsp}

This model involves 1 array of variables and 3 types of constraints: \gb{allDifferent}, \gb{sum} and \gb{instantiation}.
A series of 13 instances has been generated, by varying $n$ from 4 to 16.
We didn't use any clues (and so, the constraint  \gb{instantiation} is simply discarded at compilation).

\section{Mario}

This is a problem proposed by Amaury Ollagnier and Jean-Guillaume Fages at the 2013 Minizinc Challenge.

\subsection*{Description {\small (from Amaury Ollagnier and Jean-Guillaume Fages)}}
\begin{quote}
``This models a routing problem based on a little example of Mario's day. 
 Mario is an Italian Plumber and his work is mainly to find gold in the plumbing of all the houses of the neighborhood. 
 Mario is moving in the city using his kart that has a specified amount of fuel. Mario starts his day of work from his house 
 and always ends to his friend Luigi's house to have the supper. The problem here is to plan the best path for
 Mario in order to earn the more money with the amount of fuel of his kart ! 
% From a more general point of view, the problem is to find a path in a graph:
% - Path endpoints are given (from Mario's to Luigi's)
% - The sum of weights associated to arcs in the path is restricted (fuel consumption)
% - The sum of weights associated to nodes in the path has to be maximized (gold coins)
\end{quote}

\subsection*{Data}

As an illustration of data specifying an instance of this problem, we have:
\begin{json}
{
  "marioHouse": 0,
  "luigiHouse": 1,
  "fuelLimit": 2000,
  "houses": [
    {
      "fuelConsumption": [0, 221, 274, 808, 13, 677, 670, 943, 969, 13, 18, 217],
      "gold": 0
    },
    {
      "fuelConsumption": [0, 0, 702, 83, 813, 679, 906, 335, 529, 719, 528, 451],
      "gold": 10
    },
    ...
  ]
}
\end{json}

\subsection*{Model}

The MCSP3 model used for the competition is: 

\begin{mcsp}
class Mario implements ProblemAPI {
  int marioHouse, luigiHouse;
  int fuelLimit;
  House[] houses;

  class House {
    int[] fuel;
    int gold;
  }

  public void model() {
    int nHouses = houses.length;

    Var[] s = array("s", size(nHouses), dom(range(nHouses)),
      "s[i] is the house succeeding to the ith house (itself if not part of the route)");
    Var[] f = array("f", size(nHouses), i -> dom(houses[i].fuel),
      "f[i] is the fuel consumed at each step (from house i to its successor)");
    Var[] g = array("g", size(nHouses), i -> dom(0, houses[i].gold),
      "g[i] is the gold earned at house i");
    
    forall(range(nHouses), i -> extension(vars(s[i], f[i]), indexing(houses[i].fuel)))
      .note("fuel consumption at each step");

    sum(f, LE, fuelLimit)
      .note("we cannot consume more than the available fuel");
    
    forall(range(nHouses), i -> {
      if (i != marioHouse && i != luigiHouse)
        equivalence(eq(s[i], i), eq(g[i], 0));
    }).note("gold earned at each house");
    
    circuit(s)
      .note("Mario must make a complete tour");

    equal(s[luigiHouse], marioHouse)
      .note("Mario house is just after Luigi house");
    
    maximize(~SUM~, g)
      .note("maximizing collected gold");
  }
}
\end{mcsp}

This model involves 3 arrays of variables and 4 types of constraints: \gb{circuit}, \gb{sum}, \gb{extension} and \gb{intension} (\gb{equivalence} and \gb{equal}).
A series of 10 instances has been selected for the competition.

\section{Mistery Shopper}

This is Problem \href{http://www.csplib.org/Problems/prob004}{004} on CSPLib, called Mistery Shopper.

\subsection*{Description {\small (from Jim Ho Man Lee on CSPLib)}}
\begin{quote}
  ``A well-known cosmetic company wants to evaluate the performance of their sales people, who are stationed at the company’s counters at various department stores in different geographical locations.
  For this purpose, the company has hired some secret agents to disguise themselves as shoppers to visit the sales people.
  The visits must be scheduled in such a way that each sales person must be visited by shoppers of different varieties and that the visits should be spaced out roughly evenly.
  Also, shoppers should visit sales people in different geographic locations.''
\end{quote}

More details can be found on CSPLib.

\subsection*{Data}

As an illustration of data specifying an instance of this problem, we have:
\begin{json}
{
  "visitorGroups" : [4, 4, 4],
  "visiteeGroups" : [3, 2, 4]
}
\end{json}

\subsection*{Model}

The MCSP3 model used for the competition is: 

\begin{mcsp}
class MisteryShopper implements ProblemAPI {
  int[] visitorGroups; // visitorGroups[i] gives the size of the ith visitor group
  int[] visiteeGroups; // visiteeGroups[i] gives the size of the ith visitee group
  
  private Table numberPer(int[] t) { // numbering persons over all groups (sizes) of t
    Table table = table();
    for (int cnt = 0, i = 0; i < t.length; i++)
      for (int j = 0; j < t[i]; j++)
        table.add(i, cnt++);
    return table;
  }
  
  public void model() {
    int nVisitors = sumOf(visitorGroups), nVisitees = sumOf(visiteeGroups);
    int n = nVisitors, nDummyVisitees = nVisitors - nVisitees;
    if (nDummyVisitees > 0)
      visiteeGroups = addInt(visiteeGroups, nDummyVisitees); // dummy group added
    int nVisitorGroups = visitorGroups.length, nVisiteegroups = visiteeGroups.length;
    int nWeeks = nVisitorGroups;
      
    Var[][] vr = array("vr", size(n, nWeeks), dom(range(n)),
      "vr[i][w] is the visitor for the ith visitee at week w");
    Var[][] ve = array("ve", size(n, nWeeks), dom(range(n)),
      "ve[i][w] is the visitee for the ith visitor at week w");
    Var[][] gvr = array("gvr", size(n, nWeeks), dom(range(nVisitorGroups)),
      "gvr[i][w] is the visitor group for the ith visitee at week w");
    Var[][] gve = array("gve", size(n, nWeeks), dom(range(nVisiteeGroups)),
      "gve[i][w] is the visitee group for the ith visitor at week w");
    
    forall(range(nWeeks), w -> allDifferent(columnOf(vr, w)))
      .note("each week, all visitors must be different");
    forall(range(nWeeks), w -> allDifferent(columnOf(ve, w)))
      .note("each week, all visitees must be different");
    forall(range(n), i -> allDifferent(gvr[i]))
      .note("the visitor groups must be different for each visitee");
    forall(range(n), i -> allDifferent(gve[i]))
      .note("the visitee groups must be different for each visitor");
    
    forall(range(nWeeks), w -> channel(columnOf(vr, w), columnOf(ve, w)))
      .note("channeling arrays vr and ve, each week");    
   
    forall(range(n).range(nWeeks), (i,w) -> extension(vars(gvr[i][w],vr[i][w]), numberPer(visitorGroups)))
      .note("linking a visitor with its group");
    forall(range(n).range(nWeeks), (i,w) -> extension(vars(gve[i][w],ve[i][w]), numberPer(visiteeGroups)))
      .note("linking a visitee with its group");

    block(() -> {
      lexMatrix(vr, ~INCREASING~);
      if (nDummyVisitees > 0)
        forall(range(nWeeks), w -> strictlyIncreasing(select(columnOf(vr, w), range(nVisitees, n))));
    }).tag(SYMMETRY_BREAKING);
  }
}
\end{mcsp}

This model involves 4 arrays of variables and 5 types of constraints: \gb{channel}, \gb{allDifferent}, \gb{lexMatrix}, \gb{extension} and \gb{intension} (\gb{strictlyIncreasing}).
Note that we could have stored and reused tables instead of systematically building them.
There is a block for breaking some symmetries. 
A series of 10 instances has been generated for the competition.

\section{Nurse Rostering}

This is a realistic employee shift scheduling Problem (see, for example, \cite{MBL_solving}).
%The reader A set of 24 instances are detailed at  \href{http://www.schedulingbenchmarks.org/instances1\_24.html}{http://www.schedulingbenchmarks.org/instances1\_24.html}.

\subsection*{Description}
\begin{quote}
The description is rather complex. Hence, we refer the reader to:\\ \href{http://www.schedulingbenchmarks.org/instances1\_24.html}{http://www.schedulingbenchmarks.org/instances1\_24.html}. 
\end{quote}

\subsection*{Data}

As an illustration of data specifying an instance of this problem, we have:

\begin{json}
{
  "nDays": 14,
  "shifts": [ { "id": "D", "length": 480, "forbiddenFollowingShifts": "null" } ],
  "staffs": [
    { "id": "A",
      "maxShifts": [14],
      "minTotalMinutes": 3360,"maxTotalMinutes": 4320,
      "minConsecutiveShifts": 2,"maxConsecutiveShifts": 5,
      "minConsecutiveDaysOff": 2,
      "maxWeekends": 1, "daysOff": [0],
      "onRequests": [
        { "day": 2, "shift": "D", "weight": 2 },
        { "day": 3, "shift": "D", "weight": 2}
      ],
      "offRequests": "null"
    },
    ...
  ],
  "covers": [
     [ { "requirement": 3, "weightIfUnder": 100, "weightIfOver": 1 } ],
     [ { "requirement": 5, "weightIfUnder": 100, "weightIfOver": 1 } ],
     ...
  ]
}
\end{json}

\subsection*{Model}

The MCSP3 model used for the competition is: 

\begin{mcsp}
class NurseRostering implements ProblemAPI {
  int nDays;
  Shift[] shifts;
  Staff[] staffs;
  Cover[][] covers;
  
  class Shift {
    String id = "_off"; // value for the dummy shift
    int length;
    String[] forbiddenFollowingShifts;
  }
  
  class Request {
    int day;
    String shift;
    int weight;
  }
  
  class Staff {
    String id;
    int[] maxShifts;
    int minTotalMinutes, maxTotalMinutes;
    int minConsecutiveShifts, maxConsecutiveShifts;
    int minConsecutiveDaysOff, maxWeekends;
    int[] daysOff;
    Request[] onRequests, offRequests;
  }
  
  class Cover {
    int requirement, weightIfUnder, weightIfOver;

    int costFor(int i) {
      return i <= requirement ? (requirement - i) * weightIfUnder : (i - requirement) * weightIfOver;
    }
  }

  private Request onRequest(int person, int day) {
    return firstFrom(staffs[person].onRequests, request -> request.day == day);
  }

  private Request offRequest(int person, int day) {
    return firstFrom(staffs[person].offRequests, request -> request.day == day);
  }

  private int shiftPos(String s) {
    return firstFrom(range(shifts.length), i -> shifts[i].id.equals(s));
  }

  private int[] costsFor(int day, int shift) {
    int[] t = new int[staffs.length + 1];
    if (shift != shifts.length - 1) // if not '\_off'
      for (int i = 0; i < t.length; i++)
        t[i] = covers[day][shift].costFor(i);
    return t;
  }

  private Automaton automatonMinConsec(int nShifts, int k, boolean forShifts) {
    Range rangeOff = range(nShifts - 1, nShifts); // a range with only one value (off)
    Range rangeNotOff = range(nShifts - 1); // a range with all other values
    Range r1 = forShifts ? rangeOff : rangeNotOff, r2 = forShifts ? rangeNotOff : rangeOff;
    Transitions transitions = transitions();
    transitions.add("q0", r1, "q1").add("q0", r2, "q" + (k + 1)).add("q1", r1, "q" + (k + 1));
    for (int i = 1; i <= k; i++)
      transitions.add("q" + i, r2,"q" + (i + 1));
    transitions.add("q" + (k + 1), range(nShifts), "q" + (k + 1));
    return automaton("q0", transitions, finalState("q" + (k + 1)));
  }

  private Table rotationTable() {
    Table table = table(NEGATIVE);  // a negative table (i.e., involving conflicts)
    for (Shift shift1 : shifts)
      if (shift1.forbiddenFollowingShifts != null)
        for (String shift2 : shift1.forbiddenFollowingShifts)
          table.add(shiftPos(shift1.id), shiftPos(shift2));
    return table;
  }

  private void buildDummyShift() {
    shifts = addObject(shifts, new Shift()); // we append first a dummy off shift
    for (Staff staff : staffs)
      staff.maxShifts = addInt(staff.maxShifts, nDays); // we append no limit (nDays) for the dummy shift
  }
  
  public void model() {
    buildDummyShift();
    int nWeeks = nDays / 7, nShifts = shifts.length, nStaffs = staffs.length;
    int off = nShifts - 1; // value for '\_off'
    
    Var[][] x = array("x", size(nDays, nStaffs), dom(range(nShifts)),
      "x[d][p] is the shift at day d for person p (one shift denoting '_off')");
    Var[][] ps = array("ps", size(nStaffs, nShifts), (p, s) -> dom(range(staffs[p].maxShifts[s] + 1)),
      "ps[p][s] is the number of days such that person p works with shift s");
    Var[][] ds = array("ds", size(nDays, nShifts), dom(range(nStaffs + 1)),
      "ds[d][s] is the number of persons working on day d with shift s");
    Var[][] wk = array("wk", size(nStaffs, nWeeks), dom(0, 1),
      "wk[p][w] is 1 iff the week-end w is worked by person p");
    Var[][] cn = array("cn", size(nStaffs, nDays), (p, d) ->
        onRequest(p, d) != null ? dom(0, onRequest(p, d).weight) : null,
      "cn[p][d] is the cost of not satisfying the on-request (if it exists) of person p on day d");
    Var[][] cf = array("cf", size(nStaffs, nDays), (p, d) ->
        offRequest(p, d) != null ? dom(0, offRequest(p, d).weight) : null,
      "cf[p][d] is the cost of not satisfying the off-request (if it exists) of person p on day d");    
    Var[][] cc = array("cc", size(nDays, nShifts), (d, s) -> dom(costs(d, s)),
      "cc[d][s] is the cost of not satisfying cover for shift s on day d");
    
    instantiation(select(x, (d, p) -> contains(staffs[p].daysOff, d)), takingValue(off))
      .note("guaranteeing days off for staff");    
    forall(range(nStaffs).range(nShifts), (p, s) -> exactly(columnOf(x, p), takingValue(s), ps[p][s]))
      .note("computing number of days");
    forall(range(nDays).range(nShifts), (d, s) -> exactly(x[d], takingValue(s), ds[d][s]))
      .note("computing number of persons");
    
    forall(range(nStaffs).range(nWeeks), (p, w) -> {
      implication(ne(x[w * 7 + 5][p], off), eq(wk[p][w], 1));
      implication(ne(x[w * 7 + 6][p], off), eq(wk[p][w], 1));
    }).note("computing worked week-ends");
    
    if (rotationTable().size() > 0)
      forall(range(nStaffs), p -> slide(columnOf(x, p), range(nDays - 1), i ->
        extension(vars(x[i][p], x[i + 1][p]), rotationTable())))
      .note("rotation shifts");

    forall(range(nStaffs), p -> sum(wk[p], LE, staffs[p].maxWeekends))
      .note("maximum number of worked week-ends");
    
    int[] lengths = valuesFrom(shifts, shift -> shift.length);
    forall(range(nStaffs), p ->
      sum(ps[p], weightedBy(lengths), IN, range(staffs[p].minTotalMinutes,staffs[p].maxTotalMinutes + 1)))
    .note("minimum and maximum number of total worked minutes");
    
    forall(range(nStaffs), p -> {
      int k = staffs[p].maxConsecutiveShifts;
      forall(range(nDays - k), i ->
        atLeast1(select(columnOf(x, p), range(i,  i + k + 1)), takingValue(off)));
    }).note("maximum consecutive worked shifts");
    
    forall(range(nStaffs), p -> {
      int k = staffs[p].minConsecutiveShifts;
      forall(range(nDays - k), i ->
        regular(select(columnOf(x, p), range(i,  i + k + 1)), automatonMinConsec(nShifts, k, true)));
    }).note("minimum consecutive worked shifts");
    
    forall(range(nStaffs), p -> {
      int k = staffs[p].minConsecutiveDaysOff;
      forall(range(nDays - k), i ->
        regular(select(columnOf(x, p), range(i, i + k + 1)), automatonMinConsec(nShifts, k, false)));
    }).note("minimum consecutive days off");

   forall(range(nStaffs), p -> {
      int k = staffs[p].minConsecutiveShifts;
      if (k > 1) {
	forall(range(1, k), i -> implication(ne(x[0][p], off), ne(x[i][p], off)));
	forall(range(1, k), i -> implication(ne(x[nDays - 1][p], off), ne(x[nDays - 1 - i][p], off)));
      }
    }).note("managing off days on schedule ends");

   forall(range(nStaffs).range(nDays), (p, d) -> {
      if (onRequest(p, d) != null)
        equivalence(eq(x[d][p], shiftPos(onRequest(p, d).shift)), eq(cn[p][d], 0));
      if (offRequest(p, d) != null)
        equivalence(eq(x[d][p], shiftPos(offRequest(p, d).shift)), ne(cf[p][d], 0));
    }).note("cost of not satisfying on and off requests");

    forall(range(nDays).range(nShifts), (d,s) -> extension(vars(ds[d][s],cc[d][s]), indexing(costs(d,s))))
      .note("cost of under/over covering");
    
    minimize(~SUM~, vars(cn, cf, cc));
  }
}
\end{mcsp}

This model involves 7 arrays of variables and 7 types of constraints: \gb{regular}, \gb{slide}, \gb{count} (\gb{exactly} and \gb{atLeast}), \gb{sum}, \gb{instantiation}, \gb{intension} (\gb{implication} and \gb{equivalence}) and \gb{extension}.
Note how data are structured: we use 4 classes to describe them.
You can easily follow the structure of the automatas that are built when \mn{automatonMinConsec} is called.
% fields onRequests offRequests
A series of 20 instances has been selected from \href{http://www.schedulingbenchmarks.org/}{http://www.schedulingbenchmarks.org}.

\section{Peacable Armies}

This is Problem \href{http://www.csplib.org/Problems/prob110}{110} on CSPLib, called Peaceably Co-existing Armies of Queens.

\subsection*{Description {\small (from Ozgur Akgun on CSPLib)}}
\begin{quote}
``In the “Armies of queens” problem, we are required to place two equal-sized armies of black and white queens on a chessboard so that the white queens do not attack the black queens (and necessarily vice versa) and to find the maximum size of two such armies. Also see \cite{SPG_models}.''
\end{quote}

\subsection*{Data}

As an illustration of data specifying an instance of this problem, we have $n=10$.

\subsection*{Model}

The MCSP3 model(s) used for the competition is: 

\begin{mcsp}
class PeacableArmies implements ProblemAPI {
  int n; // order

  public void model() {
    if (modelVariant("m1")) {
      Var[][] b = array("b", size(n, n), dom(0, 1),
        "b[i][j] is 1 if a black queen is in the cell at row i and column j");
      Var[][] w = array("w", size(n, n), dom(0, 1),
        "w[i][j] is 1 if a white queen is in the cell at row i and column j");
      
      forall(range(n).range(n).range(n).range(n), (i1, j1, i2, j2) -> {
	if (i1 == i2 && j1 == j2)
	  lessEqual(add(b[i1][j1], w[i1][j1]), 1);
	else if (i1 < i2 || (i1 == i2 && j1 < j2))
	  if (i1 == i2 || j1 == j2 || Math.abs(i1 - i2) == Math.abs(j1 - j2)) {
	    lessEqual(add(b[i1][j1], w[i2][j2]), 1);
	    lessEqual(add(w[i1][j1], b[i2][j2]), 1);
	  }
      }).note("no two opponent queens can attack each other");

      int[] coeffs = range(n * n * 2).map(i -> i < n * n ? 1 : -1);
      sum(vars(b, w), weightedBy(coeffs), EQ, 0)
        .note("ensuring the same numbers of black and white queens");

      maximize(~SUM~, b)
        .note("maximizing the number of black queens (and consequently, the size of the armies)");
    }
    
    if (modelVariant("m2")) {
      Var[][] x = array("x", size(n, n), dom(0, 1, 2),
        "x[i][j] is 1 or 2 if a black or white queen is at row i and column j. It is 0 otherwise.");
      Var nb = ~\textcolor{dred}{var}~("nb", dom(range(n * n / 2)),
        "nb is the number of black queens");
      Var nw = ~\textcolor{dred}{var}~("nw", dom(range(n * n / 2)),
        "nw is the number of white queens");
      
      forall(range(n).range(n).range(n).range(n), (i1, j1, i2, j2) -> {
	if (i1 < i2 || (i1 == i2 && j1 < j2))
  	  if (i1 == i2 || j1 == j2 || Math.abs(i1 - i2) == Math.abs(j1 - j2))
	    different(add(x[i1][j1], x[i2][j2]), 3);
      }).note("No two opponent queens can attack each other");

      count(vars(x), takingValue(1), EQ, nb).note("counting the number of black queens");
      count(vars(x), takingValue(2), EQ, nw).note("counting the number of white queens");
      equal(nb, nw).note("ensuring equal-sized armies");

      maximize(nb)
        .note("maximizing the number of black queens (and consequently, the size of the armies)");
    }
  }
}
\end{mcsp}

Following \cite{SPG_models}, two model variants, called 'm1' and 'm2' have been written.
The first variant model involves 2 arrays of variables and 2 types of constraints: \gb{sum} and \gb{intension} (\gb{lessEqual}).
The second variant model involves 1 array of variables, 2 stand-alone variables and 2 types of constraints: \gb{count} and \gb{intension} (\gb{equal} and \gb{different}).
A series of $2\times7$ instances has been generated for the competition.

\section{Pizza Voucher}

This is a problem introduced in the Minizinc challenge 2015 under the name ``freepizza''.

\subsection*{Description}
\begin{quote}
  ``You are given a list of pizzas (actually, their prices) to get, and a set of vouchers. Each voucher can be used to get pizzas for free. For example a voucher 2/1 (2 being the 'pay' part and 1 the 'free' part) indicates that you need to buy 2 pizzas to get another one free (whose price must be inferior).
  You want to optimally use the vouchers so as to get all the pizzas with minimal cost.''
\end{quote}

\subsection*{Data}

As an illustration of data specifying an instance of this problem, we have:

\begin{json}
{
  "pizzaPrices": [50, 60, 90, 70, 80, 100, 20, 30, 40, 10],
  "vouchers": [
    { "payPart": 1, "freePart": 2 },
    { "payPart": 2, "freePart": 3 },
    ...
  ]
}
\end{json}

\subsection*{Model}

The MCSP3 model used for the competition is:

\begin{mcsp}
class PizzaVoucher implements ProblemAPI {
  int[] pizzaPrices;
  Voucher[] vouchers;
  
  class Voucher {
    int payPart;
    int freePart;
  }
  
  public void model() {
    int nPizzas = pizzaPrices.length, nVouchers = vouchers.length;
    
    Var[] v = array("v", size(nPizzas), dom(rangeClosed(-nVouchers, nVouchers)),
      "v[i] is the voucher used for getting the ith pizza. 0 means that no voucher is used. A negative
      (resp., positive) value i means that the ith pizza contributes to the the pay (resp., free) part of voucher |i|.");
    Var[] np = array("np", size(nVouchers), i -> dom(0, vouchers[i].payPart),
      "np[i] is the number of paid pizzas wrt the ith voucher");
    Var[] nf = array("nf", size(nVouchers), i -> dom(range(vouchers[i].freePart + 1)),
      "nf[i] is the number of free pizzas wrt the ith voucher");
    Var[] pp = array("pp", size(nPizzas), i -> dom(0, pizzaPrices[i]),
      "pp[i] is the price paid for the ith pizza");
    
    forall(range(nVouchers), i -> count(v, takingValue(-i - 1), EQ, np[i]))
      .note("counting paid pizzas");
    forall(range(nVouchers), i -> count(v, takingValue(i + 1), EQ, nf[i]))
      .note("counting free pizzas");
    forall(range(nVouchers), i -> equivalence(eq(nf[i], 0), ne(np[i], vouchers[i].payPart)))
      .note("a voucher, if used, must contribute to have at least one free pizza.");
    forall(range(nPizzas), i -> implication(le(v[i], 0), ne(pp[i], 0)))
      .note("a pizza must be paid iff a free voucher part is not used to have it free");
    
    forall(range(nPizzas).range(nPizzas), (i, j) -> {
      if (i != j && pizzaPrices[i] < pizzaPrices[j])
        disjunction(ge(v[i], v[j]), ne(v[i], neg(v[j])));
    }).note("a free pizza got with a voucher must be cheaper than any pizza paid wrt this voucher");
    
    minimize(~SUM~, pp)
      .note("minimizing summed up price paid for pizzas");

    decisionVariables(v);
  }
}
\end{mcsp}

This model involves 4 arrays of variables and 2 types of constraints: \gb{count} and \gb{intension} (\gb{equivalence}, \gb{implication} and \gb{disjunction}).
A series of 10 original instances has been generated for the competition.

\section{Pseudo-Boolean}

This problem has been already used in previous XCSP competitions.
%Hence, we very succinctly introduce it.

\subsection*{Description}
\begin{quote}
Pseudo-Boolean problems generalize SAT problems by allowing linear constraints and, possibly, a linear objective function.
\end{quote}

\subsection*{Data}

As an illustration of data specifying an instance of this problem, we have:

\begin{json}
{
  "n": 144,
  "e": 704,
  "ctrs": [
    { "coeffs": [1,1,1,1,1,1], "nums": [1,17,33,49,65,81], "op": "=", "limit": 1 },
    { "coeffs": [1,1,1,1,1,1], "nums": [2,18,34,50,66,82], "op": "=", "limit": 1 },
    ...
  ],
  "obj": {
    "coeffs": [1,2,1,1,1,2,2,1,1,2,1,1],
    "nums": [96,97,101,108,109,111,116,124,131,133,140,143]
  }
}
\end{json}

\subsection*{Model}

The MCSP3 model used for the competition is:

\begin{mcsp}
class PseudoBoolean implements ProblemAPI {
  int n, e;
  LinearCtr[] ctrs;
  LinearObj obj;
  
  class LinearCtr {
    int[] coeffs;
    int[] nums;
    String op;
    int limit;
  }
  
  class LinearObj {
    int[] coeffs;
    int[] nums;
  }
  
  public void model() {
    Var[] x = array("x", size(n), dom(0, 1), "x|i] is the Boolean value (0/1) of the ith variable");
    
    forall(range(e), i -> {
      Var[] scp = variablesFrom(ctrs[i].nums, num -> x[num]);
      sum(scp, weightedBy(ctrs[i].coeffs), TypeConditionOperatorRel.valueFor(ctrs[i].op), ctrs[i].limit)
        .note("respecting each linear constraint");
    });
    
    if (obj != null) {
      Var[] scp = variablesFrom(obj.nums, num -> x[num]);
      minimize(~SUM~, scp, weightedBy(obj.coeffs))
        .note("minimizing the linear objective");
    }
  }
}
\end{mcsp}

This problem involves 1 array of variables and 1 type of constraint: \gb{sum}.
Two series of 13 instances have been selected: one for CSP (model variant 'dec') and the other for COP (model 'opt').

\section{Quadratic Assignment}

The Quadratic Assignment Problem (QAP) is one of the fundamental combinatorial optimization problems in the branch of optimization. % or operations research in mathematics, from the category of the facilities location problems.
See, for example, \href{http://anjos.mgi.polymtl.ca/qaplib/}{QAPLIB}.

\subsection*{Description {\small (from WikiPedia)}}
\begin{quote}
``There are a set of $n$ facilities and a set of $n$ locations. For each pair of locations, a distance is specified and for each pair of facilities a weight or flow is specified (e.g., the amount of supplies transported between the two facilities).
  The problem is to assign all facilities to different locations with the goal of minimizing the sum of the distances multiplied by the corresponding flows.''
\end{quote}

\subsection*{Data}

As an illustration of data specifying an instance of this problem, we have:

\begin{json}
{
  "weights": [
    [0, 90, 10, 23, 43, 0, 0, 0, 0, 0, 0, 0],
    [90, 0, 0, 0, 0, 88, 0, 0, 0, 0, 0, 0],
    ...
  ],
  "distances": [
    [0, 36, 54, 26, 59, 72, 9, 34, 79, 17, 46, 95],
    [36, 0, 73, 35, 90, 58, 30, 78, 35, 44, 79, 36],
    ...
  ]
}
\end{json}

\subsection*{Model}

The MCSP3 model used for the competition is:

\begin{mcsp}
class QuadraticAssignment implements ProblemAPI {
  int[][] weights;   // facility weights
  int[][] distances; // location distances

  private Table channelingTable() {
    Table table = table();
    for (int i = 0; i < distances.length; i++)
      for (int j = 0; j < distances.length; j++)
        if (i != j)
          table.add(i, j, distances[i][j]);
    return table;
  }
  
  public void model() {
    int n = weights.length;
    
    Var[] x = array("x", size(n), dom(range(n)),
      "x[i] is the location assigned to the ith facility");
    Var[][] d = array("d", size(n, n), (i, j) -> dom(distances).when(i < j && weights[i][j] != 0),
      "d[i][j] is the distance between the locations assigned to the ith and jth facilities");
    
    allDifferent(x)
      .note("all locations must be different");
    
    forall(range(n).range(n), (i, j) -> {
      if (i < j && weights[i][j] > 0)
        extension(vars(x[i], x[j], d[i][j]), channelingTable());
    }).note("computing the distances");
    
    minimize(~SUM~, d, weightedBy(weights), onlyOn((i, j) -> i < j && weights[i][j] != 0))
      .note("minimizing summed up distances multiplied by flows");
  }
}
\end{mcsp}

This model involves 2 arrays of variables and 2 types of constraints: \gb{allDifferent} and \gb{extension}.
A series of 19 instances has been selected for the competition.

\section{Quasigroup}

This is Problem \href{http://www.csplib.org/Problems/prob003}{003} on CSPLib, called Quasigroup Existence.

\subsection*{Description {\small (from Toby Walsh on CSPLib)}}
\begin{quote}
  An order $n$ quasigroup is a Latin square of size $n$. That is, a $n \times n$ multiplication table in which each element occurs once in every row and column.
  A quasigroup can be specified by a set and a binary multiplication operator, $*$ defined over this set. Quasigroup existence problems determine the existence or non-existence of quasigroups of a given size with additional properties.
  For example:
  \begin{itemize}
  \item QG3: quasigroups for which $(a*b)*(b*a)=a$
  \item QG7: quasigroups for which $(b*a)*b=a*(b*a)$
  \end{itemize}
For each of these problems, we may additionally demand that the quasigroup is idempotent. That is, $a*a=a$ for every element $a$.
\end{quote}

\subsection*{Data}

As an illustration of data specifying an instance of this problem, we have $n=6$.

\subsection*{Model}

The MCSP3 model(s) used for the competition is: 

\begin{mcsp}
class QuasiGroup implements ProblemAPI {
  int n;
  
  public void model() {
    Var[][] x = array("x", size(n, n), dom(range(n)),
      "x[i][j] is the value at row i and column j of the quasigroup");
    
    allDifferentMatrix(x)
      .note("ensuring a Latin square"); 

    instantiation(diagonalDown(x), takingValues(range(n))).tag("idempotence");
      .note("enforcing x[i][i] = i");

    if (modelVariant("qg3")) {
      Var[][] y = array("y", size(n, n), dom(range(n * n)));
      
      forall(range(n).range(n), (i, j) -> {
	if (i != j) {
	  element(vars(x), at(y[i][j]), takingValue(i));
	  equal(y[i][j], add(mul(x[i][j], n), x[j][i]));
        }
      });
    }
    if (modelVariant("qg7")) {
      Var[][] y = array("y", size(n, n), dom(range(n)));
      
      forall(range(n).range(n), (i, j) -> {
	if (i != j) {
	  element(columnOf(x, j), at(x[j][i]), takingValue(y[i][j]));
	  element(x[i], at(x[j][i]), takingValue(y[i][j]));
        }
      });
    }
  }
}
\end{mcsp}

Two variants of the problem are described here.
Both involve 2 arrays of variables and 3 types of constraints: \gb{allDifferentMatrix}, \gb{instantiation} and \gb{element}.
The second variant, 'qg7', also involves another type of constraint: \gb{intension} (\gb{equal}).
Note the presence of the tag 'idempotence', which easily allows us to activate or deactivate the constraint \gb{instantiation}, at parsing time.
A series of $2\times8$ instances has been generated, for problems QG3 and QG7.

\section{RCPSP}

This is Problem \href{http://www.csplib.org/Problems/prob061}{061} on CSPLib, called 
Resource-Constrained Project Scheduling Problem (RCPSP). See also \href{http://www.om-db.wi.tum.de/psplib/}{PSPLIB}.

\subsection*{Description {\small (from Peter Nightingale and Emir Demirović on CSPLib)}}
\begin{quote}
``The resource-constrained project scheduling problem is a classical well-known problem in operations research.
  A number of activities are to be scheduled. Each activity has a duration and cannot be interrupted.
  There are a set of precedence relations between pairs of activities which state that the second activity must start after the first has finished.
%  The set of precedence relations are usually given as a directed acyclic graph (DAG), where an edge $(u,v)$ represents a precedence relation where $u$ must finish before $v$ begins.
%  The DAG contains two additional activities with duration 0, the source and sink, where the source is the first activity and sink is the last activity (these are dummy activities).
  There are a set of renewable resources.
  Each resource has a maximum capacity and at any given time slot no more than this amount can be in use.
  Each activity has a demand (possibly zero) on each resource.
  The dummy source and sink activities have zero demand on all resources.
The problem is usually stated as an optimisation problem where the makespan (i.e. the completion time of the sink activity) is minimized.''
\end{quote}

\subsection*{Data}

As an illustration of data specifying an instance of this problem, we have:

\begin{json}
{
  "horizon": 158,
  "resourceCapacities": [12, 13, 4, 12],
  "jobs": [
    { "duration": 0, "successors": [1, 2, 3], "requiredQuantities": [0, 0, 0, 0] },
    { "duration": 8, "successors": [5, 10, 14], "requiredQuantities": [4, 0, 0, 0] },
    ...,
    { "duration": 0, "successors": [], "requiredQuantities": [0, 0, 0, 0] }
  ]
}
\end{json}

\subsection*{Model}

The MCSP3 model used for the competition is: 

\begin{mcsp}
class Rcpsp implements ProblemAPI {
  int horizon;
  int[] resourceCapacities;
  Job[] jobs;
  
  class Job {
    int duration;
    int[] successors;
    int[] requiredQuantities;
  }
  
  public void model() {
    int nJobs = jobs.length;
    
    Var[] s = array("s", size(nJobs), i -> i == 0 ? dom(0) : dom(range(horizon)),
      "s[i] is the starting time of the ith job");
    
    forall(range(nJobs).range(nJobs), (i, j) -> {
      if (j < jobs[i].successors.length)
        lessEqual(add(s[i], jobs[i].duration), s[jobs[i].successors[j]]);
    }).note("precedence constraints");
    
    forall(range(resourceCapacities.length), j -> {
      int[] indexes = select(range(nJobs), i -> jobs[i].requiredQuantities[j] > 0);
      Var[] origins = variablesFrom(indexes, index -> s[index]);
      int[] lengths = valuesFrom(indexes, index -> jobs[index].duration);
      int[] heights = valuesFrom(indexes, index -> jobs[index].requiredQuantities[j]);
      cumulative(origins, lengths, heights, resourceCapacities[j]);
    }).note("resource constraints");

    minimize(s[nJobs - 1])
      .note("minimizing the makespan");
  }
}
\end{mcsp}
%    int[] lengths = valuesFrom(t, v -> jobs[v].duration);
%      int[] heights = valuesFrom(t, v -> jobs[v].requiredQuantities[j];

This model involves 1 array of variables and 2 types of constraints: \gb{cumulative} and \gb{intension} (\gb{lessEqual}).
A series of 16 instances has been selected for the competition.

\section{RLFAP}

%The Radio Link Frequency Assignment Problem (RLFAP) was abstracted from the real life application of assigning frequencies to radio links \cite{CGLSW_radio}.
%The interference level between the frequencies assigned to the different links has to be acceptable; otherwise communication will be distorted.
%The frequency assignments have to comply with certain regulations and physical characteristics of the transmitters.
%Moreover, the number of frequencies is to be minimised, because each frequency used in the network has to be reserved at a certain cost.
%In certain cases, some of the links may have pre-assigned frequencies which may be respected or preferred by the frequency assignment algorithm.

When radio communication links are assigned the same or closely related frequencies, there is a potential for interference.
Consider a radio communication network, defined by a set of radio links.
The radio link frequency assignment problem \cite{CGLSW_radio} is to assign, from limited spectral resources, a frequency to each of these links in such a way that all the links may operate together without noticeable interference.
Moreover, the assignment has to comply to certain regulations and physical constraints of the transmitters.
Among all such assignments, one will naturally prefer those which make good use of the available spectrum, trying to save the spectral resources for a later extension of the network.
%Furthermore, when several bands are available, and for various reasons (for example radio wave propagation, ease of deployment etc.), the lower bands are usually most favored and one should try to assign frequencies preferably from the lower bands.
Whereas we used simplified CSP instances of this problem in previous XCSP competitions, do note here that we have considered the original COP instances.

\subsection*{Description}
\begin{quote}
The description is rather complex. Hence, we refer the reader to \cite{CGLSW_radio}.
\end{quote}

\subsection*{Data}

As an illustration of data specifying an instance of this problem, we have:
\begin{json}
{
  "domains": {
    "1": [16, 30, 44, 58, 72, 86, 100, 114, 128, 142, 156, 254, 268, ...],
    "2":[30, 58, 86, 114, 142, 268, 296, 324, 352, 380, 414, 442, 470, ...],
    ...
  },
  "vars": [
    { "number": 13, "domain": 1, "value": "null", "mobility": "null" },
    { "number": 14, "domain": 1, "value": "null", "mobility": "null" },
    ...
  ],
  "ctrs":[
    { "x": 13, "y": 14, "equality": true, "limit": 238, "weight": 0 },
    { "x": 13, "y": 16, "equality": false, "limit": 186, "weight": 0 },
    ...
  ],
  "interferenceCosts": [0, 1000, 100, 10, 1],
  "mobilityCosts": [0, 0, 0, 0, 0]
}
\end{json}

\subsection*{Model}

The MCSP3 model used for the competition is: 

\begin{mcsp}
class Rlfap implements ProblemAPI {
  Map<Integer, int[]> domains;
  RlfapVar[] vars;
  RlfapCtr[] ctrs;
  int[] interferenceCosts;
  int[] mobilityCosts;
  
  class RlfapVar {
    int number;
    int domain;
    Integer value;
    Integer mobility;
  }
  
  class RlfapCtr {
    int x;
    int y;
    boolean equality;
    int limit;
    int weight;
  }

  private int index(int num) {
    return firstFrom(range(vars.length), i -> vars[i].number == num);
  }

  public void model() {
    int n = vars.length, e = ctrs.length;
    
    Var[] f = array("f", size(n), i -> dom(domains.get(vars[i].domain)),
      "f[i] is the frequency of the ith radio link");

    int[] indexes = select(range(n), i -> vars[i].value != null);
    Var[] fixedVars = variablesFrom(indexes, index -> mapVars.get(vars[index].number));
    int[] fixedVals = valuesFrom(indexes, index -> vars[index].value);
    instantiation(fixedVars, fixedVals).note("managing pre-assigned frequencies");
    
    forall(range(e), i -> {
      Var x = f[index(ctrs[i].x)], y = f[index(ctrs[i].y)];
      if (ctrs[i].equality)
        equal(dist(x, y), ctrs[i].limit);
      else
        greaterThan(dist(x, y), ctrs[i].limit);
    }).note("hard constraints on radio-links");
      
    if (modelVariant("span"))
      minimize(~MAXIMUM~, f)
        .note("minimizing the largest frequency");
    else if (modelVariant("card"))
      minimize(~NVALUES~, f)
        .note("minimizing the number of used frequencies");
  }
}
\end{mcsp}

Here, for simplicity, we only provide code for criteria SPAN and CARD.
The model involves 1 array of variables and 2 types of constraints: \gb{instantiation} and \gb{intension} (\gb{equal} and \gb{greaterThan}).
Note that instead of the auxiliary method \mn{index}, we could have used a map.
The complete series of 25 instances, 11 CELAR (scen) and 14 GRAPH, has been selected for the competition.
Also, the good old series 'scen11' of CSP instances has been selected.

\section{Social Golfers}

This is Problem \href{http://www.csplib.org/Problems/prob010}{010} on CSPLib, and called the Social Golfers Problem.

\subsection*{Description {\small (from Warwick Harvey on CSPLib)}}
\begin{quote}
``The coordinator of a local golf club has come to you with the following problem.
  In their club, there are 32 social golfers, each of whom play golf once a week, and always in groups of 4.
  They would like you to come up with a schedule of play for these golfers, to last as many weeks as possible, such that no golfer plays in the same group as any other golfer on more than one occasion.
  %Possible variants of the above problem include: finding a 10-week schedule with “maximum socialisation”; that is, as few repeated pairs as possible (this has the same solutions as the original problem if it is possible to have no repeated pairs), and finding a schedule of minimum length such that each golfer plays with every other golfer at least once (“full socialisation”).
The problem can easily be generalized to that of scheduling $m$ groups of $n$ golfers over $p$ weeks, such that no golfer plays in the same group as any other golfer twice (i.e. maximum socialisation is achieved).
\end{quote}

\subsection*{Data}

As an illustration of data specifying an instance of this problem, we have $(\vn{nGroups}=8, \vn{groupSize}=4, \vn{nWeeks}=6)$:

\subsection*{Model}

The MCSP3 model used for the competition is: 

\begin{mcsp}
class SocialGolfers implements ProblemAPI {
  int nGroups, groupSize, nWeeks;

  public void model() {
    int nPlayers = nGroups * groupSize;
    Range allGroups = range(nGroups);
    
    Var[][] x = array("x", size(nWeeks, nPlayers), dom(allGroups),
      "x[w][p] is the group in which player p plays in week w");

    forall(range(nWeeks), w -> cardinality(x[w], allGroups, occursEachExactly(groupSize)));
      .note("respecting the size of the groups");

    forall(range(nWeeks).range(nWeeks).range(nPlayers).range(nPlayers), (w1, w2, p1, p2) -> {
      if (w1 < w2 && p1 < p2)
        disjunction(ne(x[w1][p1], x[w1][p2]), ne(x[w2][p1], x[w2][p2]));
    }).note("ensuring that two players don't meet each other more than one time"); 
      
    block(() -> {
      instantiation(x[0], takingValues(range(nPlayers).map(p -> p / groupSize)));
      forall(range(groupSize), k -> instantiation(select(columnOf(x, k), w -> w > 0), takingValue(k)));
      lexMatrix(x, ~INCREASING~);
    }).tag(SYMMETRY_BREAKING);
  }
}
\end{mcsp}

This model involves 1 array of variables and 4 types of constraints: \gb{cardinality}, \gb{lexMatrix}, \gb{instantiation} and \gb{intension} (\gb{disjunction}).
Note the presence of a block for breaking some symmetries.
A series of 12 instances has been selected for the competition.

\section{Sports Scheduling}

This is Problem \href{http://www.csplib.org/Problems/prob026}{026} on CSPLib, called the Sports Tournament Scheduling.

\subsection*{Description (from Toby Walsh on CSPLib)}
\begin{quote}
  ``The problem is to schedule a tournament of $n$ teams over $n−1$ weeks, with each week divided into $n/2$ periods, and each period divided into two slots.
  The first team in each slot plays at home, whilst the second plays the first team away.
  A tournament must satisfy the following three constraints: every team plays once a week; every team plays at most twice in the same period over the tournament; every team plays every other team.''
\end{quote}

\subsection*{Data}

As an illustration of data specifying an instance of this problem, we have $n=10$.

\subsection*{Model}

The MCSP3 model used for the competition is: 

\begin{mcsp}
class SportsScheduling implements ProblemAPI {
  int nTeams;

  private int matchNumber(int team1, int team2) {
    int nPossibleMatches = (nTeams - 1) * nTeams / 2;
    return nPossibleMatches - ((nTeams - team1) * (nTeams - team1 - 1)) / 2 + (team2 - team1 - 1);
  }

  private Table matchs() {
    Table table = table();
    for (int team1 = 0; team1 < nTeams; team1++)
      for (int team2 = team1 + 1; team2 < nTeams; team2++)
        table.add(team1, team2, matchNumber(team1, team2));
    return table;
  }
  
  public void model() {
    int nWeeks = nTeams - 1, nPeriods = nTeams / 2, nPossibleMatches = (nTeams - 1) * nTeams / 2;
    Range allTeams = range(nTeams);
    
    Var[][] h = array("h", size(nPeriods, nWeeks), dom(allTeams),
      "h[p][w] is the number of the home opponent");
    Var[][] a = array("a", size(nPeriods, nWeeks), dom(allTeams),
      "a[p][w] is the number of the away opponent");
    Var[][] m = array("m", size(nPeriods, nWeeks), dom(range(nPossibleMatches)),
      "m[p][w] is the number of the match");
    
    allDifferent(m)
      .note("all matches are different (no team can play twice against another team)");
    forall(range(nPeriods).range(nWeeks), (p, w) -> extension(vars(h[p][w], a[p][w], m[p][w]), matchs()))
      .note("linking variables through ternary table constraints");
    forall(range(nWeeks), w -> allDifferent(vars(columnOf(h, w), columnOf(a, w))))
      .note("each week, all teams are different (each team plays each week)");
    forall(range(nPeriods), p -> cardinality(vars(h[p],a[p]), allTeams, occursEachBetween(1, 2)))
      .note("each team plays at most two times in each period"); 
    
    block(() -> {
      instantiation(columnOf(m, 0), takingValues(range(nPeriods).map(p -> matchNumber(2 * p, 2 * p + 1))))
        .note("the first week is set : 0 vs 1, 2 vs 3, 4 vs 5, etc.");
      forall(range(nWeeks), w -> exactly1(columnOf(m, w), takingValue(matchNumber(0, w + 1))))
        .note("the match '0 versus t' (with t strictly greater than 0) appears at week t-1");
    }).tag(SYMMETRY_BREAKING);
    
    block(() -> {
      Var[] hd = array("hd", size(nPeriods), dom(range(nTeams)),
        "hd[p] is the number of the home opponent for the dummy match of the period");
      Var[] ad = array("ad", size(nPeriods), dom(range(nTeams)),
        "ad[p] is the number of the away opponent for the dummy match of the period");

      allDifferent(vars(hd, ad))
        .note("all teams are different in the dummy week");
      forall(range(nPeriods), p -> cardinality(vars(h[p],hd[p],ad[p],a[p]),allTeams,occursEachExactly(2)))
        .note("Each team plays two times in each period");
      forall(range(nPeriods), p -> lessThan(hd[p], ad[p]))
        .tag(SYMMETRY_BREAKING);
    }).note("handling dummy week (variables and constraints)").tag("dummyWeek");
  }
}
\end{mcsp}

This model involves $3+2$ arrays of variables and 6 types of constraints: \gb{cardinality}, \gb{allDifferent}, \gb{count} (\gb{exactly1}), \gb{instantiation}, \gb{extension} and \gb{intension} (\gb{lessThan}).
Note that we could have used a cache for the table built by \mn{matchs}.
Also, the presence of the tag 'dummyWeek' allows us to easily activate or deactivate this part of the model, at parsing time.
A series of 10 instances has been selected for the competition.

\section{Steel Mill Slab}

This is Problem \href{http://www.csplib.org/Problems/prob038}{038} on CSPLib, called Steel Mill Slab Design.

\subsection*{Description {\small (from Ian Miguel on CSPLib)}}
\begin{quote}
  ``Steel is produced by casting molten iron into slabs.
  A steel mill can produce a finite number of slab sizes.
  An order has two properties, a colour corresponding to the route required through the steel mill and a weight.
  Given input orders, the problem is to assign the orders to slabs, the number and size of which are also to be determined, such that the total weight of steel produced is minimized. This assignment is subject to two further constraints:
\begin{itemize}
\item Capacity constraints: The total weight of orders assigned to a slab cannot exceed the slab capacity.
\item Colour constraints: Each slab can contain at most $p$ of $k$ total colours ($p$ is usually 2). 
\end{itemize}
The colour constraints arise because it is expensive to cut up slabs in order to send them to different parts of the mill.''
\end{quote}

\subsection*{Data}

As an illustration of data specifying an instance of this problem, we have:
\begin{json}
{
  "slabCapacities": [5, 7, 9, 11, 15, 18],
  "orders": [
    { "size": 1, "color": 2 },
    { "size": 3, "color": 1 },
    { "size": 2, "color": 1 },
    ...
  ]
}
\end{json}

\subsection*{Model}

The MCSP3 model(s) used for the competition is: 

\begin{mcsp}
class SteelMillSlab implements ProblemAPI {
  int[] slabCapacities;
  Order[] orders;
  
  class Order {
    int ~size~;
    int color;
  }

  // Repartition of orders in groups according to colors
  private Stream<int[]> colRep(int[] allColors) {
    return IntStream.of(allColors).mapToObj(c -> range(orders.length).select(i -> orders[i].color == c));
  }
  
  public void model() {
    slabCapacities = singleValuesIn(0, slabCapacities); // distinct sorted capacities (including 0)
    int maxCapacity = slabCapacities[slabCapacities.length - 1];
    int[] possibleLosses = range(maxCapacity+1).map(i -> minOf(select(slabCapacities, v -> v >= i)) - i);
    int[] sizes = valuesFrom(orders, order -> order.~size~);
    int totalSize = sumOf(sizes);
    int[] allColors = singleValuesFrom(orders, order -> order.color);
    int nOrders = orders.length, nSlabs = orders.length, nColors = allColors.length;
    
    Var[] sb = array("sb", size(nOrders), dom(range(nSlabs)),
      "sb[o] is the slab used to produce order o");
    Var[] ld = array("ld", size(nSlabs), dom(range(maxCapacity + 1)),
      "ld[s] is the load of slab s");
    Var[] ls = array("ls", size(nSlabs), dom(possibleLosses),
      "ls[s] is the loss of slab s");
    
    if (modelVariant("m1")) {
      forall(range(nSlabs), s -> sum(treesFrom(sb, x -> eq(x, s)), weightedBy(sizes), EQ, ld[s]))
        .note("computing (and checking) the load of each slab");
      forall(range(nSlabs), s -> extension(vars(ld[s], ls[s]), indexing(possibleLosses)))
        .note("computing the loss of each slab");
      forall(range(nSlabs), s -> sum(colRep(allColors).map(g -> or(treesFrom(g, o -> eq(sb[o],s)))),LE,2))
        .note("no more than two colors for each slab");
    }
    if (modelVariant("m2")) {
      Var[][] y = array("y", size(nSlabs, nOrders), dom(0, 1),
        "y[s][o] is 1 iff the slab s is used to produce the order o");
      Var[][] z = array("z", size(nSlabs, nColors), dom(0, 1),
        "z[s][c] is 1 iff the slab s is used to produce an order of color c");

      forall(range(nSlabs).range(nOrders), (s, o) -> equivalence(eq(sb[o], s), eq(y[s][o], 1)))
        .note("linking variables sb and y");
      forall(range(nSlabs).range(nOrders), (s, o) -> {
        int c = Utilities.indexOf(orders[o].color, allColors);
        implication(eq(sb[o], s), eq(z[s][c], 1));
      }).note("linking variables sb and z");
      forall(range(nSlabs), s -> sum(y[s], weightedBy(sizes), EQ, ld[s]))
        .note("computing (and checking) the load of each slab");
      forall(range(nSlabs), s -> extension(vars(ld[s], ls[s]), indexing(possibleLosses)))
        .note("computing the loss of each slab");
      forall(range(nSlabs), s -> sum(z[s], LE, 2))
        .note("no more than two colors for each slab");
    }
    
    sum(ld, EQ, totalSize)
      .tag(REDUNDANT_CONSTRAINTS);

    block(() -> {
      decreasing(load);
      forall(range(nOrders).range(nOrders), (i, j) -> {
	if (i < j && orders[i].~size~ == orders[j].~size~ && orders[i].color == orders[j].color)
	  lessEqual(sb[i], sb[j]);
      });
    }).tag(SYMMETRY_BREAKING);
    
    minimize(~SUM~, ls)
      .note("minimizing summed up loss");
  }
}
\end{mcsp}

Two model variants, 'm1' and 'm2', have been considered.
The first model variant involves 3 arrays of variables and 4 types of constraints: \gb{sum} (over trees), \gb{extension}, \gb{ordered} (\gb{decreasing}) and \gb{intension} (\gb{lessEqual}).
The second model variant involves $3+2$ arrays of variables and 4 types of constraints: \gb{sum}, \gb{extension}, \gb{ordered} (\gb{decreasing}) and \gb{intension} (\gb{equivalence}, \gb{implication} and \gb{lessEqual}).
Noe that there is a redundant constraint and a block of symmetry-breaking constraints.
The series 'm2s' corresponds to the model 'm2' without the redundant and symmetry-breaking constraints.
A series of $6+6+5$ instances has been selected for the main track.
For the mini-track, instances from model 'm2s' have been slightly reformulated.

\section{Still Life}

This is Problem \href{http://www.csplib.org/Problems/prob038}{032} on CSPLib, called Maximum density still life.
This problem arises from the Game of Life, invented by John Horton Conway in the 1960s and popularized by Martin Gardner in his Scientific American columns.

\subsection*{Description {\small (from Barbara Smith on CSPLib)}}
\begin{quote}
``Life is played on a squared board. %, considered to extend to infinity in all directions.
Each square of the board is a cell, which at any time during the game is either alive or dead. A cell has eight neighbours.
The configuration of live and dead cells at time $t$ leads to a new configuration at time $t+1$ according to the rules of the game:
\begin{itemize}
\item if a cell has exactly three living neighbours at time $t$, it is alive at time $t+1$
\item if a cell has exactly two living neighbours at time $t$ it is in the same state at time $t+1$ as it was at time $t$
\item otherwise, the cell is dead at time $t+1$
\end{itemize}
A stable pattern, or still-life, is not changed by these rules.
Hence, every cell that has exactly three live neighbours is alive, and every cell that has fewer than two or more than three live neighbours is dead. %(An empty board is a still-life, for instance.)
What is the densest possible still-life, i.e. the pattern with the largest number of live cells, that can be fitted into %an $n \times n$ section of
the board? %, with all the rest of the board dead?
''
\end{quote}

\subsection*{Data}

As an illustration of data specifying an instance of this problem, we have $n=8$.

\subsection*{Model}

The MCSP3 model used for the competition is: 

\begin{mcsp}
class StillLife implements ProblemAPI {
  int n;

  @NotData
  private Predicate<int[]> p = x -> {
    int s1 = x[0] + x[1] + x[2] + x[3] + x[5] + x[6] + x[7] + x[8];
    int s2 = x[0] * x[2] + x[2] * x[8] + x[8] * x[6] + x[6] * x[0] + x[1] + x[3] + x[5] + x[7];
    int s3 = x[1] + x[3] + x[5] + x[7];
    return (x[4] != 1 || s1 >= 2) && (x[4] != 1 || s1 <= 3) && (x[4] != 0 || s1 != 3)
      && (x[4] != 1 || s2 > 1 || x[9] >= 1) && (x[4] != 1 || s2 > 0 || x[9] >= 2)
      && (x[4] != 0 || s3 < 4 || x[9] >= 2) && (x[4] != 0 || s3 > 1 || x[9] >= 1)
      && (x[4] != 0 || s3 > 0 || x[9] >= 2);
  };
  
  public void model() {
    Var[][] x = array("x", size(n + 2, n + 2), dom(0, 1),
      "x[i][j] is 1 iff the cell at row i and column j is alive (note that there is a border)");
    Var[][] w = array("w", size(n + 2, n + 2), dom(0, 1, 2),
      "w[i][j] is the wastage for the cell at row i and column j");
    Var[] ws = array("ws", size(n + 2), dom(range(2 * (n + 2) * (n + 2) + 1)),
      "ws[i] is the wastage sum for cells at row i");
    Var z = ~\textcolor{dred}{var}~("z", dom(range(n * n + 1)),
      "z is the number of alive cells");
    
    block(() -> {
      instantiation(x[0], takingValue(0));
      instantiation(x[n + 1], takingValue(0));
      instantiation(columnOf(x, 0), takingValue(0));
      instantiation(columnOf(x, n + 1), takingValue(0));
    }).note("cells at the border are assumed to be dead");
    
    block(() -> {
      Table conflicts = table(NEGATIVE).add(1, 1, 1);
      slide(x[1], range(n), j -> extension(vars(x[1][j], x[1][j + 1], x[1][j + 2]), conflicts));
      slide(x[n], range(n), j -> extension(vars(x[n][j], x[n][j + 1], x[n][j + 2]), conflicts));
      slide(columnOf(x, 1), range(n), i -> extension(vars(x[i][1], x[i + 1][1], x[i + 2][1]), conflicts));
      slide(columnOf(x, n), range(n), i -> extension(vars(x[i][n], x[i + 1][n], x[i + 2][n]), conflicts));
    }).note("ensuring that cells at the border remain dead");
    
    int[][] tuples = allCartesian(vals(2, 2, 2, 2, 2, 2, 2, 2, 2, 3), p);
    forall(range(1, n + 1).range(1, n + 1), (i, j) -> {
      Var[] neighbors = select(x, range(i - 1, i + 2).range(j - 1, j + 2));
      extension(vars(neighbors, w[i][j]), tuples);
    }).note("still life + wastage constraints");
    
    block(() -> {
      forall(range(1, n + 1), j -> equal(add(w[0][j], x[1][j]), 1));
      forall(range(1, n + 1), j -> equal(add(w[n + 1][j], x[n][j]), 1));
      forall(range(1, n + 1), i -> equal(add(w[i][0], x[i][1]), 1));
      forall(range(1, n + 1), i -> equal(add(w[i][n + 1], x[i][n]), 1));
    }).note("managing wastage on the border");
    
    forall(range(n + 2), i -> sum(i == 0 ? w[0] : vars(ws[i - 1], w[i]), EQ, ws[i]))
      .note("summing wastage");
    sum(vars(z, ws[n + 1]), vals(4, 1), EQ, 2 * n * n + 4 * n)
      .note("setting the value of the objective");
    forall(range(n + 1), i -> greaterEqual(sub(ws[n + 1], ws[i]), 2 * ((n - i) / 3) + n / 3))
      .tag(REDUNDANT_CONSTRAINTS);
    
    maximize(z)
      .note("maximizing the number of alive cells");
  }
}
\end{mcsp}

This model involves 3 arrays of variables, 1 stand-alone variable and 4 types of constraints: \gb{instantiation}, \gb{extension}, \gb{sum} and \gb{intension} (\gb{equal} and \gb{greaterEqual}).
This model is in the spirit of the 'wastage' model from the Minizinc challenge 2012.
Interestingly, note how we manage both Still life and wastage constraints with table constraints.
To build them, we filter tuples from a Cartesian product by using a predicate (field $p$ which is a lambda function not being considered as a piece of data by means of the annotation @NotData) for the competition.
A series of 13 instances has been selected.

\section{Strip Packing}

This is the Two-Dimensional Strip Packing Problem (TDSP), as introduced, for example, in \cite{HT_empirical}.
See also the \href{http://people.brunel.ac.uk/~mastjjb/jeb/info.html}{OR-library}.

\subsection*{Description}
\begin{quote}
``In the Two-Dimensional Strip Packing Problem (TDSP), one has to pack a set of rectangular items into a rectangular strip.'' 
\end{quote}

\subsection*{Data}

As an illustration of data specifying an instance of this problem, we have:
\begin{json}
{
  "container": { "width": 20, "height": 20 },
  "rectangles":[
    { "width": 2, "height": 12 },
    { "width": 7, "height": 12 },
    ...
  ]
}
\end{json}

\subsection*{Model}

The MCSP3 model used for the competition is: 

\begin{mcsp}
class StripPacking implements ProblemAPI {

  Rectangle container;
  Rectangle[] items;
  
  class Rectangle {
    int width;
    int height;
  }
  
  public void model() {
    int nItems = items.length;
    
    Var[] x = array("x", size(nItems), dom(range(container.width)),
      "x[i] is the x-coordinate of the ith rectangle");
    Var[] y = array("y", size(nItems), dom(range(container.height)),
      "y[i] is the y-coordinate of the ith rectangle");
    Var[] w = array("w", size(nItems), i -> dom(items[i].width, items[i].height),
      "w[i] is the width of the ith rectangle");
    Var[] h = array("h", size(nItems), i -> dom(items[i].width, items[i].height),
      "h[i] is the height of the ith rectangle");
    Var[] r = array("r", size(nItems), dom(0, 1),
      "r[i] is 1 iff the ith rectangle is rotated by 90 degrees");
    
    forall(range(nItems), i -> lessEqual(add(x[i], w[i]), container.width))
      .note("horizontal control");
    forall(range(nItems), i -> lessEqual(add(y[i], h[i]), container.height))
      .note("vertical control");
    forall(range(nItems), i -> {
      Table table = table();
      table.add(0, items[i].width, items[i].height).add(1, items[i].height, items[i].width);
      extension(vars(r[i], w[i], h[i]), table);
    }).note("managing rotation");
    noOverlap(transpose(x, y), transpose(w, h))
     .note("no overlapping between rectangles");
  }
}
\end{mcsp}

This model involves 5 arrays of variables, and 3 types of constraints: \gb{noOverlap}, \gb{extension} and \gb{intension} (\gb{lessEqual}).
A series of 12 instances has been selected for the competition.

\section{Subgraph Isomorphism}

\subsection*{Description}
\begin{quote}
``In theoretical computer science, the subgraph isomorphism problem is a computational task in which two graphs $G$ and $H$ are given as input, and one must determine whether $G$ contains a subgraph that is isomorphic to $H$.''
\end{quote}

\subsection*{Data}

As an illustration of data specifying an instance of this problem, we have:
\begin{json}
{
  "nPatternNodes": 180,
  "nTargetNodes": 200,
  "patternEdges": [ [0,1], [0,3], [0,17], ... ],
  "targetEdges": [ [0,34], [0,65], [0,129], ...]
}
\end{json}

\subsection*{Model}

The MCSP3 model used for the competition is: 

\begin{mcsp}
class Subisomorphism implements ProblemAPI { 
  int nPatternNodes, nTargetNodes;
  int[][] patternEdges, targetEdges;
  
  private int[] selfLoops(int[][] edges) {
    return Stream.of(edges).filter(t -> t[0] == t[1]).mapToInt(t -> t[0]).toArray();
  }
  
  private int degree(int[][] edges, int node) {
    return (int) Stream.of(edges).filter(t -> t[0] == node || t[1] == node).~count~();
  }
  
  private Table bothWayTable() {
    Table table = table().add(targetEdges);
    table.add(Stream.of(targetEdges).map(t -> tuple(t[1], t[0]))); // reversed tuples
    return table;
  }
  
  public void model() {
    int[] pLoops = selfLoops(patternEdges);
    int[] tLoops = selfLoops(targetEdges);
    int[] pDegrees = range(nPatternNodes).map(i -> degree(patternEdges, i));
    int[] tDegrees = range(nTargetNodes).map(i -> degree(targetEdges, i));
    int l = pLoops.length, e = patternEdges.length;
    
    Var[] x = array("x", size(nPatternNodes), dom(range(nTargetNodes)),
      "x[i] is the node from the target graph to which the ith node of the pattern graph is mapped.");
    
    allDifferent(x)
      .note("ensuring injectivity");
    forall(range(l), i -> extension(x[pLoops[i]], tLoops))
      .note("being careful of self-loops");  
    forall(range(e), i -> extension(vars(x[patternEdges[i][0]], x[patternEdges[i][1]]), bothWayTable()))
      .note("preserving edges");
    
    forall(range(nPatternNodes), i -> {
      int[] conflicts = range(nTargetNodes).select(j -> tDegrees[j] < pDegrees[i]);
      if (conflicts.length > 0)
        extension(x[i], conflicts, NEGATIVE);
    }).tag(REDUNDANT_CONSTRAINTS);
  }
}
\end{mcsp}

This model involves 1 array of variables and 2 types of constraints: \gb{allDifferent} and \gb{extension}.
Note that we could have used a cache for the table built by \mn{bothWayTable}.
There is a block with redundant unary constraints.
A series of 11 instances has been selected for the competition.

\section{Sum Coloring}

\subsection*{Description}
\begin{quote}
``In graph theory, a sum coloring of a graph is a labeling of its nodes by positive integers, with no two adjacent nodes having equal labels, that minimizes the sum of the labels.''
\end{quote}

\subsection*{Data}

As an illustration of data specifying an instance of this problem, we have:

\begin{json}
{
  "nNodes": 30,
  "edges": [ [0,1], [0,3], [0,10], [0,12], ...]
}
\end{json}

\subsection*{Model}

The MCSP3 model used for the competition is: 

\begin{mcsp}
class SumColoring implements ProblemAPI {
  int nNodes;
  int[][] edges;
  
  public void model() {
    int nEdges = edges.length;
    
    Var[] c = array("c", size(nNodes), dom(range(nNodes)),
      "c[i] is the color assigned to the ith node");
    
    forall(range(nEdges), i -> different(c[edges[i][0]], c[edges[i][1]]))
      .note("two adjacent nodes must be colored differently");

    minimize(~SUM~, c)
      .note("minimizing the sum of colors assigned to nodes");
  }
}
\end{mcsp}

This model only involves 1 array of variables and 1 type of constraint: \gb{intension} (\gb{different}).
A series of 14 instances has been selected for the competition.

\section{TAL}

TAL is a problem of natural language processing.

\subsection*{Description {\small (from work by R\'emi Coletta and Jean-Philippe Prost})}
\begin{quote}
The description is rather complex. Hence, we refer the reader to \cite{PCLcompilaton}.
\end{quote}

\subsection*{Data}

As an illustration of data specifying an instance of this problem, we have:

\begin{json}
{
  "maxArity": 7,
  "maxHeight": -1,
  "sentence": [15, 11, 13, 9, 1, 11, 7, 4],
  "grammar": [
    null,
    [ [0,0,2147483646,0], [1,2,13,0], [1,2,26,16], ... ],
    [ [0,0,2147483646,2147483646,0], [1,2,13,2147483646,0], ... ],
    [ [0,0,2147483646,2147483646,2147483646,2147483646,0], ... ],
  ],
  "tokens": ["AP", "COORD", "NP", "PP", "SENT", ..., "VPP", "VPR", "VS"],
  "costs": [0, 1, 2, 4, 8, 16]
}
\end{json}

\begin{mcsp}
class Tal implements ProblemAPI {
  int maxArity;
  int maxHeight;
  int[] sentence;
  int[][][] grammar; // grammar[i] gives the grammar tuples of arity i
  String[] tokens;
  int[] costs;
  
  private void predicate(Var[][] l, Var[][] a, int i, int j, int[] lengths) {
    Range r = range(i != 0 && j == lengths[i] - 1 ? 1 : 0, Math.min(j + 1, maxArity));
    intension(xor(eq(l[i][j], 0), treesFrom(r, k -> ge(a[i + 1][j - k], k + 1))));
  }

  private Table tableFor(int vectorLength) {
    int arity = vectorLength + 2;
    Table table = table().add(range(arity).map(k -> k == arity - 1 ? 0 : STAR)); 
    for (int i = 0; i < vectorLength; i++)
      for (int j = 1; j < tokens.length + 1; j++) {
        int[] tuple = repeat(STAR, arity);
        tuple[0] = i;
        tuple[i + 1] = j;
        tuple[arity - 1] = j;
        table.add(tuple); 
      }
    return table;
  }
  
  public void model() {
    int nWords = sentence.length, nLevels = nWords * 2, nTokens = tokens.length;
    int[] lengths = valuesFrom(range(nLevels), i -> i==0 ? nWords : nWords - (int) Math.floor((i+1)/2)+1);
    
    Var[][] c = array("c", size(nLevels, nWords), (i, j) -> (i == 0 || i % 2 == 1) && j < lengths[i] ?
       dom(costs) : dom(0), "c[i][j] is the cost of the jth word at the ith level");
    Var[][] l = array("l", size(nLevels, nWords), (i, j) -> j < lengths[i] ?
       dom(range(nTokens + 1)) : dom(0), "l[i][j] is the label of the jth word at the ith level");
    Var[][] a = array("a", size(nLevels, nWords), (i, j) -> i % 2 == 1 && j < lengths[i] ?
       dom(range(maxArity + 1)) : dom(0), "a[i][j] is the arity of the jth word at the ith level");
    Var[][] x = array("x", size(nLevels, nWords), (i, j) -> 0 < i && i % 2 == 0 && j < lengths[i] ?
       dom(range(lengths[i])) : dom(0), "x[i][j] is the index of the jth word at the ith level");
    Var[] s = array("s", size(nLevels - 2), i -> dom(range(lengths[i + 1])));
    
    forall(range(1, nLevels-1), i -> exactly(select(l[i], j -> j<lengths[i]), takingValue(0), s[i - 1]));
    forall(range(1, nLevels - 1, 2), i -> equal(s[i - 1], s[i]));
    
    forall(range(nWords), j -> equal(c[0][j], 0))
      .note("on row 0, costs are 0");
    forall(range(nWords), j -> equal(l[0][j], sentence[j]))
      .note("on row 0, the jth label is the jth word of the sentence");
    forall(range(1, nLevels), i -> greaterThan(l[i][0], 0))
      .note("on column 0, labels are 0");
    
    forall(range(1, nLevels, 2), p -> greaterThan(a[p][0], 0));
    forall(range(1, nLevels, 2).range(1, nWords), (i, j) -> {
      if (j < lengths[i] && j + maxArity > lengths[i - 1])
        lessEqual(a[i][j], lengths[i - 1] - j);
    });
    
    forall(range(2, nLevels, 2), i -> {
      equal(x[i][0], 0);
      equal(l[i][0], l[i - 1][0]);
    });
    
    forall(range(2, nLevels, 2), i -> forall(range(1, lengths[i]), j -> {
      greaterEqual(x[i][j], j);
      implication(eq(l[i][j], 0), eq(x[i][j], lengths[i] - 1));
      implication(gt(l[i][j], 0), gt(x[i][j], x[i][j - 1]));
      implication(eq(l[i][j - 1], 0), eq(l[i][j], 0));
      Var[] vect = select(l[i - 1], k -> k < lengths[i - 1]);
      extension(vars(x[i][j], vect, l[i][j]), tableFor(vect.length));
    }));
    
    forall(range(1, nLevels, 2), i -> forall(range(lengths[i]), j -> {
      equivalence(eq(l[i][j], 0), eq(a[i][j], 0));
      int nPossibleSons = Math.min(lengths[i - 1] - j, maxArity);
      Var[] scp = vars(a[i][j], l[i][j], select(l[i - 1], range(j, j + nPossibleSons)), c[i][j]);
      extension(scp, grammar[nPossibleSons]);    
    }));
    
    forall(range(0, nLevels, 2), i -> forall(range(lengths[i]), j -> predicate(l, a, i, j, lengths)));
    
    if (0 < maxHeight && 2 * maxHeight < l.length)
      equal(l[2 * maxHeight][1], 0); 
    
    minimize(~SUM~, vars(c))
      .note("minimizing summed up cost");
  }
}
\end{mcsp}

This model involves 5 arrays of variables and 3 types of constraints: \gb{count} (\gb{exactly}, \gb{extension} and \gb{intension} (\gb{equal}, \gb{greaterThan}, \gb{implication} and \gb{equivalence}). 
A series of 10 instances has been selected for the competition.

\section{Template Design}

This is Problem \href{http://www.csplib.org/Problems/prob038}{002} on CSPLib, called Template Design.
See also \cite{PS_integer}.

\subsection*{Description {\small (from Barbara Smith on CSPLib)}}
\begin{quote}
  ``This problem arises from a colour printing firm which produces a variety of products from thin board, including cartons for human and animal food and magazine inserts.
  Food products, for example, are often marketed as a basic brand with several variations (typically flavours).
  Packaging for such variations usually has the same overall design, in particular the same size and shape, but differs in a small proportion of the text displayed and/or in colour.
  For instance, two variations of a cat food carton may differ only in that on one is printed ‘Chicken Flavour’ on a blue background whereas the other has ‘Rabbit Flavour’ printed on a green background.
  A typical order is for a variety of quantities of several design variations.
  Because each variation is identical in dimension, we know in advance exactly how many items can be printed on each mother sheet of board, whose dimensions are largely determined by the dimensions of the printing machinery.
  Each mother sheet is printed from a template, consisting of a thin aluminium sheet on which the design for several of the variations is etched.
  The problem is to decide, firstly, how many distinct templates to produce, and secondly, which variations, and how many copies of each, to include on each template.''
\end{quote}

\subsection*{Data}

As an illustration of data specifying an instance of this problem, we have:

\begin{json}
{
  "nSlots": 9,
  "demands": [250, 255, 260, 500, 500, 800, 1100]
}
\end{json}

\subsection*{Model}

The MCSP3 model(s) used for the competition is: 

\begin{mcsp}
class TemplateDesign implements ProblemAPI {
  int nSlots;
  int[] demands;

  private int lb(int v) {
    return (int) Math.ceil(demands[v] * 0.95);
  }
  
  private int ub(int v) {
    return (int) Math.floor(demands[v] * 1.1);
  }
   
  public void model() {
    int maxDemand = maxOf(demands), nVariations = demands.length, nTemplates = nVariations;
     
    Var[][] d = array("d", size(nTemplates, nVariations), dom(range(nSlots + 1)),
      "d[t][v] is the number of occurrences of variation v on template t");
    Var[] p = array("p", size(nTemplates), dom(range(maxDemand + 1)),
      "p[t] is the number of printings of template t");
    Var[] u = array("u", size(nTemplates), dom(0, 1),
      "u[t] is 1 iff the template t is used");
    
    forall(range(nTemplates), t -> sum(d[t], EQ, nSlots))
      .note("all slots of all templates are used");
    forall(range(nTemplates), t -> equivalence(eq(u[t], 1), gt(p[t], 0)))
      .note("if a template is used, it is printed at least once")
    
    if (modelVariant("m1")) {
      Var[][] pv = array("pv", size(nTemplates, nVariations), (t, v) -> dom(range(ub(v))), 
        "pv[t][v] is the number of printings of variation v by using template t");
      
      forall(range(nTemplates).range(nVariations), (t, v) -> equal(mul(p[t], d[t][v]), pv[t][v]))
        .note("linking variables of arrays p and pv");
      forall(range(nVariations), v -> sum(columnOf(pv, v), IN, range(lb(v), ub(v) + 1)))
        .note("respecting printing bounds for each variation v");
    }

    if (modelVariant("m2"))
      forall(range(nVariations), v -> sum(p, weightedBy(columnOf(d, v)), IN, range(lb(v), ub(v) + 1)))
        .note("respecting printing bounds for each variation v");
    
    block(() -> {
      decreasing(p);
      forall(range(nTemplates), t -> equivalence(eq(u[t], 0), eq(d[t][0], nSlots)));
    }).tag(SYMMETRY_BREAKING);
    
    minimize(~SUM~, u)
      .note("minimizing the number of used templates");
  }
}
\end{mcsp}

Two model variants, 'm1' and 'm2', have been considered.
These model variants involve $3(+1)$ arrays of variables, and 3 types of constraints: \gb{sum}, \gb{ordered} (\gb{decreasing}) and \gb{intension} (\gb{equivalence} and \gb{equal}).
The model variant 'm1' introduces some auxiliary variables in order to post a basic form of \gb{sum}.
Note that there is a block for breaking a few symmetries.
A series of $3 \times 5$ instances has been selected for the competition: 5 instances for models 'm1', 'm2' and 'm1s'' which is 'm1' without the symmetry-breaking constraints.

\section{Traveling Tournament}

This problem is related to Problem \href{http://www.csplib.org/Problems/prob068}{068} on CSPLib.
Many relevant information can be found at \href{http://mat.gsia.cmu.edu/TOURN/}{http://mat.gsia.cmu.edu/TOURN/}.
See also \cite{ENT_solving}.

\subsection*{Description}
\begin{quote}
``The Traveling Tournament Problem (TTP) is defined as follows.
  A double round robin tournament is played by an even number of teams.
  Each team has its own venue at its home city.
  All teams are initially at their home cities, to where they return after their last away game.
  The distance from the home city of a team to that of another team is known beforehand.
  Whenever a team plays two consecutive away games, it travels directly from the venue of the first opponent to that of the second.
  The problem calls for a schedule such that no team plays more than (two or) three consecutive home games or more than (two or) three consecutive away games,
  there are no consecutive games involving the same pair of teams, and the total distance traveled by the teams during the tournament is minimized.''
\end{quote}

\subsection*{Data}

As an illustration of data specifying an instance of this problem, we have:

\begin{json}
{
  "distances": [
    [0, 10, 15, 34],
    [10, 0, 22, 32],
    [15, 22, 0, 47],
    [34, 32, 47, 0]
  ]
}
\end{json}

\subsection*{Model}

The MCSP3 model used for the competition is: 

\begin{mcsp}
class TravelingTournament implements ProblemAPI {
  int[][] distances;

  private Table tableEnd(int i) {
    Table table = table().add(1, STAR, 0); // when playing at home, travel distance is 0
    for (int j = 0; j < distances.length; j++)
      if (j != i)
        table.add(0, j, distances[i][j]);
    return table;
  }
  
  private Table tableOther(int i) {
    int nTeams = distances.length;
    Table table = table().add(1, 1, STAR, STAR, 0);
    for (int j = 0; j < nTeams; j++)
      if (j != i) {
        table.add(0, 1, j, STAR, distances[i][j]);
        table.add(1, 0, STAR, j, distances[i][j]);
      }
    for (int j1 = 0; j1 < nTeams; j1++)
      for (int j2 = 0; j2 < nTeams; j2++)
        if (j1 != i && j2 != i && j1 != j2)
          table.add(0, 0, j1, j2, distances[j1][j2]);
    return table;    
  }

  private Automaton automat() {
    String transitions = "(q,0,q01)(q,1,q11)(q01,0,q02)(q01,1,q11)(q11,0,q01)(q11,1,q12)(q02,1,q11)(q12,0,q01)";
    if (modelVariant("a2"))
      return automaton("q", transitions, finalStates("q01", "q02", "q11", "q12"));
    transitions = transitions + "(q02,0,q03)(q12,1,q13)(q03,1,q11)(q13,0,q01)";
    return automaton("q", transitions, finalStates("q01", "q02", "q03", "q11", "q12", "q13"));
  }

  private int[] allTeamsExcept(int i) {
    return range(distances.length).select(j -> j != i);
  }
  
  public void model() {
    int nTeams = distances.length, nRounds = nTeams * 2 - 2;
    
    Var[][] o = array("o", size(nTeams, nRounds), dom(range(nTeams)),
      "o[i][k] is the opponent (team) of the ith team  at the kth round");
    Var[][] h = array("h", size(nTeams, nRounds), dom(0, 1),
      "h[i][k] is 1 iff the ith team plays at home at the kth round");
    Var[][] a = array("a", size(nTeams, nRounds), dom(0, 1),
      "a[i][k] is 0 iff the ith team plays away at the kth round");
    Var[][] t = array("t", size(nTeams, nRounds + 1), dom(distances),
      "t[i][k] is the travelled distance by the ith team at the kth round; an additional round for returning at home.");
    
    forall(range(nTeams), i -> cardinality(o[i], allTeamsExcept(i), CLOSED, occursEachExactly(2)))
      .note("each team must play exactly two times against each other team");
    forall(range(nTeams).range(nRounds), (i, k) -> element(columnOf(o, k), at(o[i][k]), takingValue(i)))
      .note("if team i plays against j at round k, then team j plays against i at round k");  
    forall(range(nTeams).range(nRounds), (i, k) -> equal(h[i][k], not(a[i][k])))
      .note("playing home at round k iff not playing away at round k");
    forall(range(nTeams).range(nRounds), (i,k) -> element(columnOf(h,k),at(o[i][k]),takingValue(a[i][k])))
      .note("channeling the three arrays");

    forall(range(nTeams).range(nRounds).range(nRounds), (i, k1, k2) -> {
      if (k1 + 1 < k2)
        implication(eq(o[i][k1], o[i][k2]), ne(h[i][k1], h[i][k2]));
    }).note("playing against the same team must be done once at home and once away");
     
    forall(range(nTeams), i -> regular(h[i], automat()))
      .note("verifying the number of consecutive games at home, and consecutive games away");
    
    forall(range(nTeams), i -> {
      extension(vars(h[i][0], o[i][0], t[i][0]), tableEnd(i)));
      extension(vars(h[i][nRounds-1], o[i][nRounds-1], t[i][nRounds]), tableEnd(i)));
    }.note("handling travelling for the first and last games");

    forall(range(nTeams).range(nRounds - 1), (i, k) ->
      extension(vars(h[i][k], h[i][k + 1], o[i][k], o[i][k + 1], t[i][k + 1]), tableOther(i)))
    .note("handling travelling for two successive games");

    forall(range(nRounds), k -> allDifferent(columnOf(o, k))).tag(REDUNDANT_CONSTRAINTS)
      .note("at each round, opponents are all different");
    lessThan(o[0][0], o[0][nRounds - 1]).tag(SYMMETRY_BREAKING);
 
    minimize(~SUM~, t)
      .note("minimizing summed up travelled distance");
  }
}
\end{mcsp}

This model involves 4 arrays of variables and 6 types of constraints: \gb{cardinality}, \gb{regular}, \gb{element}, \gb{allDifferent}, \gb{extension} and \gb{intension} (\gb{equal}, \gb{implication} and \gb{lessThan}).
Note that we could have used a cache for the tables built by \mn{tableEnd} and \mn{tableOther} as well as for the automaton built by \mn{automat}. 
A series of 14 instances has been selected for the competition.

\section{Travelling Salesman}

This is the famous Travelling Salesman Problem (TSP).
See for example \href{https://www.iwr.uni-heidelberg.de/groups/comopt/software/TSPLIB95/}{TSPLIB}.

\subsection*{Description}
\begin{quote}
``Given a list of cities and the distances between each pair of cities, what is the shortest possible route that visits each city and returns to the origin city?''
\end{quote}

\subsection*{Data}

As an illustration of data specifying an instance of this problem, we have:

\begin{json}
{
  "distances": [
    [0, 5, 6, 6, 6],
    [5, 0, 9, 8, 4],
    [6, 9, 0, 1, 7],
    [6, 8, 1, 0, 6],
    [6, 4, 7, 6, 0]
  ]
}
\end{json}

\subsection*{Model}

The MCSP3 model used for the competition is: 

\begin{mcsp}
class TravellingSalesman implements ProblemAPI {
  int[][] distances;

  private Table distTable() {
    Table table = table();
    for (int i = 0; i < distances.length; i++)
      for (int j = 0; j < distances.length; j++)
        if (i != j)
          table.add(i, j, distances[i][j]);
    return table;
  }

  public void model() {
    int nCities = distances.length;
    
    Var[] c = array("c", size(nCities), dom(range(nCities)),
      "c[i] is the ith city of the tour");
    Var[] d = array("d", size(nCities), dom(distances),
      "d[i] is the distance between the cities i and i+1");
    
    allDifferent(c)
      .note("visiting each city only once");   
    forall(range(nCities), i -> extension(vars(c[i], c[(i + 1) % nCities], d[i]), distTable()))
      .note("computing the distance between any two successive cities in the tour");
      
    minimize(~SUM~, d)
      .note("minimizing the total distance");
  }
}
\end{mcsp}

This model involves two arrays of variables and two types of constraints: \gb{allDifferent} and \gb{extension}.
Of course, for large number of cities, the table built by the auxiliary method \mn{distTable} should be stored and reused, instead of being systematically computed.
A series of 12 instances has been generated for the competition.

\chapter{Solvers}

In this chapter, we introduce the solvers and teams having participated to the \xt Competition 2018.
When names of solvers are given in italic font, it means that a short description of these solvers are given by authors in the following pages. 

\begin{itemize}
\item  {\em BTD, miniBTD} (Philippe Jegou, Helene Kanso and Cyril Terrioux)
\item  {\em BTD\_12, miniBTD\_12} (Philippe Jegou, Djamal Habet, Helene Kanso and Cyril Terrioux)
\item  {\em Choco-solver} (Charles Prud'homme and Jean-Guillaume Fages)
\item  {\em Concrete} (Julien Vion)
\item  {\em CoSoCo} (Gilles Audemard)
\item  GG's minicp (Arnaud Gellens and Simon Gustin)
\item  {\em macht, minimacht} (Djamal Habet and Cyril Terrioux)
\item  MiniCPFever (Victor Joos and Antoine Vanderschueren)
\item  {\em Mistral-2.0} ( Emmanuel Hebrard and Mohamed Siala)
\item  {\em NACRE} (Gaël Glorian)
\item  {\em OscaR} (OscaR Team)
\item  {\em PicatSAT} (Neng-Fa Zhou and Hakan Kjellerstrand)
\item  {\em Sat4j-CSP} (Daniel Le Berre and Emmanuel Lonca)
\item  {\em scop} (Takehide Soh, Daniel Le Berre, Mutsunori Banbara, Naoyuki Tamura)
\item  slowpoke (Alexandre Gerlache and vincent vandervilt)
\item  Solver of Xavier Schul and Yvhan Smal (Xavier Schul and Yvhan Smal)
\item  SuperSolver (Florian Stevenart Meeus and Jean-Baptiste Macq)
\item  The dodo solver (Alexandre Dubray)
\end{itemize}

%\section{BTD}
\addcontentsline{toc}{section}{\numberline{}BTD}
\includepdf[pages=-,pagecommand={\thispagestyle{plain}}]{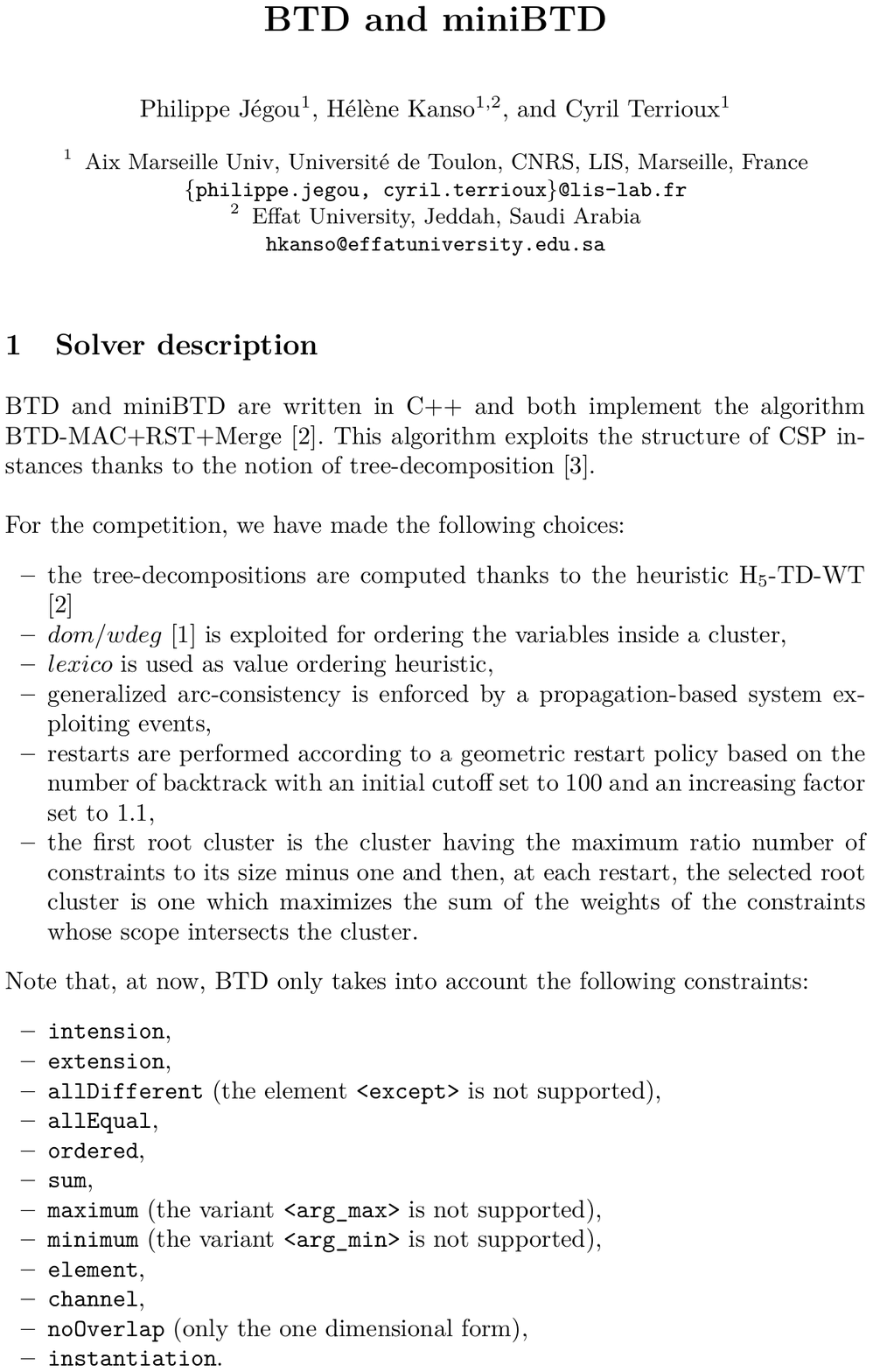}
\addcontentsline{toc}{section}{\numberline{}BTD\_12}
\includepdf[pages=-,pagecommand={\thispagestyle{plain}}]{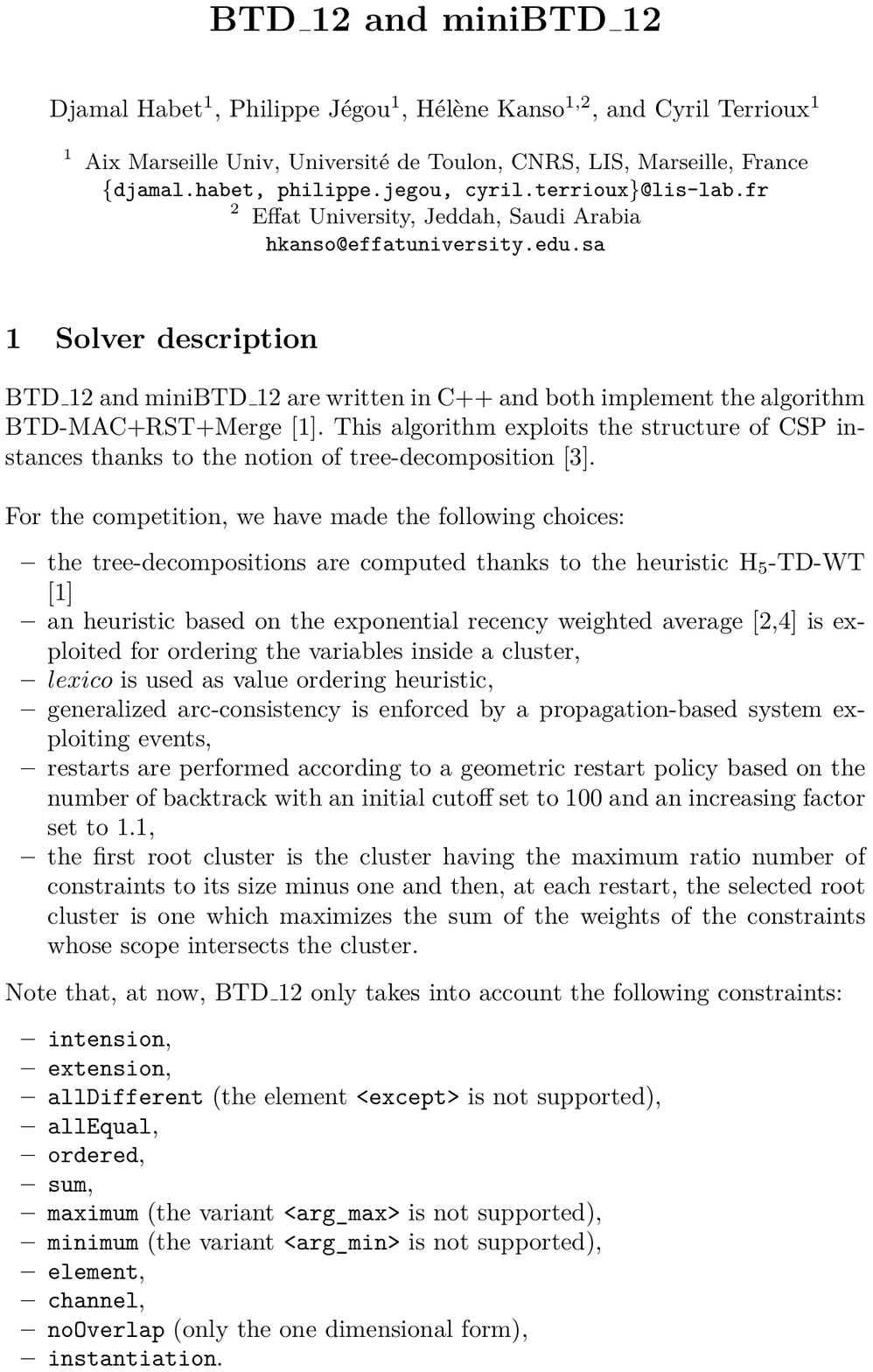}
\addcontentsline{toc}{section}{\numberline{}Choco}
\includepdf[pages=-,pagecommand={\thispagestyle{plain}}]{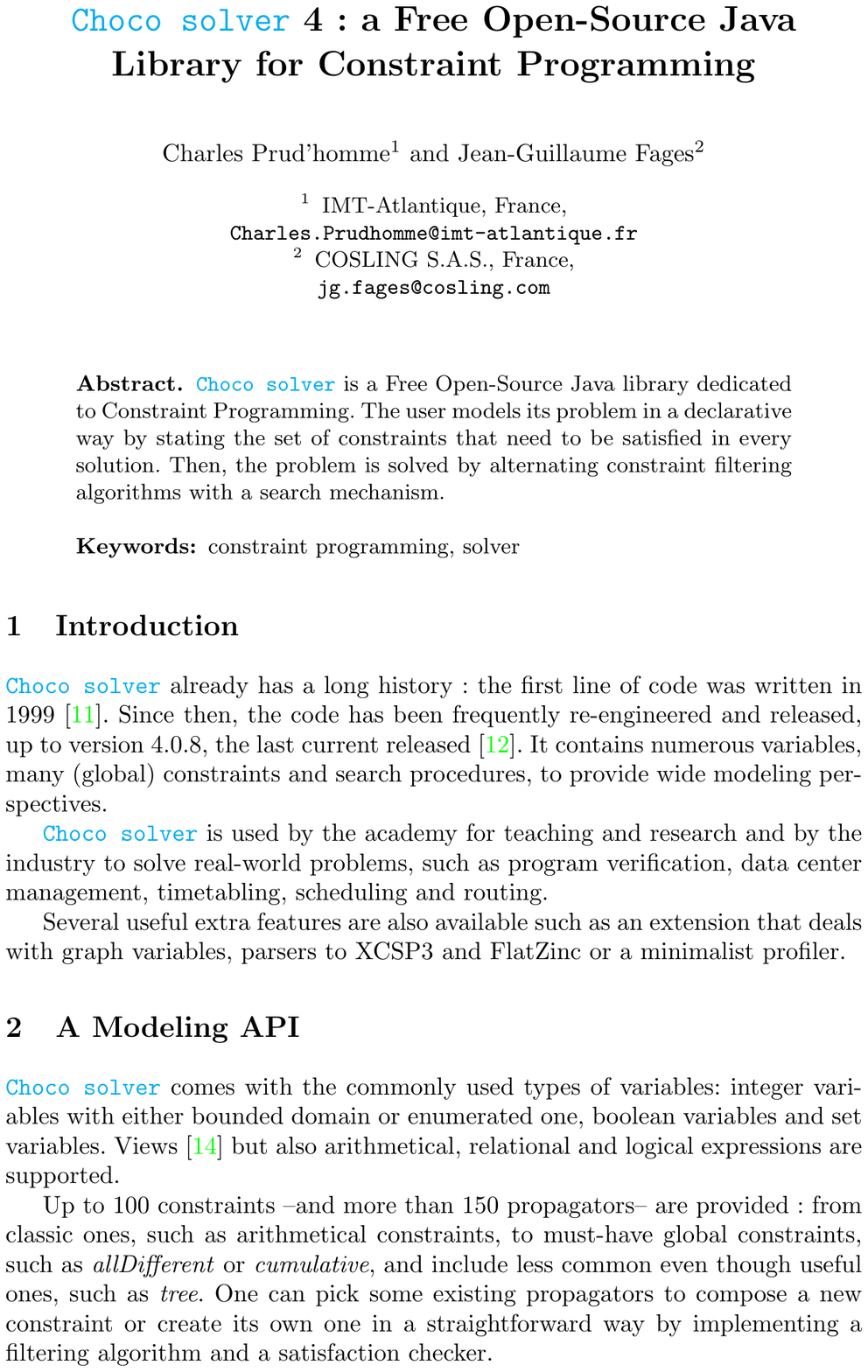}
\addcontentsline{toc}{section}{\numberline{}Concrete}
\includepdf[pages=-,pagecommand={\thispagestyle{plain}}]{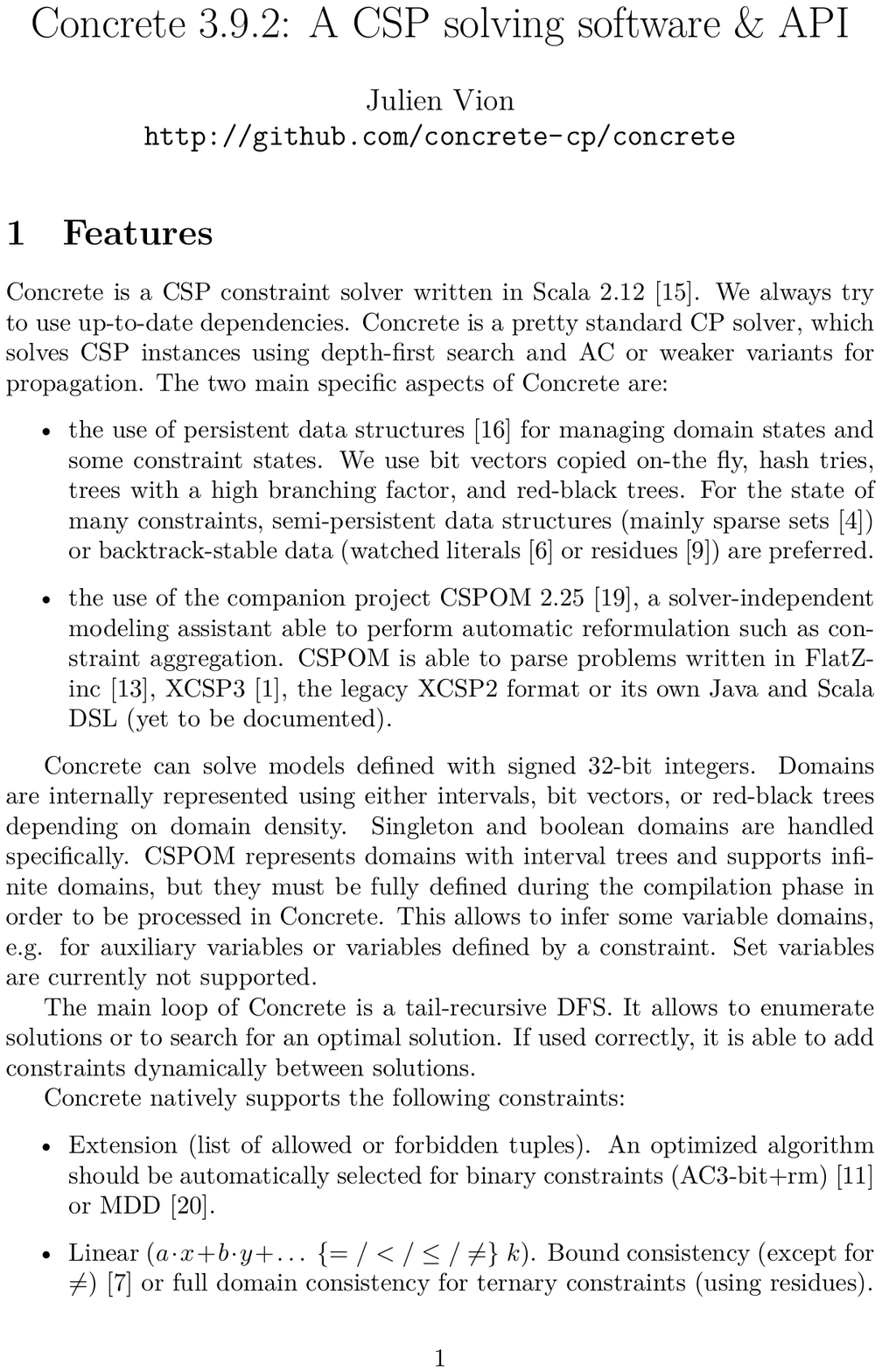}
\addcontentsline{toc}{section}{\numberline{}CoSoCo}
\includepdf[pages=-,pagecommand={\thispagestyle{plain}}]{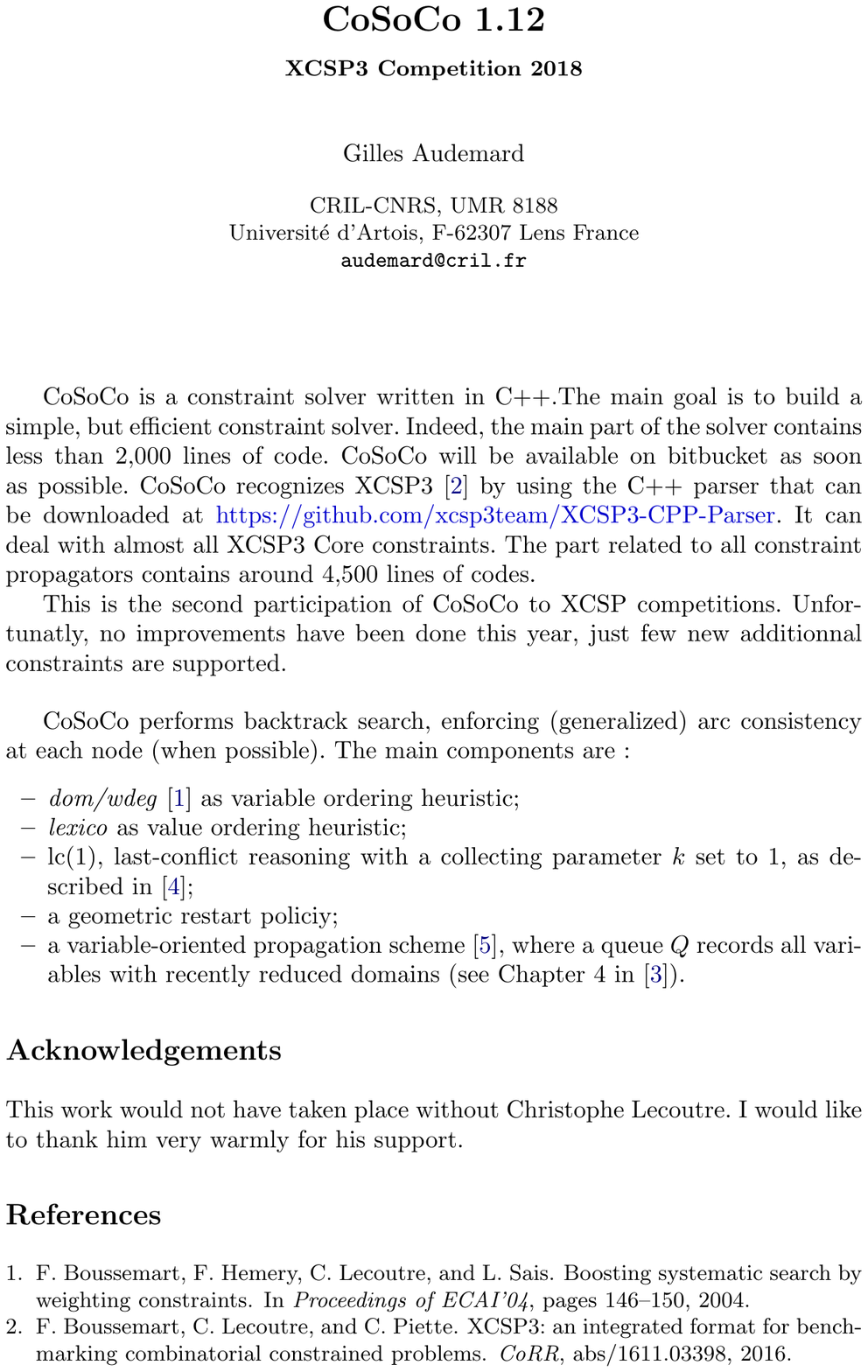}
\addcontentsline{toc}{section}{\numberline{}macht}
\includepdf[pages=-,pagecommand={\thispagestyle{plain}}]{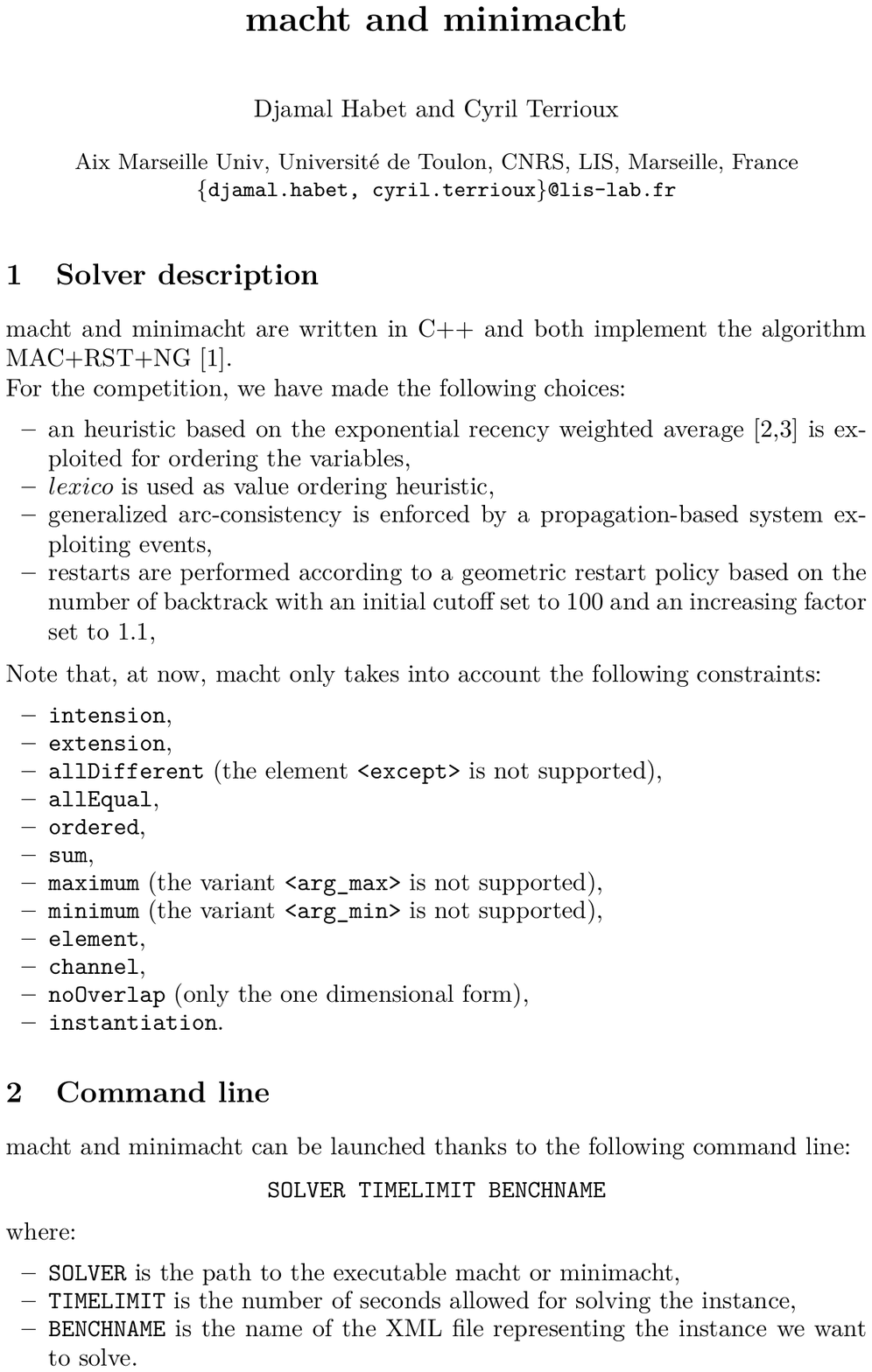}
\addcontentsline{toc}{section}{\numberline{}Mistral}
\includepdf[pages=-,pagecommand={\thispagestyle{plain}}]{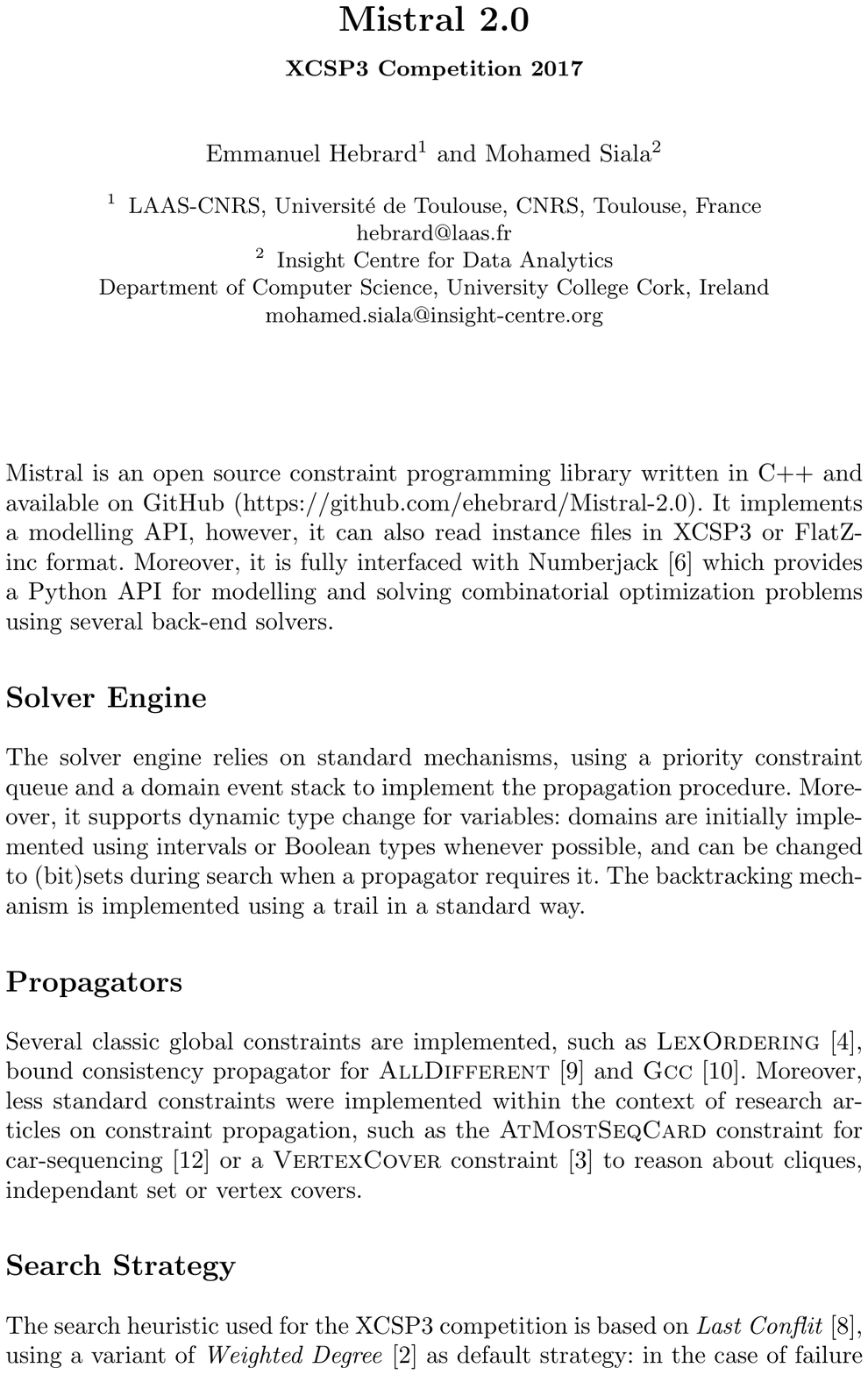}
\addcontentsline{toc}{section}{\numberline{}NACRE}
\includepdf[pages=-,pagecommand={\thispagestyle{plain}}]{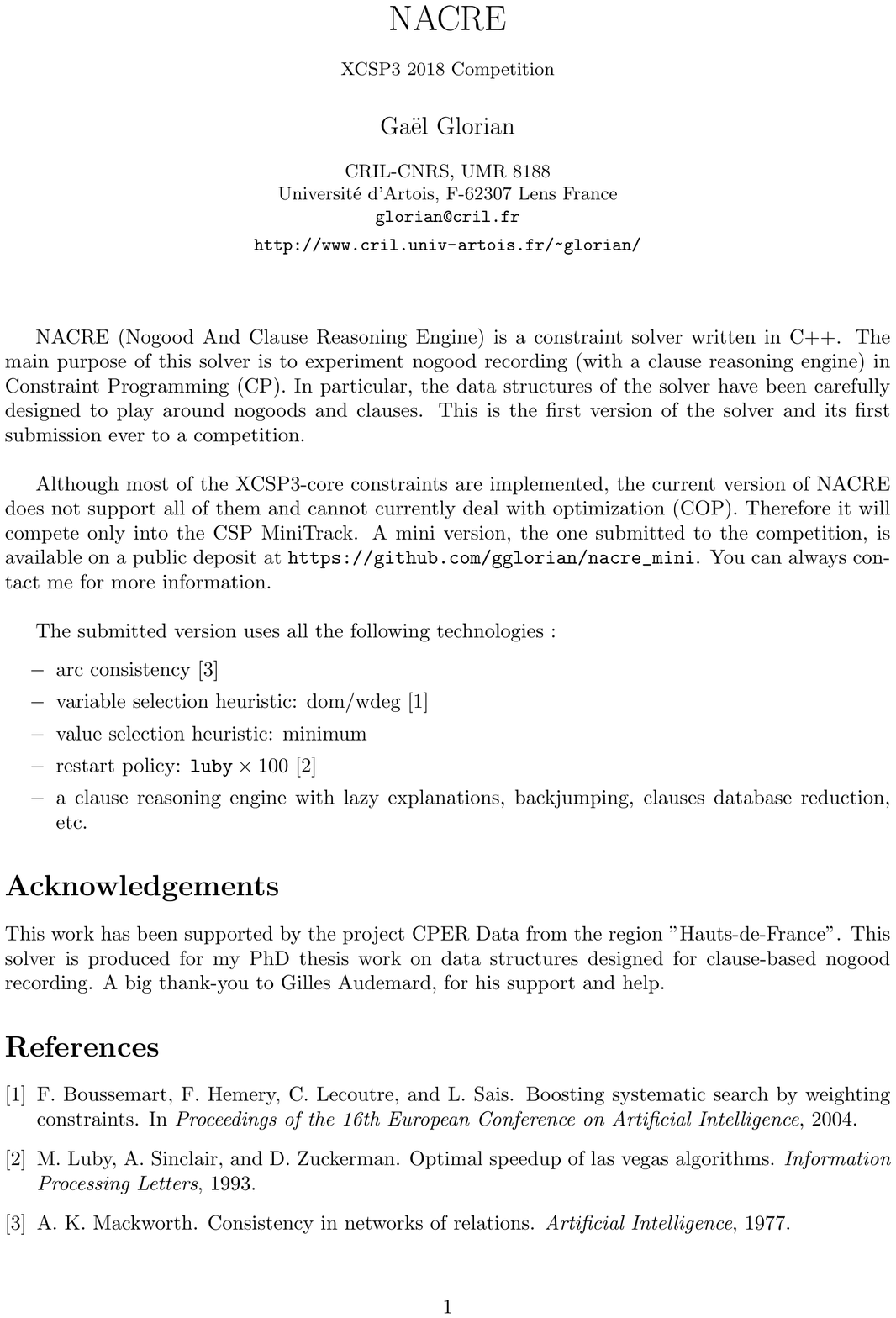}
\addcontentsline{toc}{section}{\numberline{}OscaR}
\includepdf[pages=-,pagecommand={\thispagestyle{plain}}]{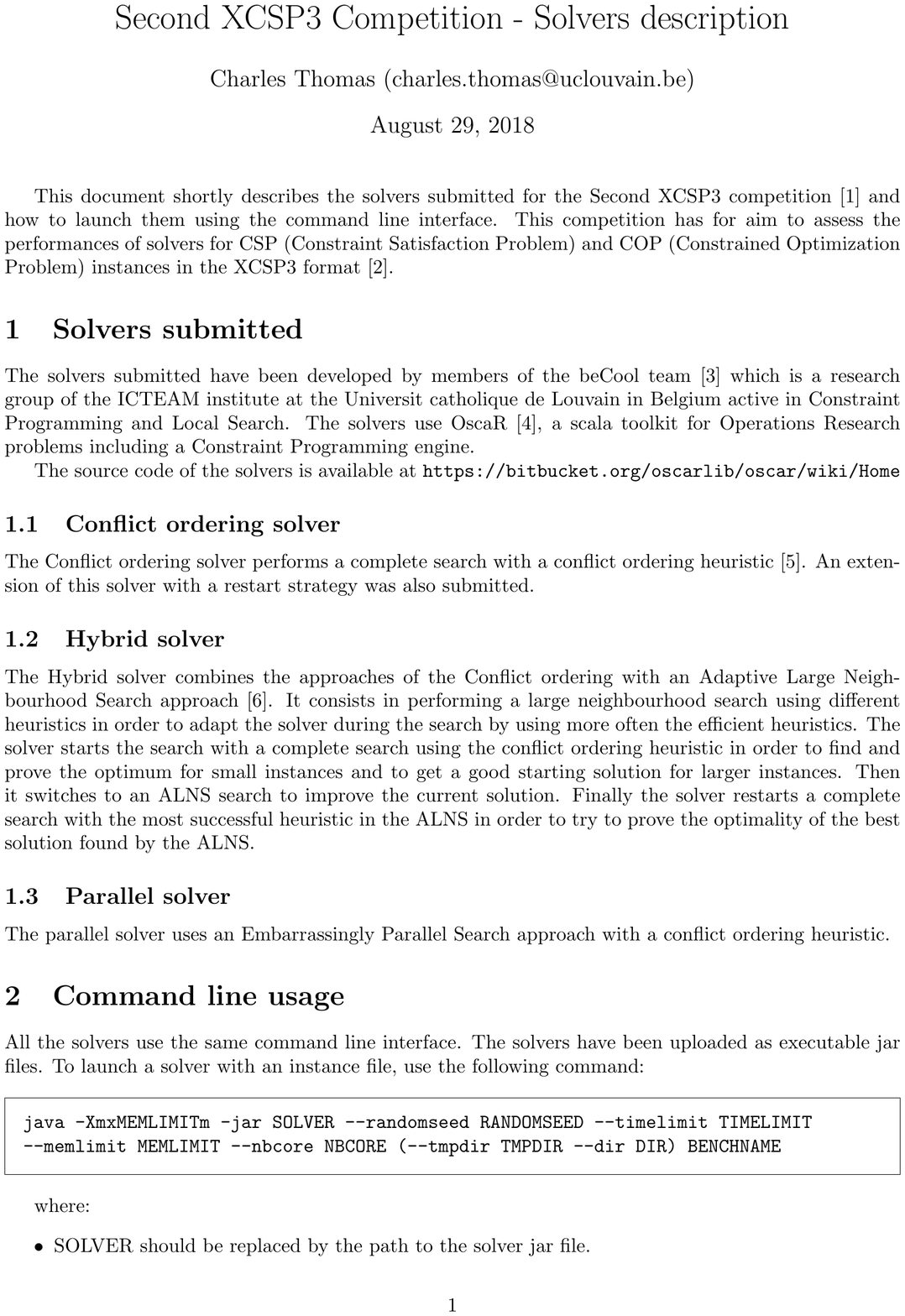}
\addcontentsline{toc}{section}{\numberline{}PicatSAT}
\includepdf[pages=-,pagecommand={\thispagestyle{plain}}]{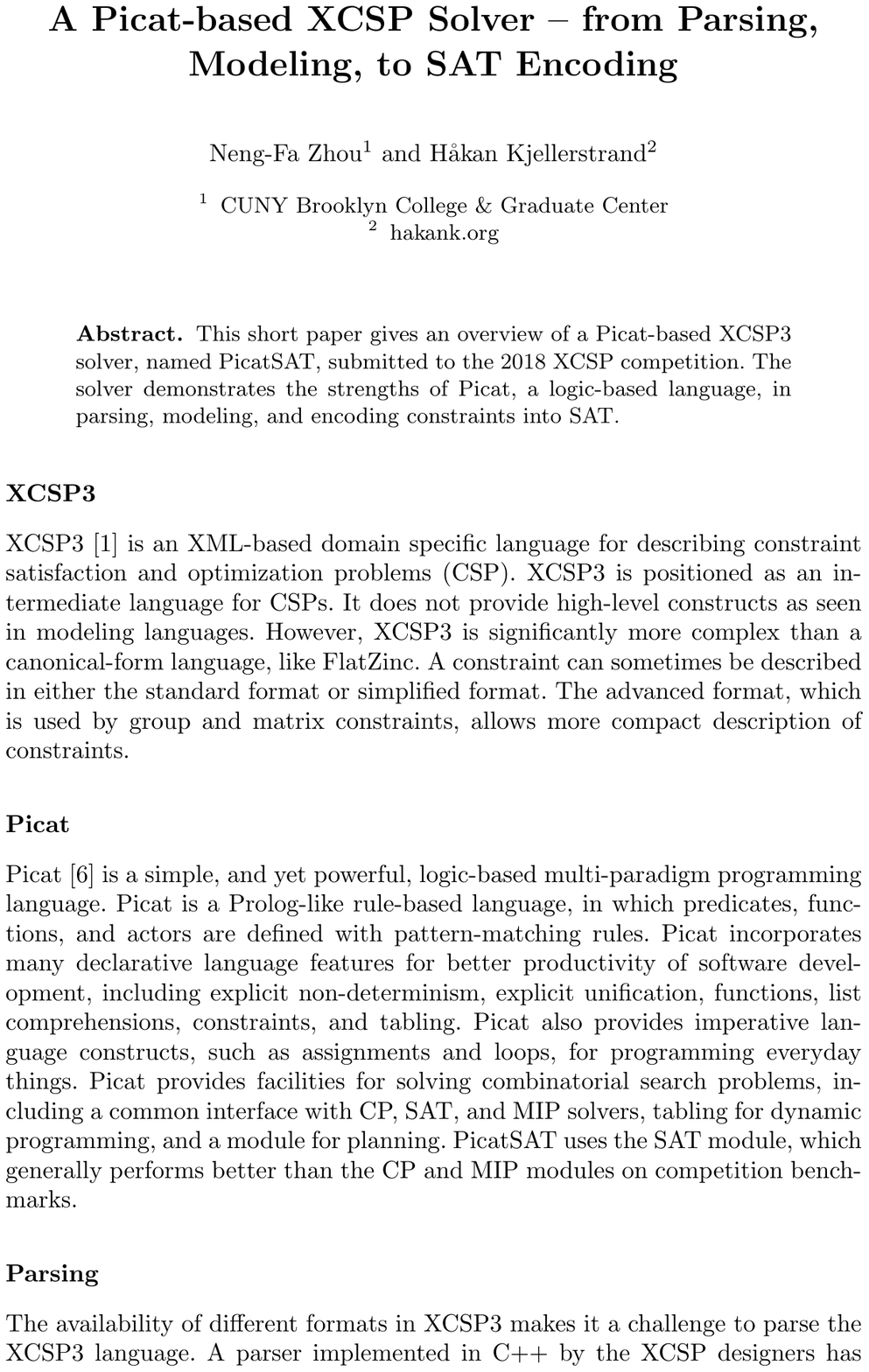}
\addcontentsline{toc}{section}{\numberline{}Sat4j}
\includepdf[pages=-,pagecommand={\thispagestyle{plain}}]{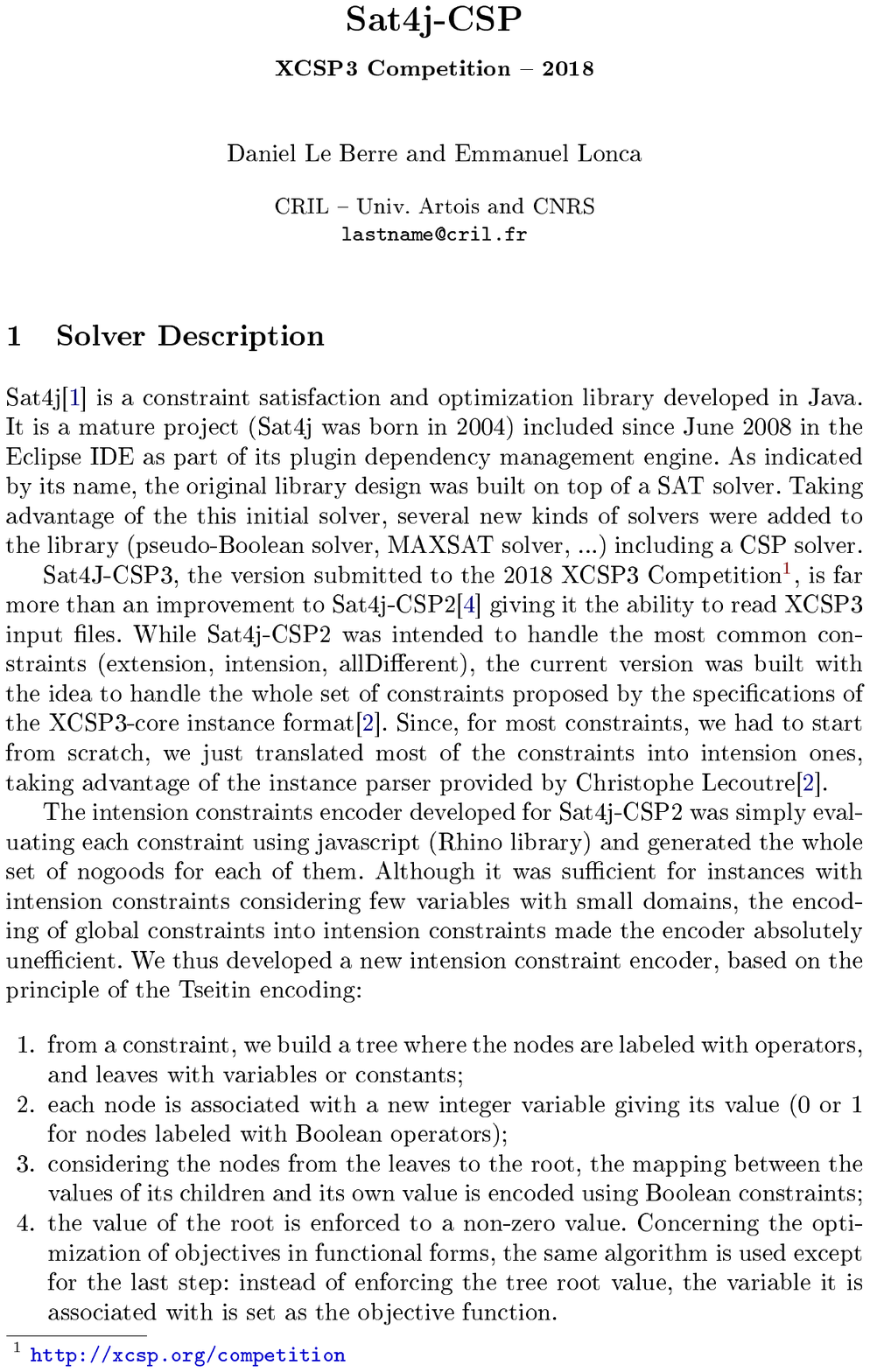}
\addcontentsline{toc}{section}{\numberline{}sCOP}
\includepdf[pages=-,pagecommand={\thispagestyle{plain}}]{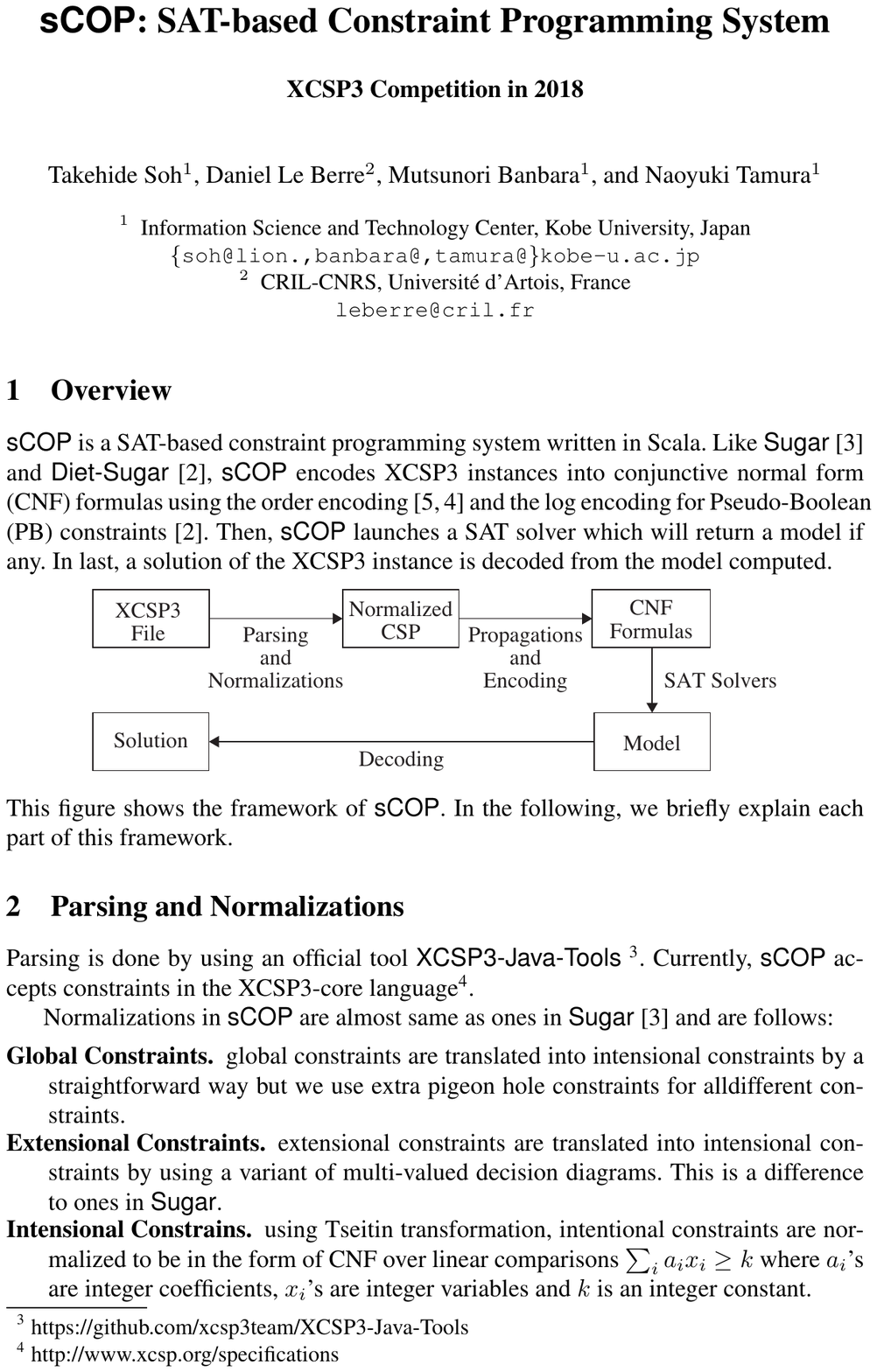}

%\includepdf[nup=1x1, frame, scale=0.9,pages=-,pagecommand={\thispagestyle{plain}}]{BTD_12.pdf}

\chapter{Results}

In this chapter, rankings for the \xt Competition 2018 are given.
We also make a few general comments about the results.
Importantly, remember that you can find all detailed results, including all traces of solvers at \href{http://www.cril.fr/XCSP18/}{http://www.cril.fr/XCSP18/}.

\section{Rankings}

% mini
\newcommand{\solverchgb}{cosoco }
\newcommand{\solverchgc}{GG's minicp}
\newcommand{\solverchgd}{MiniCPFever}
\newcommand{\solverchge}{NACRE }
\newcommand{\solverchgf}{slowpoke}
\newcommand{\solverchgg}{Solver of Schul \& Smal }
\newcommand{\solverchgh}{SuperSolver} %\_Macq\_Stevenart }
\newcommand{\solverchgi}{The dodo solver }
\newcommand{\solverchgj}{minimacht }
\newcommand{\solverchha}{miniBTD\_12 }
\newcommand{\solverchhb}{miniBTD }

% sequential
\newcommand{\solvercheb}{Mistral-2.0 }
\newcommand{\solverchec}{Sat4j-CSP }
\newcommand{\solverched}{OscaR - Conflict Ordering with restarts }
\newcommand{\solverchee}{OscaR - Conflict Ordering }
\newcommand{\solvercheg}{OscaR - Hybrid }
\newcommand{\solvercheh}{Choco-solver {\it 4.0.7 seq (493a269)}}
\newcommand{\solverchei}{BTD\_12 {\it 2018-06-11\_12}}
\newcommand{\solverchej}{BTD {\it 2018.06.11\_3}}
\newcommand{\solverchfa}{Concrete 3.8 {\it 2018-06-13}}
\newcommand{\solverchfb}{Concrete 3.8-SuperNG {\it 2018-06-13}}
\newcommand{\solverchfc}{cosoco {\it 1.12}}
\newcommand{\solverchfd}{macht }
\newcommand{\solverchfe}{PicatSAT }
\newcommand{\solverchff}{scop {\it both+MapleCOMSPS}}
\newcommand{\solverchfg}{scop {\it order+MapleCOMSPS}}
\newcommand{\solverchhi}{PicatSAT }
\newcommand{\solverchhj}{Mistral-2.0 }
\newcommand{\solverchia}{Concrete {\it 3.9.2}}
\newcommand{\solverchib}{Concrete {\it 3.9.2-SuperNG}}
\newcommand{\solverchic}{scop {\it order+MapleCOMSPS}}
\newcommand{\solverchid}{scop {\it both+MapleCOMSPS }}
\newcommand{\solverchie}{macht }
\newcommand{\solverchif}{BTD\_12 }
\newcommand{\solverchig}{BTD }
\newcommand{\solverchih}{Choco-solver {\it 4.0.7b seq }}
\newcommand{\solverchja}{OscaR - Conflict Ordering }
\newcommand{\solverchjb}{OscaR - Conflict Ordering with restarts }
\newcommand{\solverchjc}{OscaR - Hybrid {\it 2018-08-14}}
\newcommand{\solverchje}{PicatSAT {\it 2018-08-14}}
\newcommand{\solverchjj}{OscaR-Conf. Ordering+restarts}

% parallel
\newcommand{\solverchef}{OscaR - Parallel with EPS }
\newcommand{\solverchfi}{scop {\it order+glucose-syrup}}
\newcommand{\solverchfj}{scop {\it both+glucose-syrup}}
\newcommand{\solverchga}{Choco-solver {\it 4.0.7b par}}
\newcommand{\solverchii}{scop {\it order+glucose-syrup}}
\newcommand{\solverchij}{scop {\it both+glucose-syrup }}
\newcommand{\solverchjd}{OscaR - Parallel with EPS }

% fast cop
\newcommand{\solverchhc}{Choco-solver {\it 4.0.7b seq}}
\newcommand{\solverchhd}{Sat4j-CSP}
\newcommand{\solverchhe}{cosoco}
\newcommand{\solverchhg}{Concrete {\it 3.9.2}}
\newcommand{\solverchhh}{Concrete {\it 3.9.2-SuperNG}}
\newcommand{\solverchjf}{OscaR - Conflict Ordering with restarts}
\newcommand{\solverchjg}{OscaR - Hybrid}
\newcommand{\solverciaa}{OscaR - Conflict Ordering with restarts}
\newcommand{\solverciab}{Mistral-2.0}

Remember that the tracks of the competition are given by the following tables:

\begin{table}[h]
\begin{center}
\begin{tabular}{cccc} 
\toprule
\textcolor{dred}{\bf Problem} &  \textcolor{dred}{\bf Goal} &  \textcolor{dred}{\bf Exploration} &  \textcolor{dred}{\bf Timeout} \\
\midrule
CSP  & one solution & sequential & 40 minutes \\
CSP  & one solution & parallel & 40 minutes \\
\midrule
COP  & best solution & sequential & 4 minutes \\
%COP  & best solution & parallel & 4 minutes \\
COP  & best solution & sequential & 40 minutes \\
COP  & best solution & parallel & 40 minutes \\
%\midrule
%COP  & optimal solution & sequential & 40 minutes \\
%COP  & optimal solution & parallel & 40 minutes \\
\bottomrule
\end{tabular}
\end{center}
\caption{Standard Tracks. \label{tab:anysolver}}
\end{table}

\begin{table}[h]
\begin{center}
\begin{tabular}{cccc} 
\toprule
\textcolor{dred}{\bf Problem} &  \textcolor{dred}{\bf Goal} &  \textcolor{dred}{\bf Exploration} &  \textcolor{dred}{\bf Timeout} \\
\midrule
CSP  & one solution & sequential & 40 minutes \\
COP  & best solution & sequential & 40 minutes \\
%COP  & optimal solution & sequential & 40 minutes \\
\bottomrule
\end{tabular}
\end{center}
\caption{Mini-Solver Tracks. \label{tab:minisolver}}
\end{table}

\noindent Also, note that:

\begin{itemize}
\item The cluster was provided by CRIL and is composed of nodes with two quad-cores (Intel @ 2.67GHz with 32 GiB RAM).
\item Hyperthreading was disabled. % for the final runs.
\item Sequential solvers were run on one processor (4 cores) and were allocated $15,500$ MiB of memory. 
\item Parallel solvers were run on two processors (8 cores) and were allocated $31,000$ MiB of memory.
\item The selection of instances for the Standard tracks was composed of 236 CSP and 346 COP instances.
\item The selection of instances for the Mini-solver tracks was composed of 176 CSP and 188 COP instances.
\end{itemize}
%\noindent Concerning the selection of instances, we end up with: 
%\begin{itemize}
%\item Standard tracks: 236 CSP and 346 COP instances
%\item Mini-solver tracks: 176 CSP and 188 COP instances 
%\end{itemize}

\paragraph{About the Ranking.}
It is based on the number of times a solver is able to prove a result (satisfiability for CSP, optimality for COP).
Notice that for COP, another viewpoint is the number of times a solver can give the best known answer (optimality or best known bound); it is then given between parentheses in some tables.
%In that case, we give  

%\bigskip
%\paragraph{This chapter will be completed soon. Meanwhile, see all detailed results on \href{http://www.cril.fr/XCSP18/}{http://www.cril.fr/XCSP18/}}

%\subsection{Standard Tracks}

\bigskip\noindent
Tables \ref{tab:standardCSP}, \ref{tab:standardCOP} and \ref{tab:standardCOPFast} respectively give rankings for sequential (standard) solvers on CSP, COP and 'fast COP' instances.
Table \ref{tab:minisolverCSP} gives the ranking for parallel solvers on CSP instances; note that there was not enough contestants for having a relevant ranking for parallel solvers on COP instances (but look at good results of \solverchga~ on the website). 
Finally, Tables \ref{tab:minisolverCSP} and \ref{tab:minisolverCOP} respectively give rankings for mini-solvers on CSP and COP instances.

\begin{table}[p]
  \begin{center}
    \begin{footnotesize}
      \begin{tabular}{clcccc}
\toprule
 &  & \#solved &  & \%inst. & \%VBS\\
\midrule
\midrule
\multicolumn{2}{c}{\it Virtual Best Solver (VBS)} & 164 & ~ 104 SAT, 60 UNSAT ~ & 69\% & 100\%\\
\midrule
\midrule
1 & \solverchic & 146 & 92 SAT, 54 UNSAT & 62\% & 89\% \\
\midrule
2 & \solverchid & 140 & 87 SAT, 53 UNSAT & 59\% & 85\% \\
\midrule
3 & \solverchje & 138 & 85 SAT, 53 UNSAT & 58\% & 84\% \\
\midrule
4 & \solverchhj & 116 & 80 SAT, 36 UNSAT & 49\% & 71\% \\
\midrule
5 & \solverchih & 115 & 77 SAT, 38 UNSAT & 49\% & 70\% \\
\midrule
6 & \solverchia & 92 & 64 SAT, 28 UNSAT & 39\% & 56\% \\
\midrule
7 & \solverchjj & 90 & 62 SAT, 28 UNSAT & 38\% & 55\% \\
\midrule
8 & \solverchib & 84 & 55 SAT, 29 UNSAT & 36\% & 51\% \\
\midrule
9 & \solverchec & 83 & 40 SAT, 43 UNSAT & 35\% & 51\% \\
\midrule
10 & \solverchja & 81 & 51 SAT, 30 UNSAT & 34\% & 49\% \\
\midrule
11 & \solverchfc & 79 & 53 SAT, 26 UNSAT & 33\% & 48\% \\
\midrule
12 & \solverchif & 76 & 32 SAT, 44 UNSAT & 32\% & 46\% \\
\midrule
13 & \solverchig & 76 & 31 SAT, 45 UNSAT & 32\% & 46\% \\
\midrule
14 & \solverchie & 66 & 33 SAT, 33 UNSAT & 28\% & 40\% \\
\bottomrule
\end{tabular}
\caption{Ranking for (Standard Track -- CSP -- sequential) over 236 instances. \label{tab:standardCSP}}

%\captionof{table}{Ranking for (Standard Track -- CSP -- sequential) over 236 instances. \label{tab:standardCSP}}
%    \end{scriptsize}
%    \end{center}
%    
%\end{table}
%
%\begin{table}[p]
%\begin{center}
%\begin{tiny}
\bigskip
\begin{tabular}{clcccc}
\toprule
 &  & \#solved &  & \%inst. & \%VBS\\
\midrule
\midrule
\multicolumn{2}{c}{\it Virtual Best Solver (VBS)} & 146 & ~ 146 OPT~ & 42\% & 100\%\\
\midrule
\midrule
1 & \solverchje & 132 (132) & 132 OPT & 38\% & 90\% \\
\midrule
2 & \solverchia & 105 (148) & 105 OPT & 30\% & 72\% \\
\midrule
3 & \solverchih & 102 (154) & 102 OPT & 29\% & 70\% \\
\midrule
4 & \solverchjj & 99 (132) & 99 OPT & 29\% & 68\% \\
\midrule
5 & \solverchib & 99 (139) & 99 OPT & 29\% & 68\% \\
\midrule
6 & \solverchfc & 64 (112) & 64 OPT & 18\% & 44\% \\
\midrule
7 & \solverchjc & 61 (132) & 61 OPT & 18\% & 42\% \\
\midrule
8 & \solverchec & 54 (86) & 54 OPT & 16\% & 37\% \\
\bottomrule
\end{tabular}
\caption{Ranking for (Standard Track -- COP -- sequential) over 346 instances. Between parentheses, the number of times the solver can give the best known result. % (not necessarily, a proved optimal one).
  \label{tab:standardCOP}}
%\captionof{table}{Ranking for (Standard Track -- COP -- sequential) over 346 instances. Between parentheses, the number of times the solver can give the best known result (not necessarily, a proved optimal one).\label{tab:standardCOP}}
%\end{scriptsize}
%\end{center}
%\end{table}
%\begin{table}[p]
%\begin{center}
%\begin{tiny}

\bigskip
\begin{tabular}{clccc}
\toprule
 &  & \#best & \%inst. & \%VBS\\
\midrule
\midrule
\multicolumn{2}{c}{\it Virtual Best Solver (VBS)} & 316 & 91\% & 100\%\\
\midrule
\midrule
1 & \solverchhg & 151 & 44\% & 48\% \\
\midrule
2 & \solverchhc & 146 & 42\% & 46\% \\
\midrule
3 & \solverchjg & 139 & 40\% & 44\% \\
\midrule
4 & \solverciaa & 133 & 38\% & 42\% \\
\midrule
5 & \solverchhh & 129 & 37\% & 41\% \\
\midrule
6 & \solverciab & 123 & 36\% & 39\% \\
\midrule
7 & \solverchhe & 107 & 31\% & 34\% \\
\midrule
8 & \solverchhd & 78 & 23\% & 25\% \\
\bottomrule
\end{tabular}
\caption{Ranking for (Standard Track -- COP -- sequential - 4') over 346 instances. For this fast track, we consider the number of times the solver gives the best known result. %(not necessarily, a proved optimal one).
  \label{tab:standardCOPFast}}
\end{footnotesize}
\end{center}
\end{table}

\begin{table}[p]
  \begin{center}
    \begin{footnotesize}
\begin{tabular}{clcccc}
\toprule
 &  & \#solved &  & \%inst. & \%VBS\\
\midrule
\midrule
\multicolumn{2}{c}{\it Virtual Best Solver (VBS)} & 168 & 104 SAT, 64 UNSAT & 71\% & 100\%\\
\midrule
\midrule
1 & \solverchii & 151 & 95 SAT, 56 UNSAT & 64\% & 90\% \\
\midrule
2 & \solverchij & 138 & 82 SAT, 56 UNSAT & 58\% & 82\% \\
\midrule
3 & \solverchga & 134 & 88 SAT, 46 UNSAT & 57\% & 80\% \\
\midrule
4 & \solverchjd & 89 & 56 SAT, 33 UNSAT & 38\% & 53\% \\
\bottomrule
\end{tabular}
\caption{Ranking for (Standard Track -- CSP -- parallel) over 236 instances.\label{tab:standardCSPPar}}

%\bigskip
%Not enough contestants for having a relevant ranking, but look at good results of \solverchga~ 
%\caption{Ranking for (Standard Track -- COP -- parallel) over 346 instances.\label{tab:standardCSOPar}}

 \bigskip
\begin{tabular}{clcccc}
\toprule
  &  & \#solved &  & \%inst. & \%VBS\\
\midrule
\midrule
\multicolumn{2}{c}{\it Virtual Best Solver (VBS)} & 113 & ~ 53 SAT, 60 UNSAT ~ & 64\% & 100\%\\
\midrule
\midrule
1 & \solverchge & 86 & 43 SAT, 43 UNSAT & 49\% & 76\% \\
\midrule
2 & \solverchha & 79 & 36 SAT, 43 UNSAT & 45\% & 70\% \\
\midrule
3 & \solverchhb & 75 & 32 SAT, 43 UNSAT & 43\% & 66\% \\
\midrule
4 & \solverchgb & 72 & 42 SAT, 30 UNSAT & 41\% & 64\% \\
\midrule
5 & \solverchgj & 69 & 37 SAT, 32 UNSAT & 39\% & 61\% \\
\midrule
6 & \solverchgc & 56 & 37 SAT, 19 UNSAT & 32\% & 50\% \\
\midrule
7 & \solverchgg & 54 & 23 SAT, 31 UNSAT & 31\% & 48\% \\
\midrule
8 & \solverchgd & 54 & 34 SAT, 20 UNSAT & 31\% & 48\% \\
\midrule
9 & \solverchgf & 38 & 38 SAT & 22\% & 34\% \\
\midrule
10 & \solverchgh & 31 & 31 SAT & 18\% & 27\% \\
\midrule
11 & \solverchgi & 25 & 25 UNSAT & 14\% & 22\% \\
\bottomrule
\end{tabular}
\caption{Ranking for (Mini-Solver Track -- CSP) over 176 instances. \label{tab:minisolverCSP}}
%\captionof{table}{Ranking for (Mini-Solver Track -- CSP) over 176 instances. \label{tab:minisolverCSP}}

\bigskip
\begin{tabular}{clcccc}
\toprule
 &  & \#solved &  & \%inst. & \%VBS\\
\midrule
\midrule
\multicolumn{2}{c}{\it Virtual Best Solver (VBS)} & 48 & ~ 48 OPT ~ & 26\% & 100\%\\
\midrule
\midrule
1 & \solverchgb & 46 (122) & 46 OPT & 24\% & 96\% \\
\midrule
2 & \solverchgg & 35 (44) & 35 OPT & 19\% & 73\% \\
\midrule
3 & \solverchgc & 3 (22) & 3 OPT & 2\% & 6\% \\
\midrule
4 & \solverchgd & 0 (50) &  & 0\% & 0\% \\
\midrule
5 & \solverchgh & 0 (23) &  & 0\% & 0\% \\
\midrule
6 & \solverchgi & 0 (18) &  & 0\% & 0\% \\
\midrule
7 & \solverchgf & 0 (12) &  & 0\% & 0\% \\
\bottomrule
\end{tabular}
\caption{Ranking for (Mini-Solver Track -- COP) over 188 instances. Between parentheses, the number of times the solver can give the best known result. %(not necessarily, a proved optimal one).
  \label{tab:minisolverCOP}}
%\captionof{table}{Ranking for (Mini-Solver Track -- COP) over 188 instances. Between parentheses, the number of times the solver can give the best known result (not necessarily, a proved optimal one).  \label{tab:minisolverCOP}}
\end{footnotesize}
\end{center}
\end{table}

%\caption{Mini-Solver -- COP -- 188 instances. Between parentheses, the number of times the solver can give the best known result (not necessarily, a proved optimal one).  \label{tab:minisolverCOP}}
%\end{table}

\section{A few Comments}

For simplicity, the solvers that encode constraints in SAT (PicatSAT, Sat4j and sCOP)  will be called {\em SAT-based} solvers.
The solvers that use more traditional CP techniques (Choco, Concrete, CoSoCo, Mistral, and OscaR) will be called {\em CP-based} solvers.
Nacre is hybrid as it involves a strong clause reasoning engine.
Finally, the solvers that exploit tree decomposition techniques (BTD, BTD\_12, and macht) will be called {\em decomposition-based} solvers.

Overall, one can observe that three types of solvers are rather complementary.
It means that there are problems where SAT-based solvers, CP-based solvers and decomposition-based solvers are clearly superior.
This is an interesting lesson from this competition. 

\subsection{Standard Track -- CSP -- Sequential}

\paragraph{BIBD}
Recall that two series 'sum' and 'sc' of 6 instances each have been selected.
The series 'sc' is obtained by introducing auxiliary variables.
Rather surprisingly, the solvers have very similar results on instances from the two series.
Best solvers are PicatSAT, sCOP and Mistral.

\paragraph{Car Sequencing}
SAT-based solvers sCOP and PicatSAT are the best competitors, i.e., most robust, for this problem, succeeding in solving all (19) instances.

\paragraph{ColouredQueens}
All Solvers have difficulty in scaling up. All competitors only solve 3 or 4 instances out of 12.

\paragraph{Crosswords}
These instances involve large table constraints.
8 instances out of 13 remain unsolved.
Although the results obtained by the competitors are close, BTD appears to be the most efficient solver (decomposition seems to be effective).

\paragraph{Dubois}
Unsurprisingly, because of the SAT origin of the problem, SAT-based solvers PicatSAT and sCOP are very efficient. Interestingly, BTD is also able to solve all instances within a few milliseconds.

\paragraph{Eternity}
Mistral and Choco are the best competitors.
7 instances out of 15 remain unsolved.

\paragraph{Frb.}
On these random instances from Model RB (and forced to be satisfiable), sCOP slightly outperforms Concrete, CoSoCo and OscaR.
9 instances out of 16 remain unsolved.

\paragraph{Graceful Graph}
All solvers are close in term of performance, with Mistral succeeding in solving one more instance. 4 instances out of 11 remain unsolved. 

\paragraph{Haystack}
SAT-based solvers Picat-SAT and sCOP are the best competitors, solving all instances, followed by BTD. Learning and decomposition methods are quite operational for this problem.

\paragraph{Langford}
OscaR is the best competitor, followed by Mistral, PicatSAT and sCOP. Note that CoSoCo can be very fast for solving some instances.
Only one instance out of 11 remains unsolved.

\paragraph{MagicHexagon}
Concrete, CoSoCo, and Mistral are the best competitors. Sat-Based solvers and BTD are not very effective for this problem.
5 instances out of 11 remain unsolved.

\paragraph{MisteryShopper}
Mistral is the best competitor, followed by Choco and Concrete.
All instances have been solved.

\paragraph{PseudoBoolean}
sCOP is the best competitor, followed by Mistral.
5 instances out of 13 remain unsolved.

\paragraph{QuasiGroups}
sCOP is the best competitor, followed by macht and PicatSAT.
7 instances out of 16 remain unsolved.

\paragraph{RLFAP}
The most difficult instance from this classical (decision version of this) problem, is only solved by SAT-based solvers sCOP and PicatSAT as well as BTD. sCOP appears to be very robust and regular whereas BTD can be very efficient at solving some of these instances.   

\paragraph{SocialGolfers}
Compared to Choco, Concrete, Mistral and OscaR, SAT-based solvers PicatSAT and sCOP are able to solve 2 additional instances. Note how PicatSAT is very efficient for this problem. 4 instances out of 12 remain unsolved.

\paragraph{SportsScheduling}
Oscar and Mistral obtain the best results.
5 instances out of 10 remain unsolved.

\paragraph{StripPacking}
Sat-based solvers sCOP and PicatSAT are the best competitors.

\paragraph{SubIsomorphism}
Choco, followed by BTD and Mistral.
3 instances out of 11 instances remain unsolved.

\subsection{Standard Track -- COP -- Sequential}

Note that Mistral is sometimes mentioned during analysis; although it has been disqualified for this track. %, we think that for many problems, it behaves correctly. 

\paragraph{Auction}
OscaR is clearly the best competitor for this problem, always giving the best bounds. Choco is the best second solver. There is no clear advantage of using the model variant 'cnt' (with \gb{atMost1}) or the model variant 'sum'.

\paragraph{BACP}
PicatSAT is the only solver able to prove optimality for 4 instances.
On the other instances, solvers are rather close, almost always providing the same bounds. There is no clear advantage of using the model variant 'm1' or the model variant 'm2' that reformulates some constraints (to be compatible with the restrictions of the Mini Tracks).

\paragraph{CrosswordsDesign}
This difficult optimization problem involves large short tables.
For example, for $n=10$, there are 20 tables containing $186,809$ tuples and 20 other tables containing $143,417$ tuples.
CoSoCo is the best performer, succeeding in finding solutions for the most difficult instances. Concrete has also a good behavior on some instances of intermediary difficulty. Because of the huge size of some variable domains, Compact-Table \cite{DHLPPRS_efficiently} may be not the best table propagator.

\paragraph{FAPP}
It is important to note that the model used for the competition exactly corresponds to the original and complete formulation of the problems used for the 2001 ROADEF challenge. Results obtained by teams are available at \url{http://www.roadef.org/challenge/2001/fr/resultats_phase1.php}
On instance m2s-01-0200, Choco finds a solution with $k=4$ (the main parameter related to some form of violation).
On instance m2s-02-0250, Concrete finds a solution with $k=2$.
On instance m2s-03-0300, Choco and Concrete find a solution with $k=7$.
This corresponds to the best bounds found in 2001 by the best teams.
Although it is rather unfair to compare results that are 17 years away, one can observe that a pure CP approach (benefiting from progress on table constraints) can be very effective (and note that no specific tuning is possible in the context of the XCSP competition).
Note that optimality has been proved for 5 instances (and also for 2 easy ones); PicatSAT being the unique solver in proving optimality of instance test-01-0150.

\paragraph{GolombRuler}
Concrete, CoSoCo, Mistral and OscaR are the best competitors.
When decision variables are not provided (see instances with 'nodv' in their names), solvers are far less efficient in general. 

\paragraph{GraphColoring}
Concrete is the most robust solver for this problem, while CoSoCo is usually very fast.

\paragraph{Knapsack}
Mistral and OscaR most often prove optimality. 
4 instances out of 11 remains unsolved (to optimality).

\paragraph{LowAutocorrelation}
PicatSAT is impressive on this problem (involving intensional and sum constraints): it proves optimality for 10 instances (out of 14).
Mistral is also very efficient on this problem.

\paragraph{Mario}
Surprisingly, four solvers (Choco, Concrete, Mistral and OScaR) have solved all 10 instances of the series.

\paragraph{NurseRostering}
The results are interesting.
While all solvers prove optimality on the two first instances, OscaR gives the best bound on the third instance, Sat4j gives the best bounds on the next 3 instances, Choco gives the best bounds on the next 6 instances, and finally CoSoCo gives the best bounds on the next 4 instances. This is rather curious.

\paragraph{PeacableArmies}
On model variant m1, Oscar is the best competitor.
On model variant m2, obtained bounds are usually worse; Sat4j and OscaR being dominating.

\paragraph{PizzaVoucher}
SAT-based solvers are very competitive for this problem.
PicatSAT proves optimality on 7 instances (out of 10), and Sat4j provides the best bounds for the three most difficult instances.

\paragraph{PseudoBoolean}
Overall, solvers embedding SAT techniques (PicatSAT, and SAT4j) are good competitors for this problem, although there is no clear winner. CoSoCo and Choco also obtains good results.

\paragraph{QuadraticAssignment}
This is a classical problem.
Clearly, OscaR is the best competitor, being ranked first 16 times (out of 19 instances). Note that SAT-based solvers have strong difficulties.

\paragraph{RCPSP}
This is another classical problem.
OscaR is impressive here: proving rapidly optimality for 14 instances (out of 16 instances). Mistral, Choco and Concrete can also be very fast for many instances. Finally, PicatSAT is rather robust, succeeding in proving optimality for many instances.

\paragraph{RLFAP}
It is important to note that the model used for the competition exactly corresponds to the original and complete formulation of the problem (see \cite{CGLSW_radio}).
CoSoCo is the most robust solver, being ranked first 13 times (out of 14+11 instances). Sat4j is also ranked first 7 times. Choco and Concrete also usually obtain interesting results.

\paragraph{SteelMillSlab}
Except for the very simple instances, no instances from models 'm1' and 'm2' have been solved at all. For model 'm2s', CoSoCo and Concrete obtain the best results.

\paragraph{StillLife}
PicatSAT is quite impressive here: proving optimality for all instances. All other solvers can do that for at most 3 instances.

\paragraph{SumColoring}
PicatSAT proves optimality for 5 instances. OscaR and Choco are rather robust.

\paragraph{Tal}
This is a problem of natural language processing.
OscaR is the most robust solver.
Choco also obtains good results.

\paragraph{TemplateDesign}
Instances of model variant 'm1s' are similar to those of model variant 'm1', except that symmetry breaking constraints are not present. Rather surprisingly, solver usually obtain better results without such constraints.
The best competitor is Choco, followed by PicatSAT and Concrete.

\paragraph{TravellingTournament}
Choco obtains the best results, followed by OscaR and Concrete.
Optimality has only been proved for two instances.

\paragraph{TravellingSalesman}
Optimality is proved for 3 instances by PicatSAT.
CoSoCo and OscaR also obtain good results.

%\subsection{Mini Tracks}

%\subsection{Parallel Tracks}

%\subsection{COP - Fast Track}

%\bibliographystyle{plain} %alpha}
%\bibliography{./globalBiblio}

\end{document}